\pdfoutput=1
\documentclass[10pt,twocolumn,letterpaper]{article}

\usepackage{wacv}
\usepackage{times}
\usepackage{epsfig}
\usepackage{graphicx}
\usepackage{amsmath}
\usepackage{amssymb}
\usepackage{float}
\usepackage{subfig}
\usepackage[export]{adjustbox}
\usepackage{siunitx}
\usepackage{dblfloatfix}
\makeatletter
\@namedef{ver@everyshi.sty}{}
\makeatother
\usepackage{tikz}
\usepackage{xcolor}
\usepackage{bm}
 
\def\wacvPaperID{} %

\wacvfinalcopy %

\ifwacvfinal
\def\assignedStartPage{1} %
\fi

\ifwacvfinal
\usepackage[breaklinks=true,bookmarks=false]{hyperref}
\else
\usepackage[pagebackref=true,breaklinks=true,colorlinks,bookmarks=false]{hyperref}
\fi

\ifwacvfinal
\else
\fi

\begin{document}

\newcommand{\todo}[1]{{\color{red}\textbf{\textit{note: #1}}}}
\newcommand{\loss}{\mathcal{L}}
\def\Wrt{W.r.t\onedot}
\title{VAE-CE: Visual Contrastive Explanation using Disentangled VAEs}

\author{Yoeri Poels\hspace{1cm}Vlado Menkovski\\
Eindhoven University of Technology, the Netherlands
\\
{\tt\small \{y.r.j.poels, v.menkovski\}@tue.nl}
}

\maketitle

\begin{abstract}
The goal of a classification model is to assign the correct labels to data. In most cases, this data is not fully described by the given set of labels. Often a rich set of meaningful concepts exist in the domain that can much more precisely describe each datapoint. Such concepts can also be highly useful for interpreting the model's classifications. %
In this paper we propose a model, 
denoted as Variational Autoencoder-based Contrastive Explanation (VAE-CE), that represents data with high-level concepts and uses this representation for both classification and generating explanations. The explanations are produced in a contrastive manner, conveying why a datapoint is assigned to one class rather than an alternative class. An explanation is specified as a set of transformations of the input datapoint, with each step depicting a concept changing towards the contrastive class.
We build the model using a disentangled VAE, extended with a new supervised method for disentangling individual dimensions.
An analysis on synthetic data and MNIST shows that the approaches to both disentanglement and explanation provide benefits over other methods\footnote{Code is available at \ifwacvfinal
\url{https://github.com/yoeripoels/vce}
\else
\url{}
\fi
}.

\end{abstract}

\section{Introduction}
\begin{figure}
    \centering
    \includegraphics[width=0.8\linewidth]{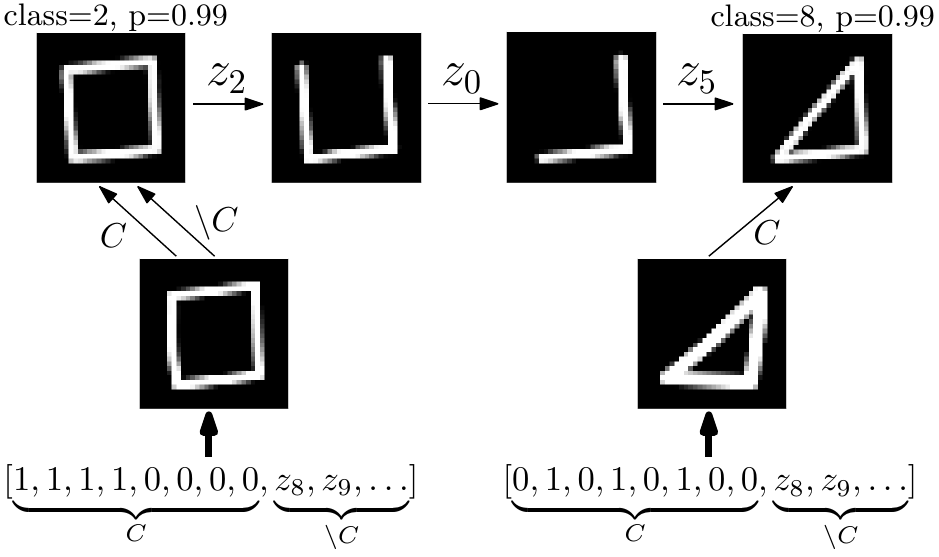}
    \caption{An example of an explanation where lines are concepts, and combinations thereof define classes. The query (left) and exemplar (right) datapoint differ in some class-relevant concepts in $C$. This difference is conveyed by transforming in domain $C$, one concept at a time.}
    \label{fig:example}
\end{figure}

Discriminative models for classification based on deep neural networks achieve outstanding performance given a sufficient amount of training data. They are highly practical as they can be trained in an end-to-end fashion to develop a map $f: X \rightarrow Y$ given pairs of datapoints and labels $(x, y)$, with $x \in X$ and $y\in Y$. Much of this success is due to their hierarchical nature, which allows them to learn an effective high-level representation of low-level input data. However, these characteristics are also the reason for one of their major limitations. Even though the models learn high-level representations, in most cases the model's reasoning is difficult to interpret. The learned representations are often hard to align with existing concepts in the domain. So, these models are commonly considered black boxes that directly map observations to target variables. Such black-box models often lack user trust\cite{kim2015ibcm,mercado2016intelligent}, as we cannot accurately gauge why they make the predictions they do.

Many interpretability approaches have been proposed that focus on developing interpretations of models' decisions and internal representations. When the data consists of natural images, some of these interpretations rely on the human visual system such that interpretations are visualizations of the model's internal representations of the data (\eg saliency maps\cite{lundberg2017unified,selvaraju2017grad} or component visualizations\cite{olah2020zoom,qin2018convolutional}). In general, such interpretability approaches are limited to certain types of data and to the qualitative interpretation by domain experts. 
In this paper, we propose an approach that includes interpretability as an integral part of the model. 

Specifically, our model consists of maps $f_c: X\rightarrow C$ and $f_y: C \rightarrow Y$, where $C$ indicates the domain of higher-level human-understandable \textit{concepts}. The first map $f_c$ develops an encoding of each datapoint into the domain $C$ that we then use to explain the model's decisions. The second map $f_y$ implements the downstream task of assigning a class value to the datapoint. 

The explanations that we produce are \textit{contrastive}. Contrastive explanation follows the human tendency of explaining an event by (implicitly) comparing it to some alternative event that did not take place\cite{lipton1990contrastive,miller2019explanation}. In our case, they convey why a datapoint belongs to a given class in contrast to some other class, by highlighting the differences as a \textit{sequence of transformations} of the datapoint.

For the empirical evaluation in this study we used image data. This allows us to present the explanations in a \textit{visual} form; an example is depicted in Fig.~\ref{fig:example}. In general, the method could produce contrastive explanations by a sequence of transformations in any data space.%

To create suitable explanations we need to be able to represent the data in an interpretable concept space and be able to generate interpretable transformations against a contrastive target. For these purposes we use a generative latent variable model, specifically a disentangled Variational Autoencoder (VAE)\cite{kingma2013auto,rezende2014stochastic}. We employ existing methods that disentangle class-relevant from irrelevant information\cite{cai2019learning,ilse2020diva,zheng2019disentangling}, as only the former is of interest \wrt domain $C$. We expand this model further with a new method for representing individual concepts in individual dimensions, focused on generating high-quality transitions when changing a single dimension. To develop the contrastive explanation, we define a target datapoint that is associated with the target class, referred to as the exemplar\cite{nosofsky1992}. We then infer a sequence of transformations in the concept domain $C$ that interpolates between the query datapoint and the exemplar. The exemplar is selected such that it is representative of its class, and such that the sequence of transformations is of minimum length.

We denote our approach as Variational Autoencoder-based Contrastive Explanation (VAE-CE). To be able to validate VAE-CE, we define a method for quantitatively evaluating explanations, resting on access to the true generating process of the data. Using synthetic and real data we quantitatively and qualitatively compare our method to similar methods and evaluate the individual components of our method. The two main contributions of this paper can be summarized as follows:
\begin{itemize}
\item We propose a method for disentangling latent dimensions in a VAE. This method is guided by pairs of images indicating changes that should (not) correspond to a change in a single dimension. (\S\ref{ss:m:pair})
\item We propose a method for generating visual contrastive explanations of a datapoints' class assignment. This method considers conditioning a VAE to represent class concepts in individual dimensions in a subpart of the latent space, and uses this space to generate interpolations depicting the class-relevant concepts. (\S\ref{s:m})
\end{itemize}

\section{Related work}
\begin{figure*}[!htb]
\centering
    \includegraphics[width=0.9\textwidth]{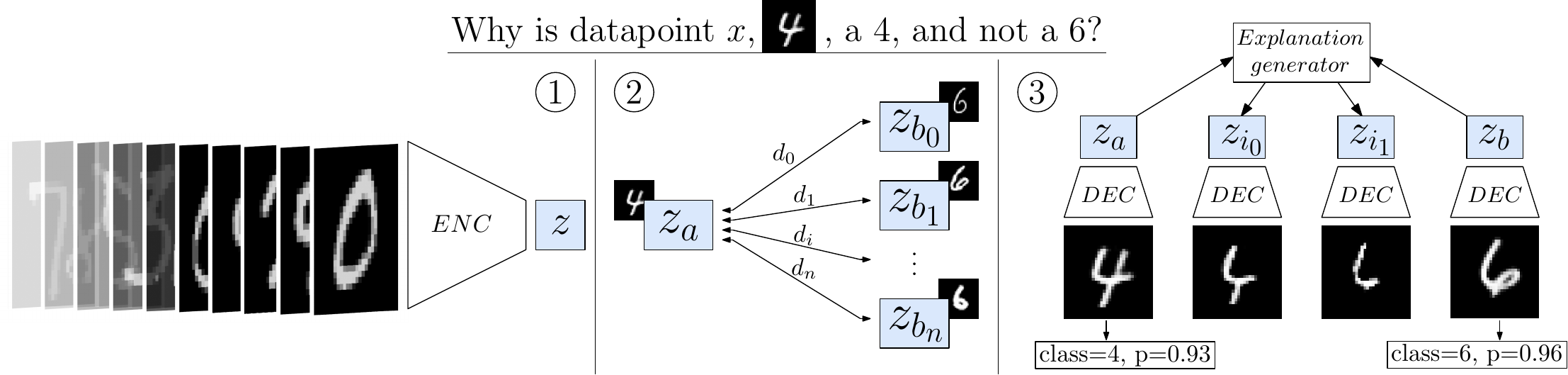}
    \caption{A simplified overview of the explanation generation method (we omit non-class factors for simplicity). (1) Encode a dataset to a semantically meaningful latent space. (2) In this space, find a suitable exemplar to contrast our query datapoint with. (3) Generate an interpolation highlighting the relevant concepts that distinguish the query from the exemplar.}
    \label{fig:mo}
\end{figure*} %
Literature most relating to our work considers the topics of explanation in the context of image classification using neural networks, and disentanglement in VAEs.

\textbf{Image-classification explanations} come in many shapes. Saliency-based methods\cite{lundberg2017unified,selvaraju2017grad,zeiler2014visualizing} explain classifications by highlighting the contribution of pixels \wrt the decision. This approach is extended to project contributions \wrt an alternative class, creating contrastive explanations\cite{pazzani2018explaining,prabhushankar2020contrastive}. To evaluate the pixel contributions in a black-box manner, one can perturb images rather than inspect model components, \eg  for class detections\cite{ribeiro2016should} or for contrasting classes\cite{dhurandhar2018explanations}. Models with an attention mechanism\cite{dosovitskiy2020image,xiao2015application} fundamentally possess the mechanic of inspecting input contributions, although attention likely only noisily predicts the input features' importance\cite{serrano2019attention}.

One can generate examples showing changes necessary to flip a class decision, denoted as counterfactuals, to explain differences in classification decisions\cite{van2019interpretable}.
Another method for conveying such differences considers highlighting regions within the query image and an alternative image, showing which pieces must be swapped to flip the decision\cite{goyal2019counterfactual}. The use of deep generative models, such as Generative Adversarial Networks (GANs)\cite{goodfellow2014generative} and VAEs\cite{kingma2013auto,rezende2014stochastic}, has been proposed to explain classification boundaries using a high-level space. Such methods, \eg \cite{feghahati2018cdeepex,joshi2019towards,liu2019generative,o2020generative,samangouei2018explaingan,singla2020explanation}, involve different approaches to training a generative model and interpolating in its latent space to convey the boundary between different classification targets.
Alternatively, one can create translations conveying these boundaries using datapoints\cite{poyiadzi2020face}.

To interpret decisions one can also work with high-level concepts, \eg by evaluating learned components\cite{olah2020zoom,qin2018convolutional} or by identifying associations between prespecified concepts and classification decisions\cite{kim2018interpretability}. Concepts can also be an integral part of a model, \eg by first detecting concepts and using them to classify in an interpretable fashion\cite{alvarez2018towards,zhou2018interpretable}. Other examples of self-explaining methods consider generating text descriptions that provide a (contrastive) reason for the classification\cite{hendricks2018grounding} and matching image parts to other samples in order to assign a class\cite{chen2019looks}.

\textbf{Disentanglement in VAEs} can be defined as the notion that single latent dimensions are sensitive to changes in single generative factors while being invariant to changes in others\cite{bengio2013representation,higgins2016beta,locatello2019challenging}. It is also often used in the context of separating information related to some factor (\eg a class label or a specific grouping) from unrelated information\cite{bouchacourt2017multi,cai2019learning,zheng2019disentangling}.

Unsupervised approaches generally extend the \textit{ELBO}'s regularization term with extra assumptions about the latent space, \eg \cite{chen2018isolating,higgins2016beta,kim2018disentangling}.
Sharing similarities with our approach to disentanglement (\S\ref{ss:m:pair}), \cite{zhu2020learning} propose learning a disentangled representation using pairs differing in a single dimension, maximizing mutual information. Regarding unsupervised disentanglement, \cite{locatello2019challenging} raised the question whether we can expect well-disentangled representations, showing that strong inductive biases or supervision are a necessity for learning and validating such representations.

Incorporating supervision, one approach is to group datapoints according to some shared feature and optimize a subpart of the latent space to share a representation for this group\cite{bouchacourt2017multi,hosoya2019group,jha2018disentangling}. A weakly-supervised variant of this idea considers heuristically finding which dimensions are common and sharing those\cite{locatello2020weak}. Alternatively, one can use labels to encourage a disentangled latent space, \eg by optimizing subspaces to contain or exclude information about a label using auxiliary classification objectives\cite{cai2019learning,ding2020guided,ilse2020diva,zheng2019disentangling}.

\section{Method: VAE-CE}\label{s:m}

\subsection{Learning a data representation for explanation}\label{ss:m:base} %
To represent the data in a higher-level space, we use a VAE\cite{kingma2013auto,rezende2014stochastic}. A VAE aims to approximate a dataset's distribution under the assumption that its samples $x$ are generated according to some latent variable $z$. In other words, the aim is to model $p(x, z) = p(x|z)p(z)$. This relation is approximated using an encoder $q_\phi(z|x)$ and decoder $p_\theta(x|z)$ distribution, parameterized by deep neural networks, and optimized using a lower bound on the true likelihood of the data, the \textit{ELBO}. The reparametrization trick\cite{kingma2013auto} is used to (back)propagate through %
the latent variables.

Using a VAE we can both infer latent variables $z$ given data $x$, and generate modified samples $\widetilde{x}$ given some modification in $z$. It provides us with the tools to work in concept domain $C$, for both classification and explanation purposes.
However, not all information in $x$, and consequently in $z$, is necessarily class related. To overcome this issue we build upon work aimed at disentangling class-relevant from irrelevant information in a VAEs latent representation.

The VAE's \textit{ELBO} objective is extended with classification terms, 
in line with works such as \cite{cai2019learning,ding2020guided,ilse2020diva,zheng2019disentangling}. Latent variable $z$ is split into subspaces $z_y$ and $z_x$, where the former aims to contain class-relevant information and the latter should contain the remaining information. We use a separate encoder for inferring each latent subspace; the $z_y$ encoder, $q_{{\phi{_y}}}(z_y|x)$, serves as the concept encoder, $f_c$.

We introduce categorical distributions $q_{{\psi{_y}}}(y|z_y)$ and $q_{{\psi{_x}}}(y|z_x)$, parameterized by neural networks and optimized using their log-likelihoods. We refer to these as the latent spaces' classifiers. The former, $q_{{\psi{_y}}}(y|z_y)$, is also used to infer class predictions, serving as $f_y$. 

For training, we simultaneously optimize the parameters of both classifiers and both encoders using categorical cross-entropy. However, $z_x$ should contain little information about label $y$. To learn such a label-agnostic subspace we reverse the loss' gradients for $z_x$'s encoder, $q_{{\phi{_x}}}(z_x|x)$, through a Gradient Reversal Layer\cite{ganin2015unsupervised}.

\begin{figure}[!htb]
    \centering
    \includegraphics[width=0.35\textwidth]{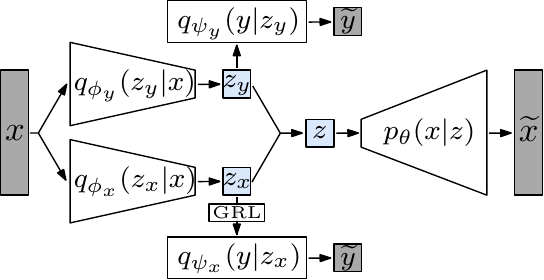}
    \caption{The architecture of the disentangled VAE. Datapoint $x$ is encoded by two separate encoders into $z_x$ and $z_y$, which are concatenated to reconstruct $\widetilde{x}$. Disentanglement is encouraged by auxiliary classifiers. We omit the sampling procedure of the latent variables for clarity.}
    \label{fig:vaedis}
\end{figure}

For each loss term, the subscript denotes the parameters it optimizes. The loss terms are as follows:
\begin{alignat}{1}
\loss_{\theta,{\phi{_y}},{\phi{_x}},{\psi{_y}}}&(x, y) = \beta_y KL(q_{{\phi{_y}}}(z_y|x)||p_\theta(z))\label{eq:kl0}\\
&+\beta_x KL(q_{{\phi{_x}}}(z_x|x)||p_\theta(z))\label{eq:kl1}\\
&-\mathbb{E}_{q_{{\phi{_y}}}(z_y|x),q_{{\phi{_x}}}(z_x|x)}[\log p_\theta(x|z_y,z_x)]\label{eq:rec}\\
&-\alpha \mathbb{E}_{q_{\phi{_y}}(z_y|x)}[\log(q_{{\psi{_y}}}(y|z_y))]\label{eq:cl0}\\
&+\alpha \mathbb{E}_{q_{\phi{_x}}(z_x|x)}[\log(q_{{\psi{_x}}}(y|z_x))],\label{eq:cl1}\\
&\hspace{-.576cm}\loss_{\psi{_x}}(x, y) = -\mathbb{E}_{q_{\phi{_x}}(z_x|x)}[\log(q_{{\psi{_x}}}(y|z_x))],\label{eq:cl1_adv}
\end{alignat}
with hyperparameters $\beta_y$, $\beta_x$ and $\alpha$. We approximate all expectations with single-sample Monte Carlo estimation. Prior distribution $p_\theta(z)$ is set to a standard factorized Gaussian, $\mathcal{N}{(0, I)}$, which allows us to compute (\ref{eq:kl0}) and (\ref{eq:kl1}) analytically\cite{kingma2013auto}. Distribution $p_\theta(x|z_y, z_x)$ is assumed to be a factorized Gaussian with fixed variance, allowing us to approximate (\ref{eq:rec}) by taking the squared error between the input and its reconstruction. (\ref{eq:cl0}), (\ref{eq:cl1}) and (\ref{eq:cl1_adv}) optimize the log-likelihood of the categorical distributions and are computed using categorical cross-entropy. Note that (\ref{eq:cl1}) is a negation of (\ref{eq:cl1_adv}): Both are computed in a single pass. An overview of the model is depicted in Fig. \ref{fig:vaedis}.

\subsection{Pair-based dimension conditioning}\label{ss:m:pair}
To produce explanations that convey differences in class concepts, we must manipulate concepts individually. To exercise such control, we aim to learn a representation where individual $z_y$-dimensions control individual concepts. We introduce a new disentanglement method based on two assumptions: (1) a significant change in a single latent dimension should correspond to changing a single concept and (2) we can train a model to evaluate whether changes fit this criterion. This method acts as additional regularization and is added on top of the previously described objective.

Two auxiliary models are used to aid the regularization procedure: A `Change Discriminator' ($CD$) and a regular `Discriminator' ($D$), both predicting a value in the range [0, 1]. $CD$ is trained beforehand, and infers whether a pair of datapoints exhibits a desirable change. In our implementation, we train $CD$ as a binary classifier with pairs that either indicate a good change (a single concept change) or a bad change (no or multiple concept changes); for details we refer to the supplementary material. %
$D$ is trained to distinguish between generated and real datapoints, as done in a GAN\cite{goodfellow2014generative}.

\begin{figure}[!b]
    \centering
    \includegraphics[width=\linewidth]{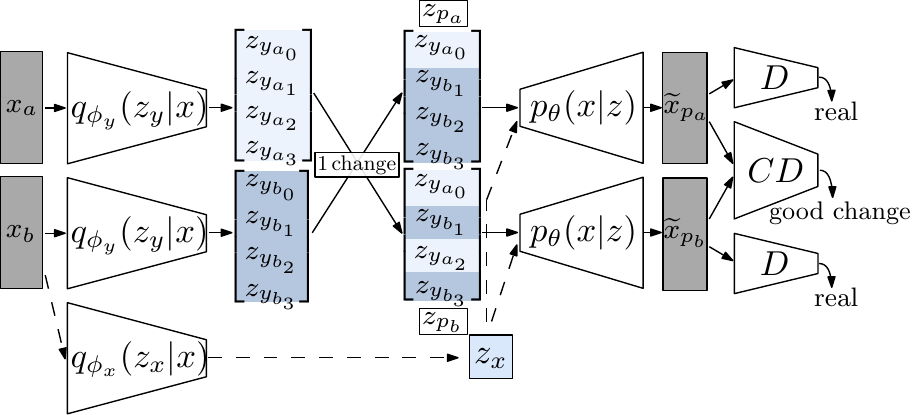}
    \caption{Individual dimensions are disentangled in an amortized fashion: Randomly constructed latent spaces differing in a single dimension are optimized to exhibit a desirable change in data space.}
    \label{fig:cpair}
\end{figure}

By optimizing latent-dimension changes using $CD$ as a critic, individual dimensions should better represent single concepts. $D$ is used to optimize the quality of the samples to avoid a degenerate solution where non-realistic changes are produced that merely trick $CD$, rather than representing meaningful concept changes (\ie an adversarial attack\cite{szegedy2013intriguing}). 

A visualization of the regularization procedure is depicted in Fig.~\ref{fig:cpair}. One step works as follows:
\begin{enumerate}
    \item Encode two arbitrary (non-identical) datapoints $x_a$ and $x_b$ to their latent representations in $z_y$-space, giving us $z_{y_a}$ and $z_{y_b}$. For the remaining information only encode the representation of datapoint $x_b$ to $z_{x}$. %
    \item Construct two latent variables that share all but one dimension by combining $z_{y_a}$ and $z_{y_b}$ stochastically. We denote these variables as $z_{p_a}$ and $z_{p_b}$. Each individual dimension comes from either $z_{y_a}$ or $z_{y_b}$ (equally likely), and all but one dimension are shared.
    \item Map the constructed pair back to data space. That is, synthesize $\widetilde{x}_{p_a}$ and $\widetilde{x}_{p_b}$ by decoding latent representations $(z_{p_a},z_x)$ and $(z_{p_b},z_x)$.
    \item Optimize the encoders and the decoder such that $CD$ predicts a high-quality change between $\widetilde{x}_{p_a}$ and $\widetilde{x}_{p_b}$ and $D$ predicts that the samples are real.
\end{enumerate}
The corresponding loss term is as follows:
\begin{alignat}{1}
\hspace{0.1cm}\loss_{\theta,{\phi{_y}},{\phi{_x}}}(\widetilde{x}_{p_a}, \widetilde{x}_{p_b}) &=  {-}\alpha_r\log(D(\widetilde{x}_{p_a}))\\ &\hspace{0.465cm}{-}\alpha_r\log(D(\widetilde{x}_{p_b}))\\
&\hspace{-1.5cm}+ \alpha_{p} n_y\frac{|z_{p_a}-z_{p_b}|}{|z_{y_a}-z_{y_b}|} \cdot - \log(CD(\widetilde{x}_{p_a}, \widetilde{x}_{p_b})),
\end{alignat}
with hyperparameters $\alpha_r$ and $\alpha_p$, and $n_y$ denoting the number of dimensions in $z_y$. This term optimizes the VAE such that $CD$ and $D$ predict high-quality changes and realistic datapoints. We scale the loss of $CD$'s prediction according to the difference in the dimension compared to the overall difference, multiplied by the number of dimensions. This extra scalar term ensures that we do not penalize `bad' changes when the differing dimension is insignificant.

Discriminator $D$ is trained in the same manner as a GAN's discriminator, using $\widetilde{x}_{p_a}$ and $\widetilde{x}_{p_b}$ as fake data alongside real data from the training set; it learns to distinguish between them by minimizing the binary cross-entropy between the predicted labels and true/false labels.

\subsection{Explanation generation}\label{ss:m:gen}
To explain a datapoint we focus on two aspects: Identifying a suitable exemplar and producing an explanation that displays the class concepts that differ between the datapoint and this exemplar. The exemplar is chosen from an alternative class, \eg the second most likely class (given $q_{{\psi{_y}}}(y|z_y)$) or user selected. Alternatively, one could select a specific datapoint. An overview of the explanation procedure is provided in Fig.~\ref{fig:mo}. When creating explanations we use mean values, rather than samples, of latent variable $z$. As such, we substitute $z$ for $\mu$ in this subsection.

\textbf{Exemplar identification} rests on two principles: (1) how representative a datapoint is of its class and (2) how similar it is to the datapoint we contrast it with (as more similarity implies fewer concepts to change). To capture the former we only consider datapoints whose class probability is above a given threshold: $q_{{\psi{_y}}}(y_i|\mu_y) > t$. For the latter, we select the datapoint with the minimum squared difference in the class-specific subspace: $\min\limits_{b} \ (\mu_{y_a} - \mu_{y_b})^2$, with $a$ indicating the query datapoint and $b$ the exemplar.%

\textbf{Explanation generation} works by transforming the class-relevant latent embedding from the query (${\mu_{y_a}}$) to the exemplar (${\mu_{y_b}}$) and showcasing the intermediate steps; the class-irrelevant embedding (${\mu_{x_a}}$) is left unchanged. Dimension values are changed at once, as dimensions represent individual concepts. For each interpolation step, we allow multiple such dimension values to be switched, as there is no guarantee that every dimension difference depicts a concept changing (\ie small differences are likely---but not necessarily---meaningless). We consider all orders of changing (groups of) dimensions; as dimensions can still be entangled, the interpolation path can have a significant effect on the quality of the intermediate states\cite{chen2019homomorphic,yan2020semantics}.

The path we take to interpolate from ${\mu_{y_a}}$ to ${\mu_{y_b}}$ should be of minimum length, in line with the Minimum Description Length (MDL)\cite{grunwald2007minimum} principle. Additionally, it is optimized \wrt two aspects: (1) each step should depict a single concept change and (2) each state should represent the dataset's underlying distribution. These properties are optimized using auxiliary models $CD$ and $D$.%

\begin{figure}[!htb]
    \centering
    \includegraphics[width=0.24\textwidth]{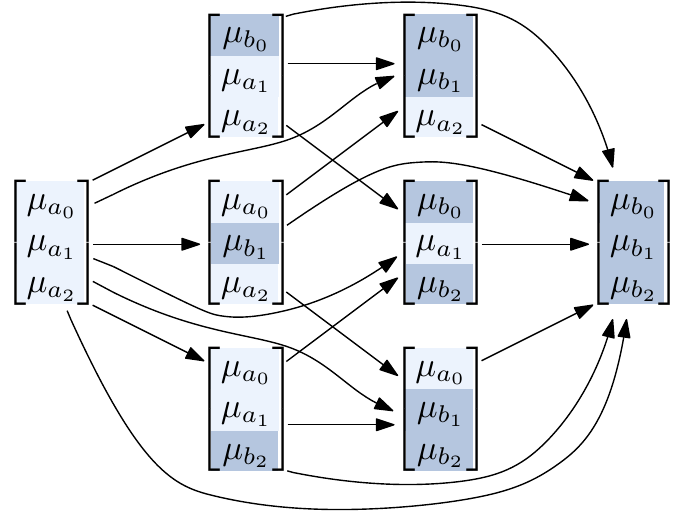}
    \caption{The interpolation graph of the transition between two latent variables of size 3 (weights omitted for clarity).}
    \label{fig:graph}
\end{figure}

Not all interpolation paths are explicitly computed, as the quantity of paths changing (groups of) dimensions grows extremely fast\footnote{Equivalent to the number of weak orderings of a set: Given $n$ latent dimensions, the $n^{th}$ Ordered Bell number\cite{mezo2019combinatorics}.}. Rather, we build a graph denoting all paths, where each edge denotes the cost of adding this state to the interpolation: A weighted sum of the probabilities of the change being undesirable ($CD$) and the datapoint being fake ($D$), adjusted by a normalization coefficient. For the change from $\mu_i$ to $\mu_j$ this can be computed as follows:
\begin{alignat}{1}
w_{ij} = {[\alpha \big(1-D(\widetilde{x}_j)\big) + \beta \big(1 - CD(\widetilde{x}_i, \widetilde{x}_j)\big)]} \cdot {k^\gamma},
\end{alignat}
where $\widetilde{x}_i$ and $\widetilde{x}_j$ are the reconstructed datapoints of states $i$ and $j$, $k$ is the number of dimensions changed, and $\alpha$, $\beta$, and $\gamma$ are hyperparameters. The shortest path in this graph represents the interpolation path optimized for our desiderata. An example of an interpolation graph is depicted in Fig.~\ref{fig:graph}.

While the shortest path can be found in linear time \wrt the nodes and edges (since the graph is directed and acyclic\cite{cormen2009algo}), the graph itself grows quickly. For $n$ dimensions to change there are $2^n$ nodes and $3^n-2^n$ edges (we refer to the supplementary material for a derivation). As such, this approach is only applicable to problems with a limited number of dimensions.
\section{Experimental setup}

\subsection{Datasets}\label{ss:s:data}
\begin{figure}[!b]
    \centering
    \subfloat[The underlying concepts determining a datapoints' class.]{{\label{fig:synb}
    \adjincludegraphics[trim={0 {0.05\height} 0 {0.04\height}},clip,width=.85\linewidth]{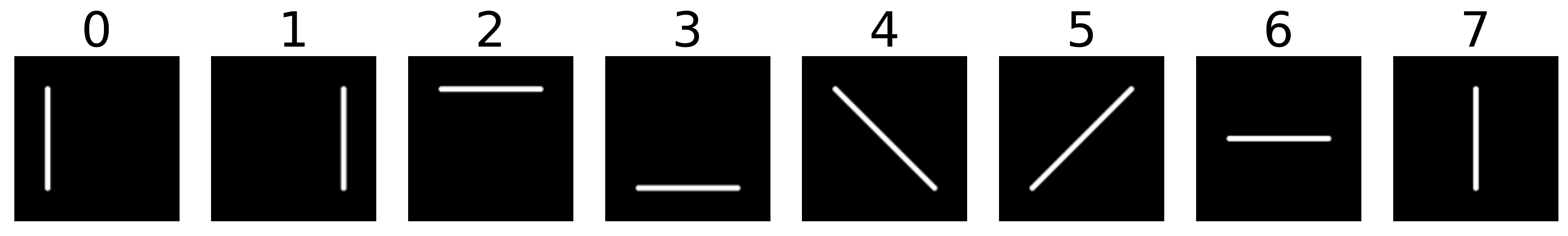}
    }} \\
    \vspace{-.1cm} %
    \subfloat[The ten classes in the dataset. The value above depicts the class index, whereas the value below depicts the indices of the lines that determine it.]{\label{fig:sync}
    \raisebox{.1cm}[2cm][0cm]{
    {\adjincludegraphics[trim={0 {0.022\height} 0 {0.025\height}},clip,width=1\linewidth]{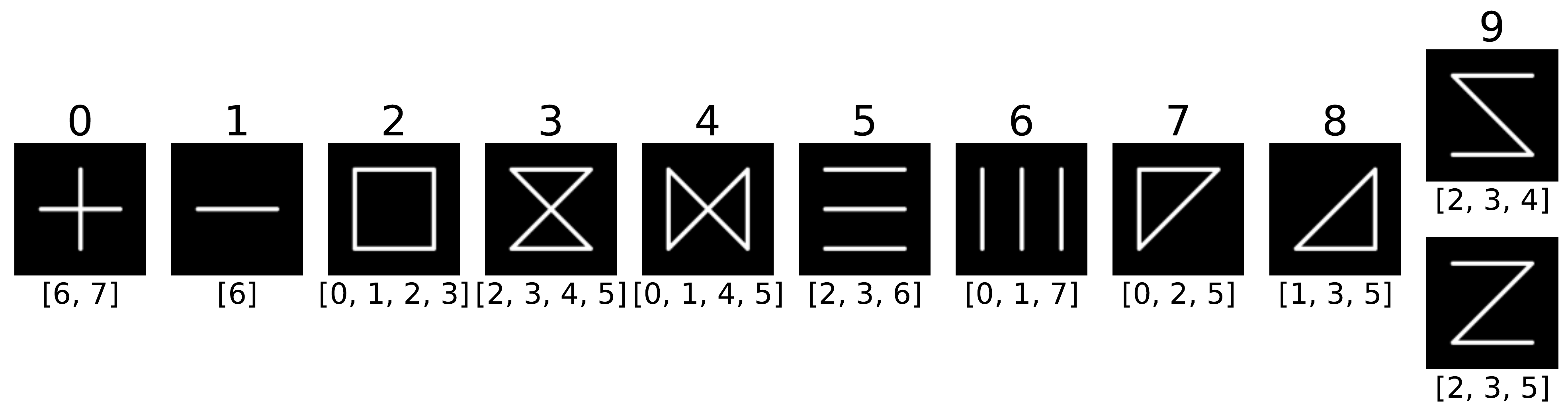} }}}
    \caption{An overview of the synthetic data's structure.}%
    \label{fig:synfactor}%
\end{figure}
\textbf{Synthetic data} with a known generating process and set of concepts is used to validate our method in a controlled setting. The class determines the datapoints' concepts, which together with added noise determine the datapoint. Concepts are defined as the occurrence of lines, where each line is defined by its orientation, length, and relative position. We use eight variables determining whether a specific line occurs in the data. %
The dataset consists of ten classes, with each class consisting of some combination(s) of lines. %
These lines and classes are depicted in Fig.~\ref{fig:synfactor}.

\begin{figure}
    \centering
    \subfloat[10 synthetic datapoints.]{\label{fig:synd}
    \raisebox{.1cm}{
    \adjincludegraphics[width=.4\linewidth]{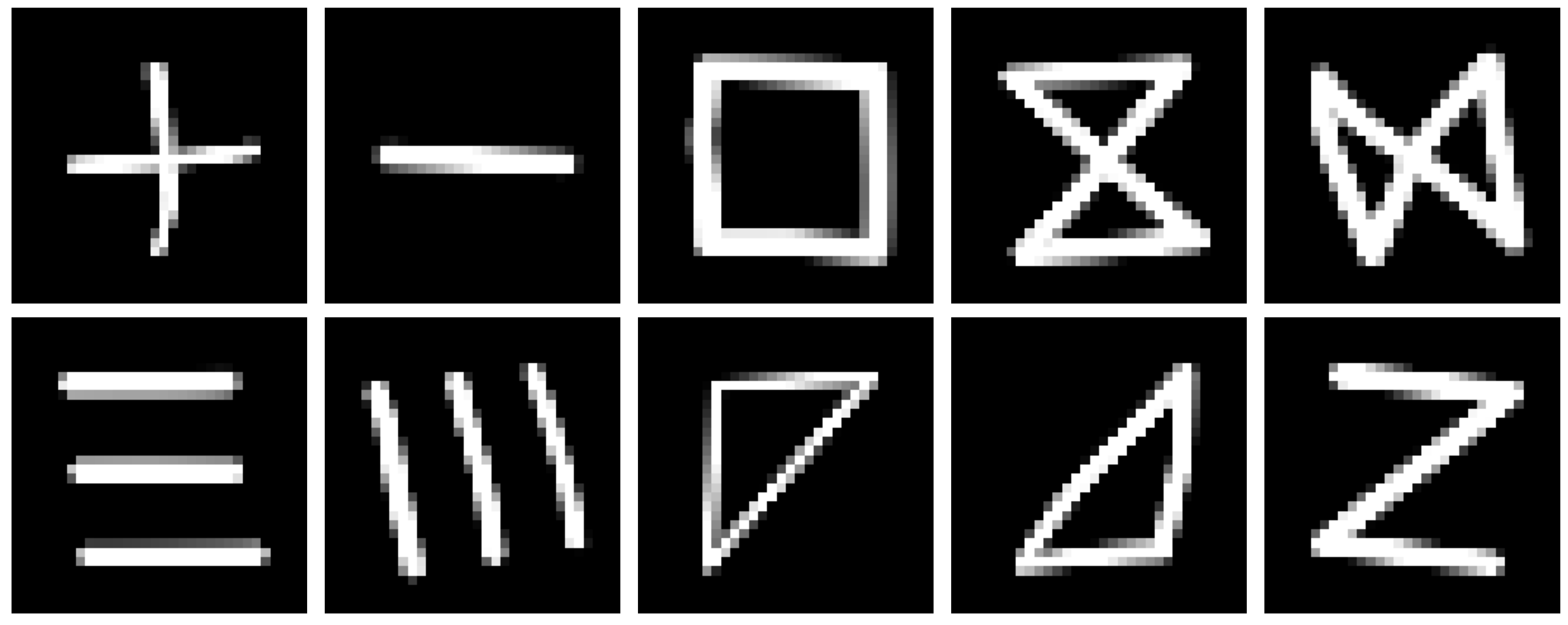}
    }
    }
    \hspace{.05\linewidth}
    \subfloat[Synthetic change pairs.]{\label{fig:syncp}
    \raisebox{.57cm}[0cm][0cm]{
    \makebox[.4\linewidth][c]{
    \begin{tabular}{l|l}
    \adjincludegraphics[width=.15\linewidth,trim={0 {.33\height} 0 0},clip]{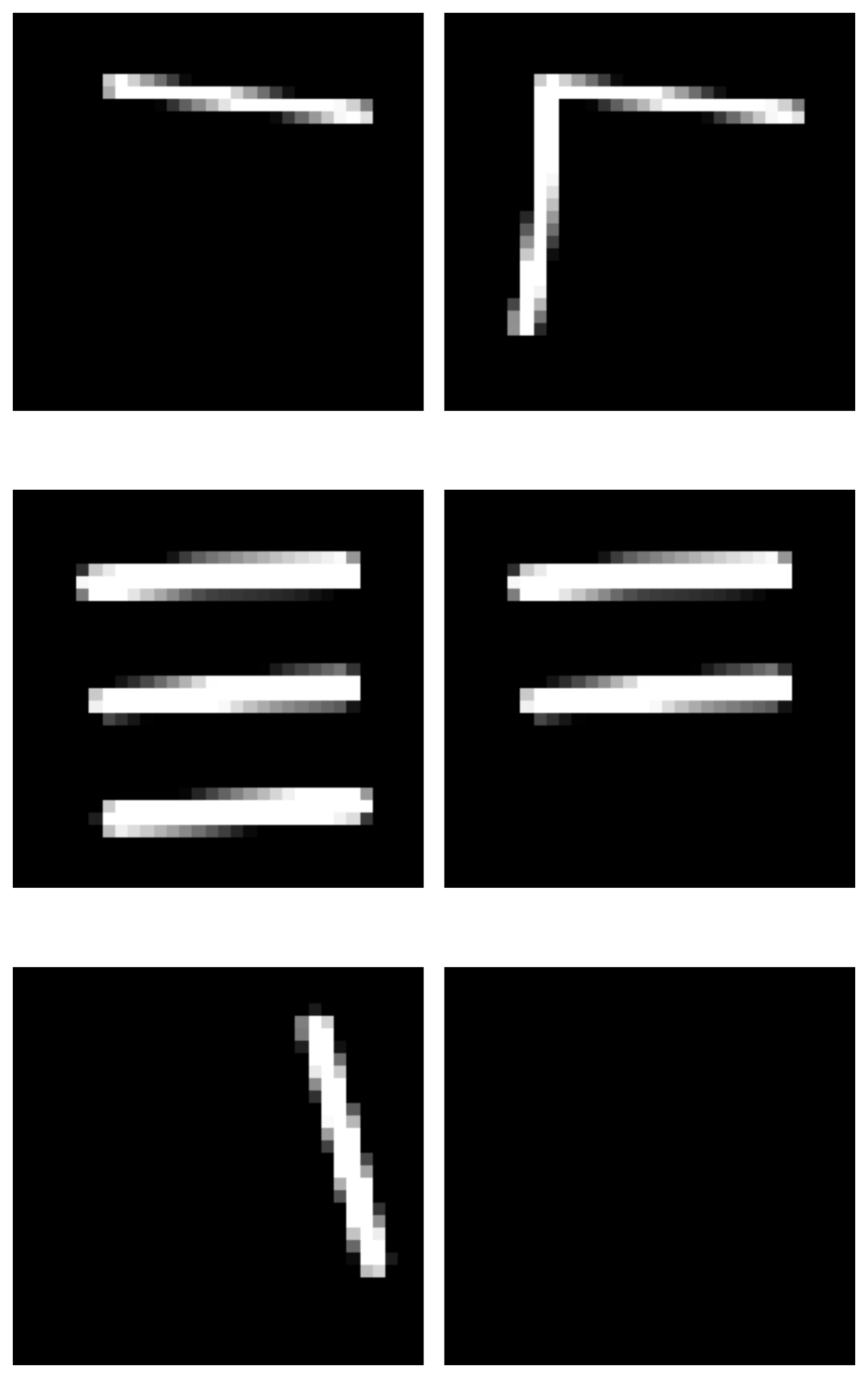} &  \adjincludegraphics[width=.15\linewidth,trim={0 0 0 {.33\height}},clip]{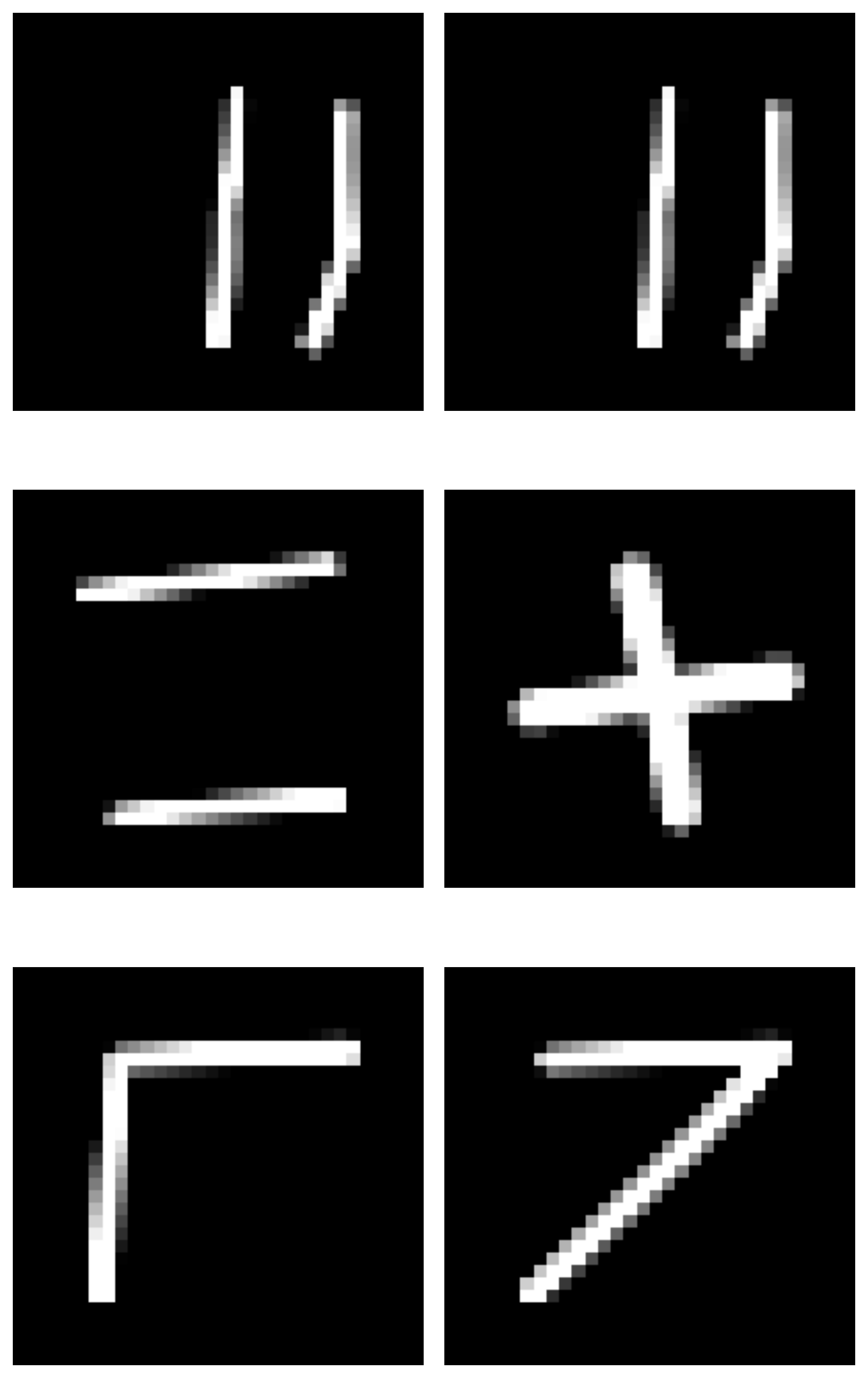}
    \end{tabular}
    }
    }
    }
    \\
    \vspace{-.15cm} %
    \subfloat[10 samples from MNIST.]{\label{fig:mnistd}
    \raisebox{.1cm}{
    \adjincludegraphics[width=.4\linewidth]{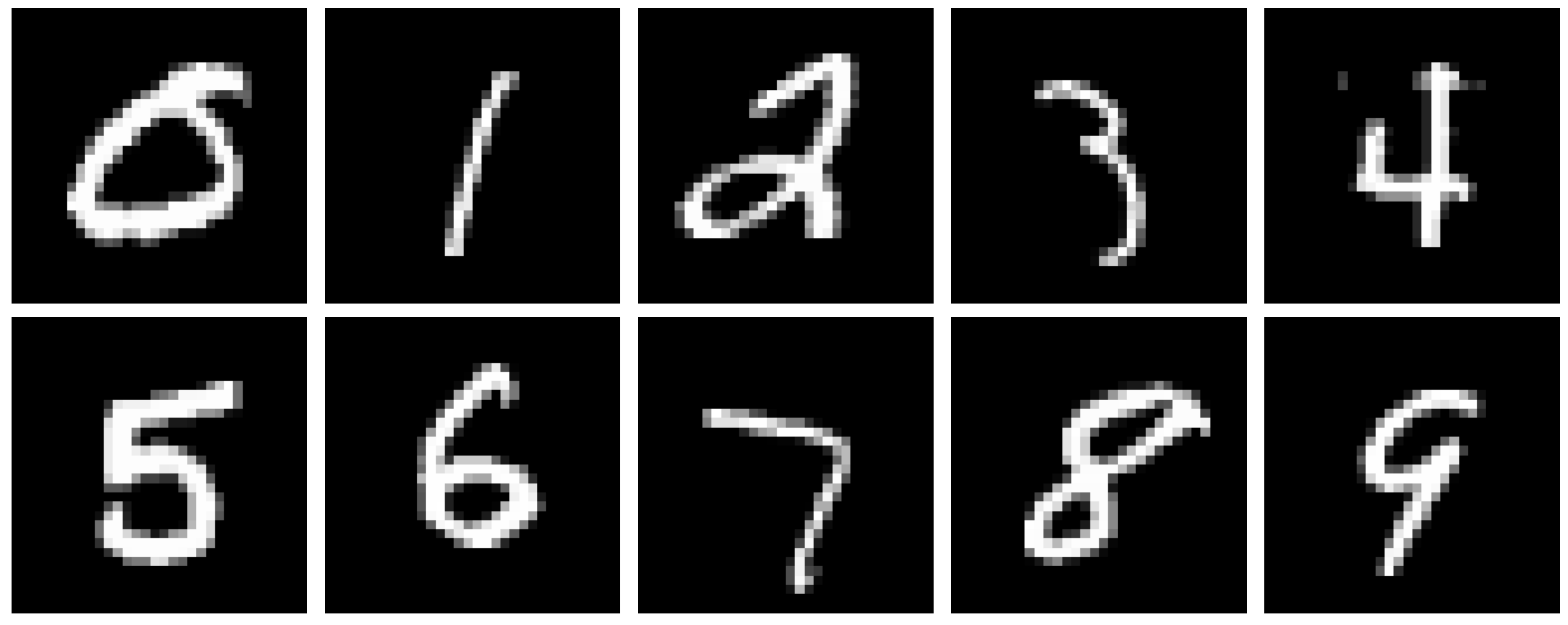}
    }
    }
    \hspace{.05\linewidth}
    \subfloat[MNIST change pairs.]{\label{fig:mnistcp}
    \raisebox{.57cm}[0cm][0cm]{
    \makebox[.4\linewidth][c]{
    \begin{tabular}{l|l}
    \adjincludegraphics[width=.15\linewidth,trim={0 {.33\height} 0 0},clip]{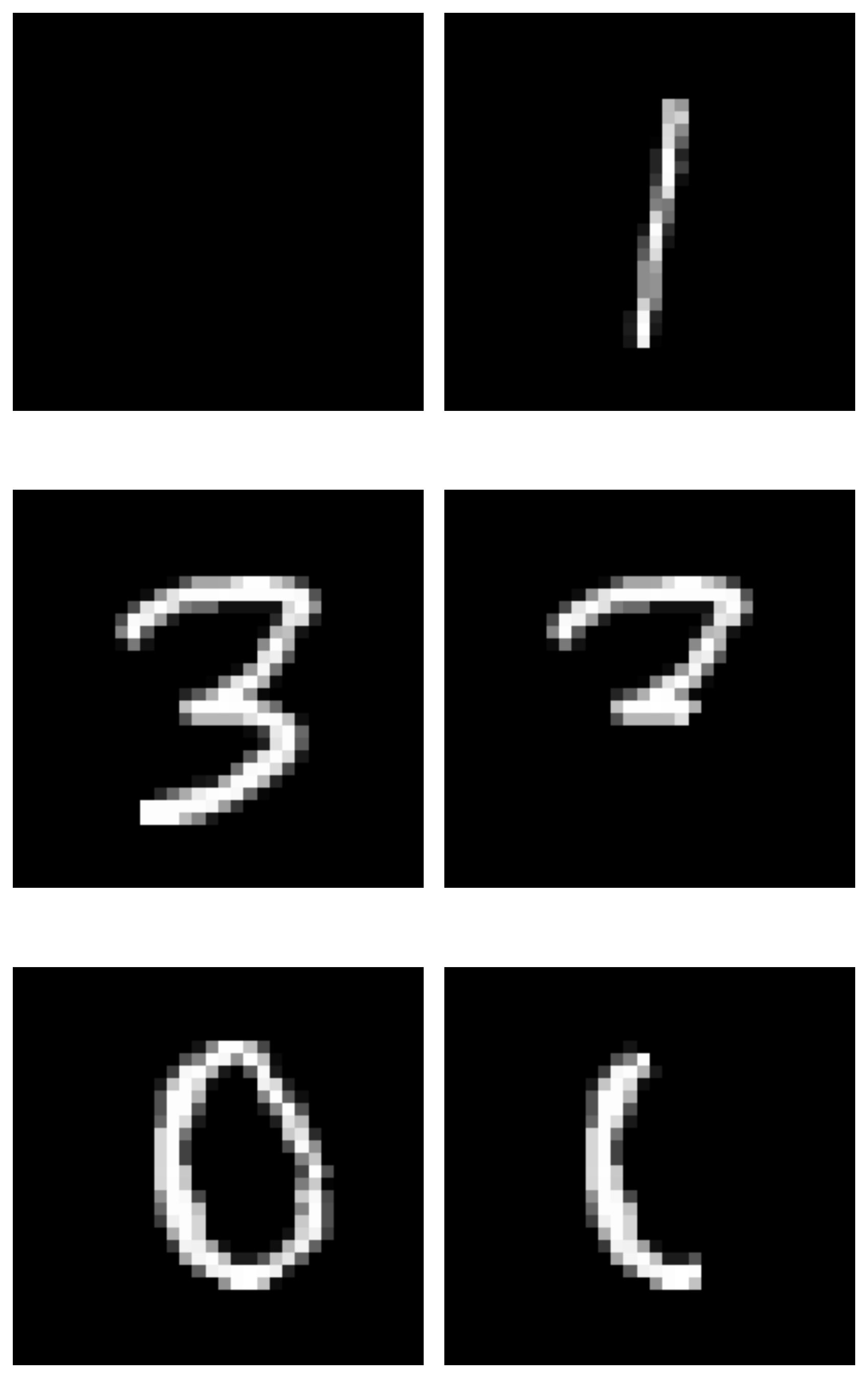} &  \adjincludegraphics[width=.15\linewidth,trim={0 0 0 {.33\height}},clip]{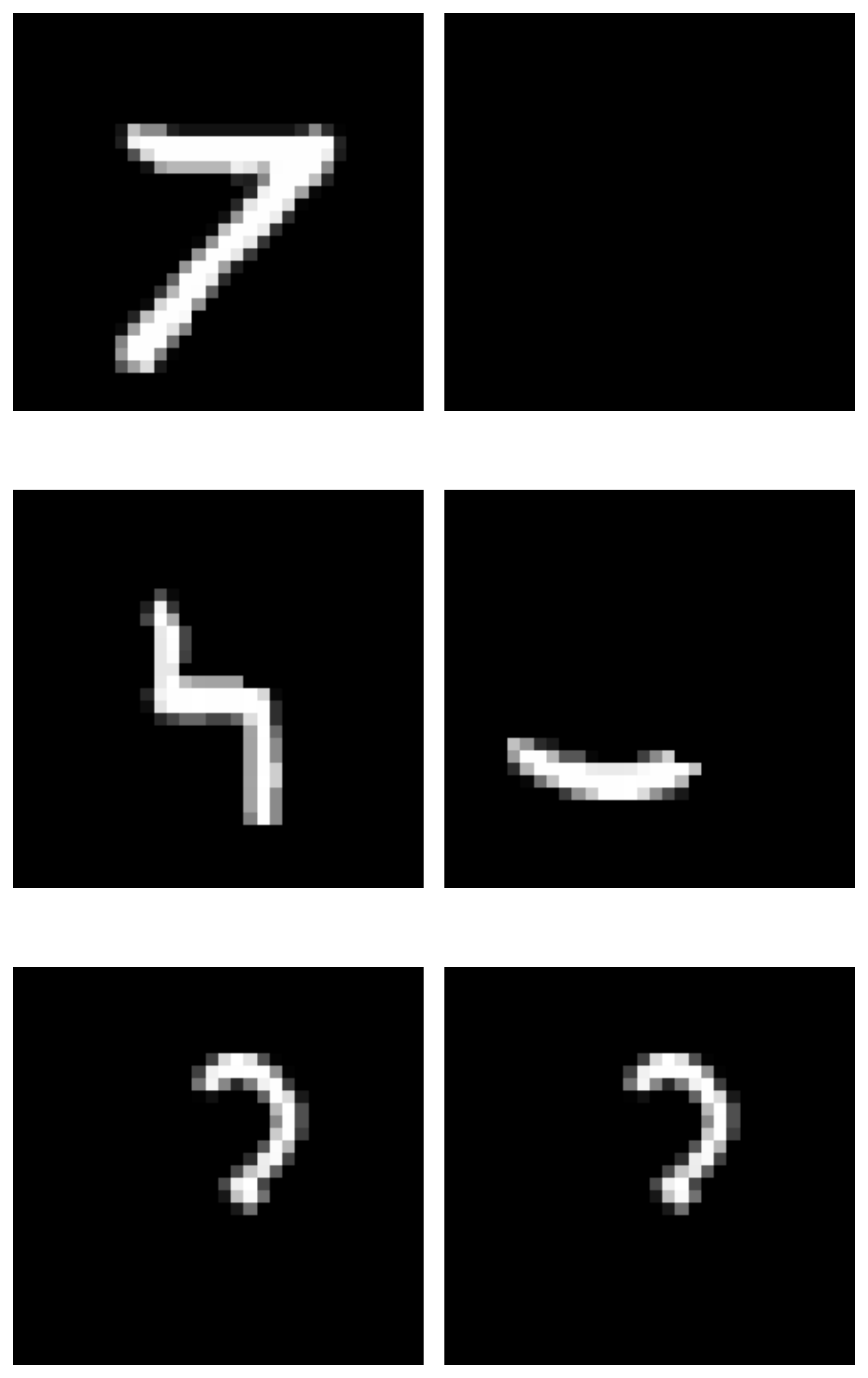}
    \end{tabular}
    }
    }
    }
    \caption{Synthetic data and MNIST, as used for training. Change pairs depicted on the left are positive (1 change), whereas those on the right are negative (0/2+ changes).}%
    \label{fig:synex}%
\end{figure}

 Datapoints are generated by taking these `base shapes' and adding non-trivial noise. The noise process seeks to mimic that of handwritten shapes (such as MNIST digits) and consists of shape distortion and line-width variation. We refer to the supplementary material for a detailed description of this generation procedure. Examples of synthetic datapoints are depicted in Fig.~\ref{fig:synd}.
 
 The training and test set consist of \num{10000} and \num{1000} $32\times32$-pixel images for each class, respectively. Model selection is done according to an explanation-quality metric that samples directly from the generative process (see \S\ref{ss:s:eval}), no validation set is used for tuning the model. Change pairs (for $CD$) are created by taking a class configuration and hiding some line(s) in both images in the pair, such that only 1 (positive) or 0/2+ lines differ (negative). Examples of such pairs are depicted in Fig~\ref{fig:syncp}. Supervision used by other methods can be created using knowledge of the generative process. For each type of supervision we generate the same number of samples in total, \num{100000}.

\textbf{MNIST}\cite{lecun1998gradient} is used to evaluate our method in a more realistic setting, \ie with noisy supervision.
For ease of implementation, all images are padded to $32\times32$ pixels. 
No ground-truth concepts are available for MNIST. Consequently, we can only evaluate methods for which we can approximate the required supervision, and cannot evaluate metrics requiring ground-truth concept labels. %

To create change pairs, images are augmented according to the notion that the concepts we reason with are continuous lines. Digits are reduced to individual lines and pixels are clustered according to these lines (we refer to the supplementary material for details). Using this line split, pairs are created that exhibit 1 (positive) or 0/2+ (negative) line changes. We create as many augmented pairs as there are training datapoints: \num{60000}. Examples of MNIST datapoints and change pairs are depicted in Figs.~\ref{fig:mnistd} and~\ref{fig:mnistcp}. Creating a labeling of line types is a significantly more challenging task than augmenting individual images to create change pairs. As such, we do not consider methods requiring such supervision when evaluating MNIST.

\subsection{Considered evaluations}\label{ss:s:eval}

\textbf{Explanation alignment cost (\textit{eac})}. To the best of our knowledge there is no method for quantitatively evaluating explanations of our defined structure. As such, we introduce the explanation alignment cost ($eac$). The $eac$ seeks to quantify the quality of a contrastive explanation based on a pair of datapoints $a$ and $b$ as input. The explanation consists of an interpolation starting at datapoint $a$, gradually transitioning to the \textit{class-relevant} concepts of $b$ (\ie the final state of the transition is not necessarily identical to $b$). Each step should indicate a single concept being changed. %

A candidate explanation for ($a, b$) is evaluated according to the cost of aligning it to a ground-truth explanation. We define a ground-truth explanation as a minimum length sequence starting at $a$, with each subsequent state changing a single concept from $a$ to $b$,
with no other changes. The last state depicts a datapoint with all class-relevant concepts from $b$ and the remaining information from $a$.

The alignments we identify must map every state in the candidate explanation to at least one state in the ground-truth explanation, and vice versa. Additionally, we constrain this mapping such that both aligned sequences are increasing. Such an alignment can be computed using Dynamic Time Warping (DTW)\cite{senin2008dynamic} in $O(nm)$ time (with $n$ and $m$ denoting the length of the explanations). We compute the cost of each individual state-to-state mapping as the per-pixel squared error and a constant, for discouraging (empty) repetitions in the alignment: $(x_c - x_t)^2 + \epsilon$,
with $x_c$ and $x_t$ as states of the candidate and true explanation, and $\epsilon = .001$. 
We compute this cost for all possible ground-truth explanations ($n!$ orders, given $n$ concepts to change) and take the minimum alignment cost as the $eac$. For evaluating the $eac$ on the synthetic data, we compute the $eac$ for 90 generated ($a, b$) pairs and report the average $eac$.%

\textbf{Representation quality metrics}. Additionally, we explore the (adverse) effects of the conditioning methods on the learned representations. To quantify concept-disentanglement, the mutual information gap ($mig$)\cite{chen2018isolating} is used. We estimate the $mig$ for the class concepts in $z_y$ following the same procedure as \cite{locatello2019challenging}. The \textit{ELBO} metrics are also evaluated, denoted as $rec$ (reconstruction error), $kl_y$, and $kl_x$ (KL divergences of the subspaces). The classification accuracy, using the learned distribution $q_{{\psi{_y}}}(y|z_y)$, is denoted as $acc$. Finally, we evaluate the disentanglement of the latent subspaces \wrt class information, by training logistic regression classifiers on the latent space embeddings. Their accuracies are denoted as $l$-$acc_y$ and $l$-$acc_x$. %

\newcommand{\querybox}[1]{
\makebox[1cm]{ %
\raisebox{-.46cm}[0cm][0cm]{#1} %
}\hspace{-.2cm}}

\newcommand{\drawoutline}[1]{
    \begin{tikzpicture}
    \node[anchor=south west, inner sep=0] (image) at (0, 0) {\adjincludegraphics[valign=c,height=.023\textwidth]{#1}}; %
    \begin{scope}[x={(image.south east)},y={(image.north west)}]
       \draw[red,ultra thick,rounded corners] (-0.02,-0.02) rectangle (1.02,1.02);
    \end{scope}
    \end{tikzpicture}
}

\newcommand{\explanationbox}[1]{ %
\marginbox{0 .1\height}{\adjincludegraphics[trim={0 0 0 0},clip,valign=c,height=.0285\textwidth]{#1}}
}

\begin{figure*}[!hb]
\centering
    \subfloat[For synthetic samples.]{\label{fig:synonly}
    \addtolength{\tabcolsep}{-5pt}    
    \renewcommand*{\arraystretch}{1.1}
    \begin{tabular}{cc}
    \multicolumn{1}{c|}{
    \querybox{\drawoutline{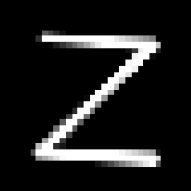}}
    \explanationbox{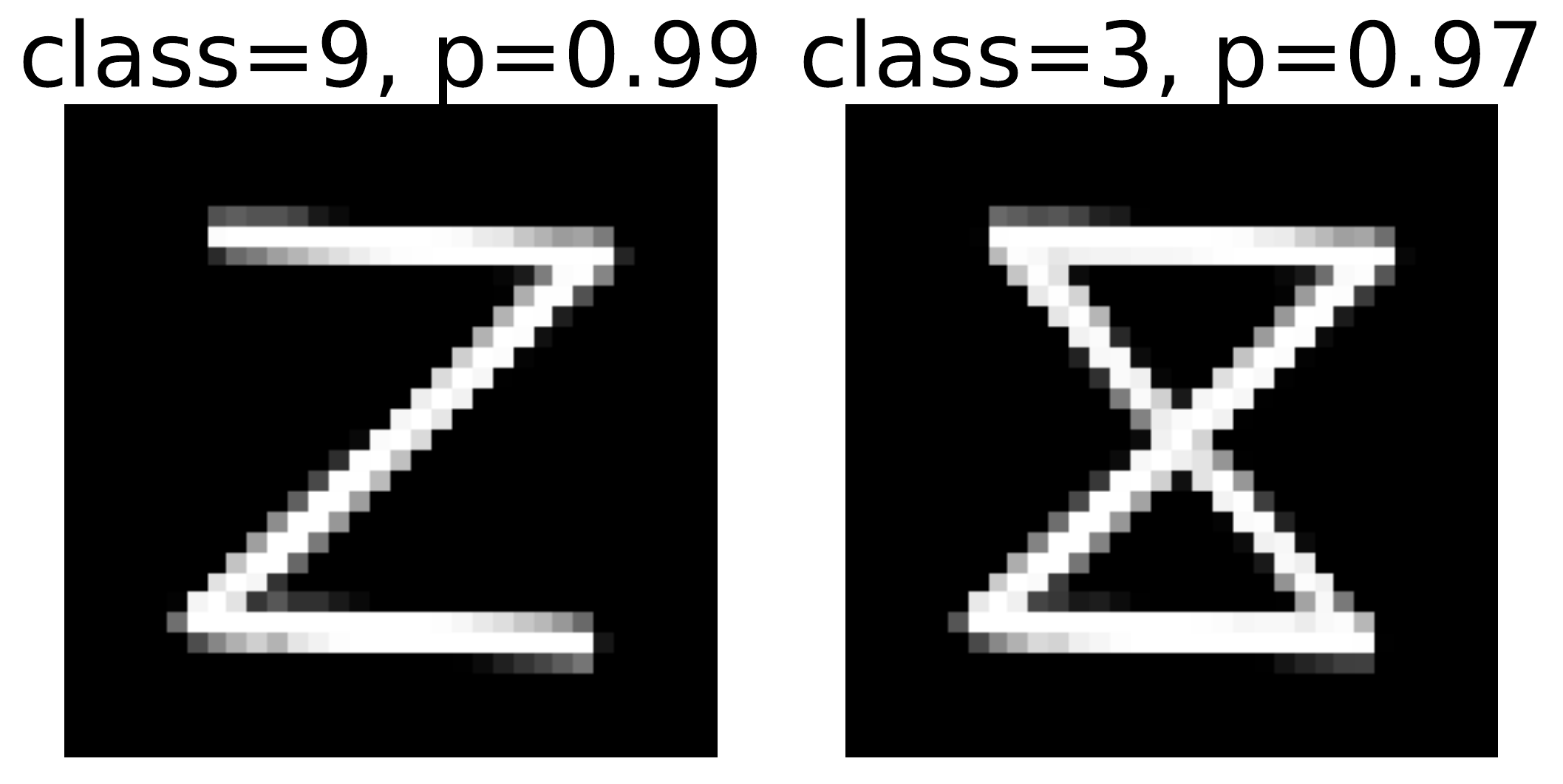}
    }
    & 
    \querybox{\drawoutline{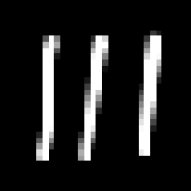}}
    \explanationbox{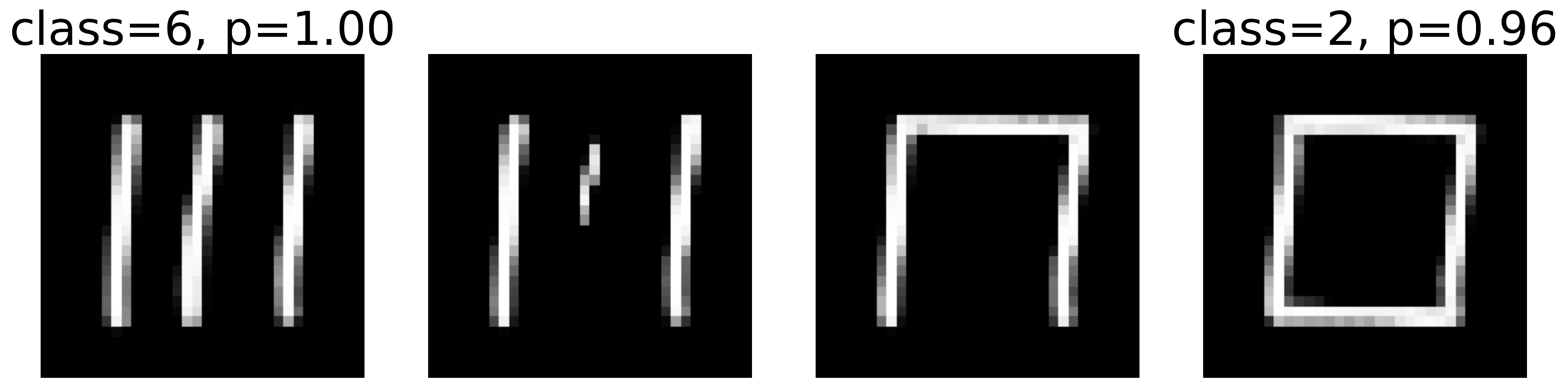} \\ \hline
    \multicolumn{1}{c|}{
    \querybox{\drawoutline{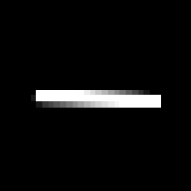}}
    \explanationbox{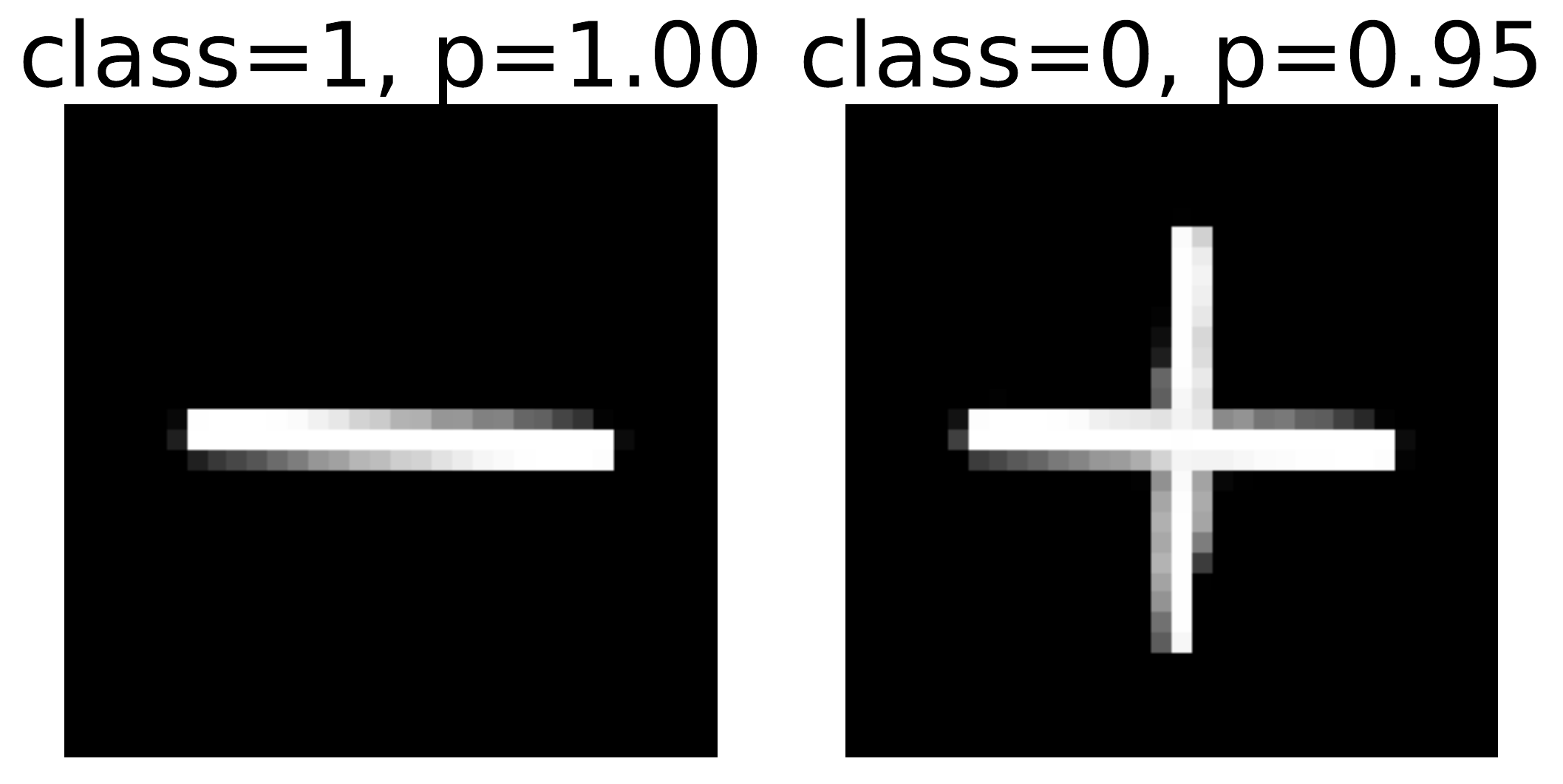}
    }
    & 
    \querybox{\drawoutline{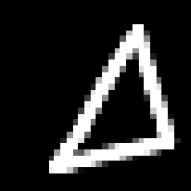}}
    \explanationbox{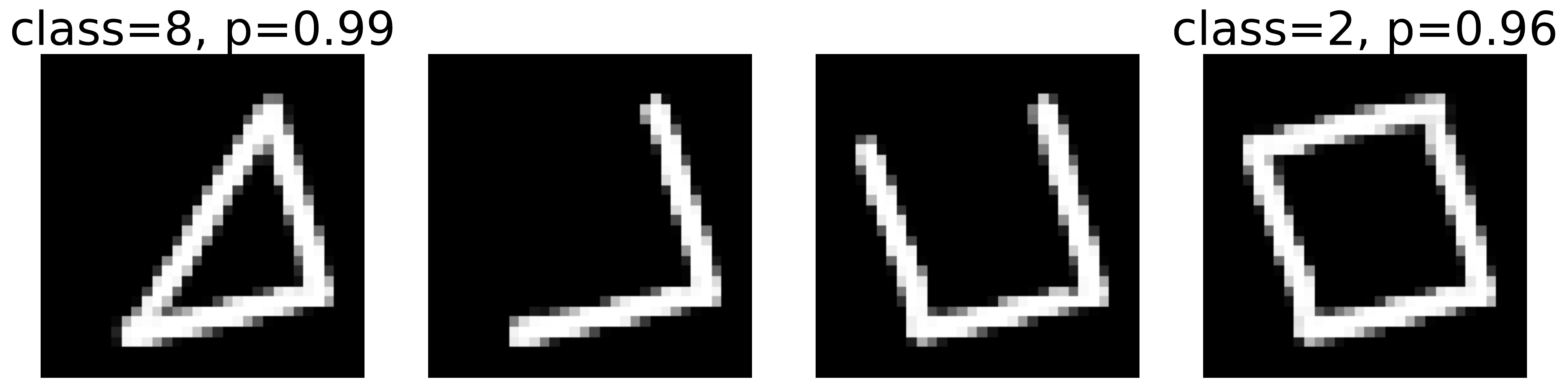}
    \end{tabular}
    \addtolength{\tabcolsep}{5pt}
    }
    \subfloat[For MNIST samples.]{\label{fig:mnistonly}
        \addtolength{\tabcolsep}{-5pt}    
    \renewcommand*{\arraystretch}{1.1}
    \begin{tabular}{cc}
    \multicolumn{1}{c|}{
    \querybox{\drawoutline{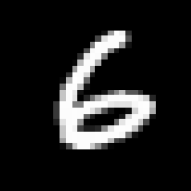}}
    \explanationbox{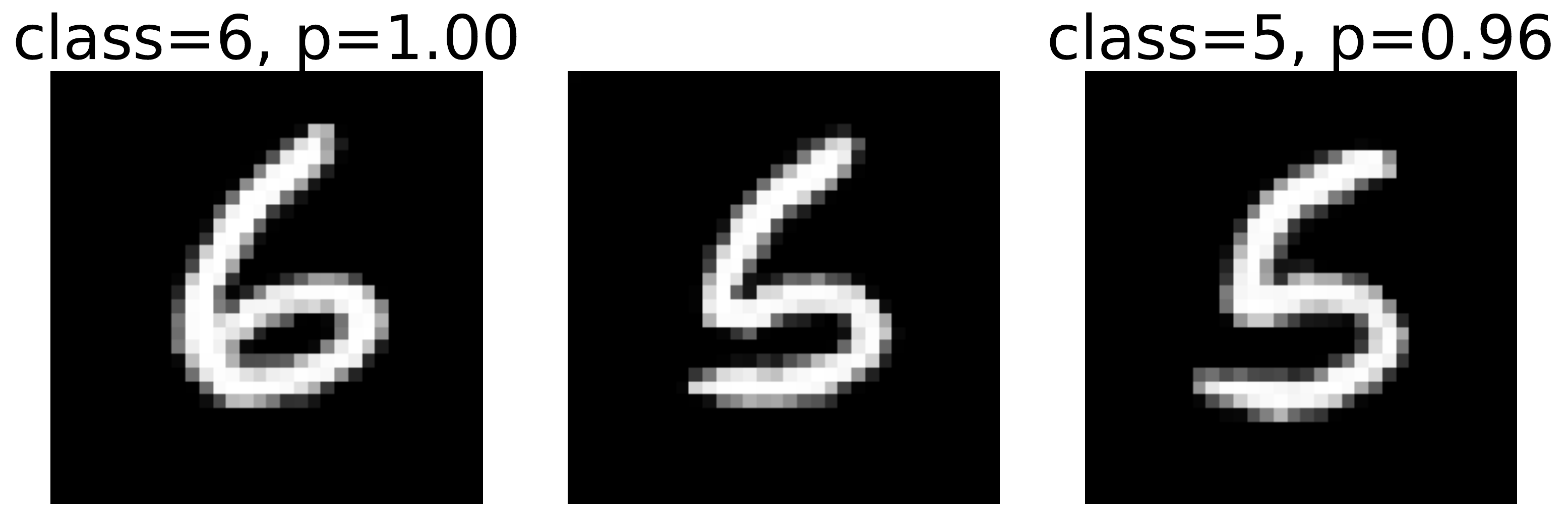}
    }
    & 
    \querybox{\drawoutline{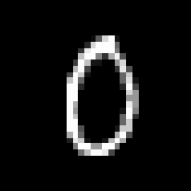}}
    \explanationbox{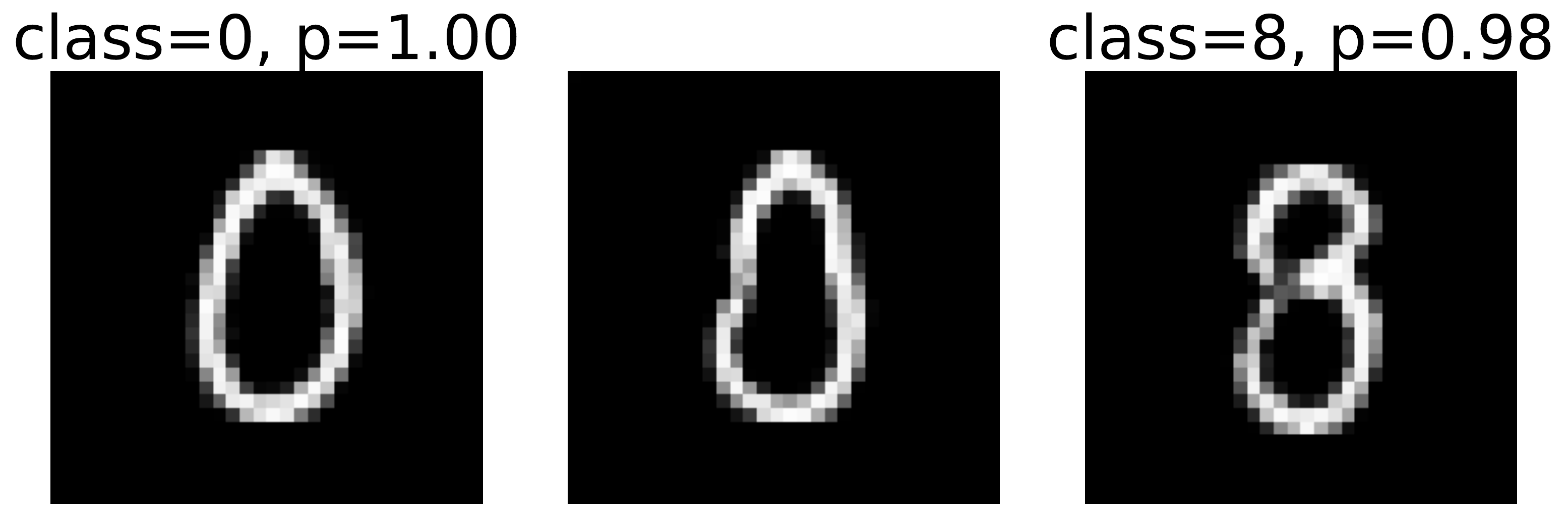} \\ \hline
    \multicolumn{1}{c|}{
    \querybox{\drawoutline{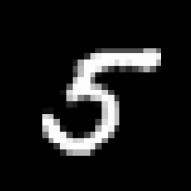}}
    \explanationbox{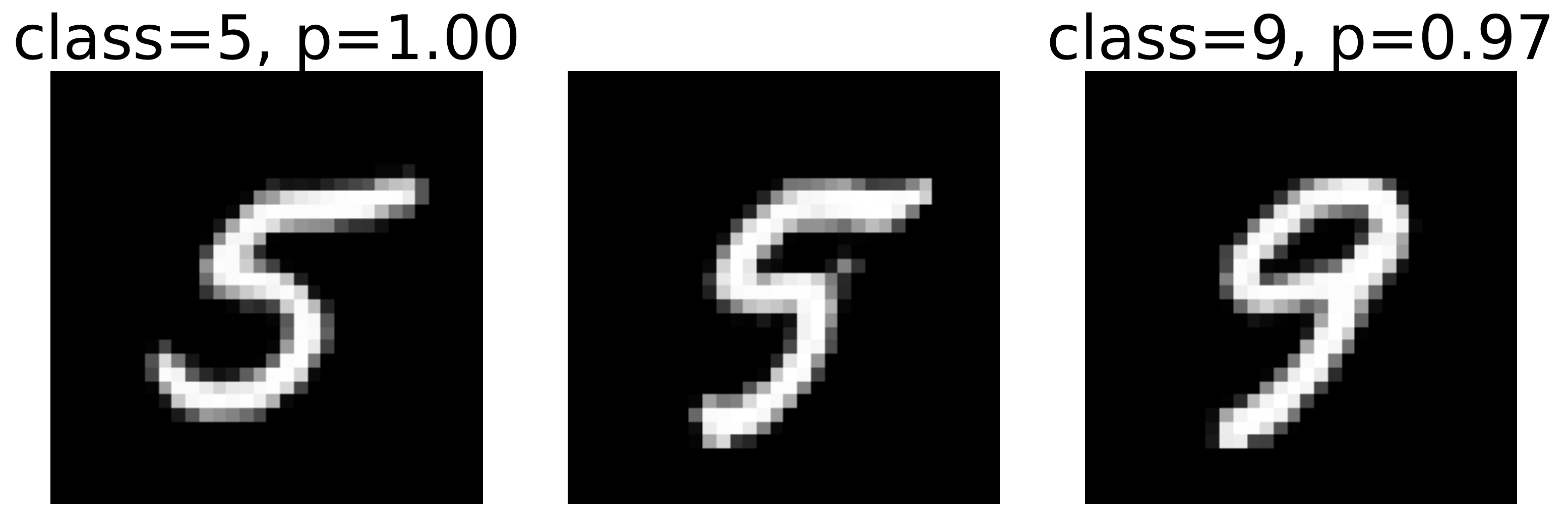}
    }
    & 
    \querybox{\drawoutline{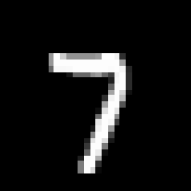}}
    \explanationbox{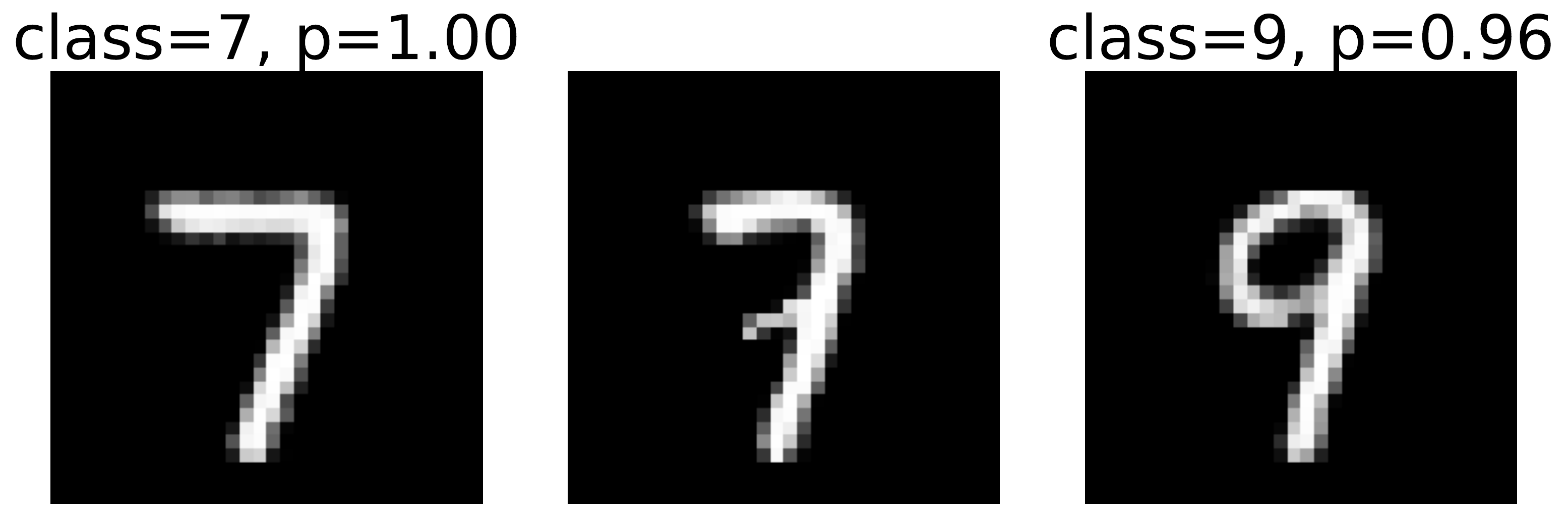}
    \end{tabular}
    \addtolength{\tabcolsep}{5pt}
    }

    \caption{Explanations generated by VAE-CE. The query datapoints are outlined in red, followed by an explanation transforming the datapoint to the second most likely class.}
    \label{fig:}
\end{figure*}

\textbf{Other evaluations}. We evaluate the exemplar identification by checking whether datapoints with more common concepts are more likely to be chosen. Class 9 has 2 variations, of which one variant has more concepts in common with classes 7 and 8. The remaining classes have the same number of concepts in common with both variants. We query for exemplars using 2000 test samples and compare the probability of selecting the more common variant using classes 7 and 8 to the probability when using other classes. Also, we qualitatively analyze the explanations, using both single datapoints to explain and input pairs to contrast. %
\subsection{Comparison overview}\label{ss:s:comp}
We compare VAE-CE to methods with similar capabilities, staying within the domain of VAE-based representation methods. The model described in \S\ref{ss:m:base} forms the baseline. We compare a set of alternative approaches to regularizing the $z_y$-space, alongside other interpolation approaches.

\textbf{Concept-disentanglement methods}. For each disentanglement approach we denote how we refer to it, alongside a short summarization of the regularization procedure and supervision. Some details differ from the original approaches as we adapt them to disentangle single dimensions and to be able to compare different types of supervision.

\textbf{DVAE} denotes the baseline model (\S\ref{ss:m:base}). \textbf{LVAE} denotes an extension of label-based disentanglement as described in \S\ref{ss:m:base}. For each $z_y$-dimension a label is provided indicating whether a concept is present. Each dimension is disentangled by two auxiliary classifiers, one predicting the label from the dimension value and one predicting the label from the remaining $z_y$ dimensions. The latter objective's gradients are reversed for the encoders. \textbf{GVAE} denotes an adaption of \cite{hosoya2019group} using pairs of datapoints with (at least) one specified matching concept. The inferred values for the $z_y$-dimension corresponding to this concept are averaged out, forcing this information to be shared through optimizing the \textit{ELBO}. \textbf{ADA-GVAE} denotes an adaption of \cite{locatello2020weak} that uses positive change pairs as supervision, allowing us to compare to a method using similar supervision. Training is done using pairs of datapoints that differ in a single concept. We infer latent dimensions for both datapoints and average all but one dimension between the pair. The independent dimension is selected as the dimension with the highest KL divergence (between the pair). Optimization is again done using the \textit{ELBO}. \textbf{VAE-CE} denotes our method (\S\ref{s:m}).

\textbf{Model implementations}. All methods share the same encoder and decoder architecture, and have a dimensionality of 8 for both $z_x$ and $z_y$. Hyperparameters are optimized using the $eac$ on a validation set of explanation pairs using synthetic data. As this cannot be evaluated for MNIST we use the same hyperparameters as chosen for the synthetic data; this approach resulted in reasonable models since the synthetic data was designed to share characteristics with MNIST. For details on architectures, training, and hyperparameters we refer to the supplementary material.

\textbf{Interpolation methods}. To evaluate the graph-based explanation approach, we also consider two naïve approaches to creating explanations. First, a smooth interpolation (denoted as $sm$), where each intermediate state of $z_y$ is a convex combination of $z_{y_a}$ and $z_{y_b}$. All dimensions are adjusted at once according to a predefined number of steps, in equal proportion for each step. We use five interpolation states. Second, a dimension-wise interpolation (denoted as $dim$), where we identify significantly differing dimensions with a simple heuristic: $|z_{y_{a_i}} - z_{y_{b_i}}| > 1$ (the $\sigma$ of the prior). All significantly different dimensions are changed one at a time, in arbitrary order. The non-significant dimensions are changed at once, in the first step. Finally, for the graph-based interpolation (denoted as $graph$), we use explanation parameters $t=.95$ (exemplar threshold), $\alpha=.5$ (realism), $\beta=1$ (change quality), and $\gamma=1$ (normalization).

\begin{table*}[!ht]
\fontsize{8.5}{10}\selectfont
\setlength\tabcolsep{3.5pt}
\begin{center}
\begin{tabular}{l l l l l l l l l}
 \cline{3-9}
\textbf{Dataset} & \textbf{Model} & $mig$ $\uparrow$ & $rec$ $\downarrow$ & $kl_y$ $\downarrow$ & $kl_x$ $\downarrow$ & $acc$ $\uparrow$ & $l$-$acc_y$ $\uparrow$ & $l$-$acc_x$ $\downarrow$ \\ \hline
Synthetic & DVAE & .1206 $\pm$ .030 & 11.93 $\pm$ .28 & \textbf{4.425} $\pm$ .16 & 7.540 $\pm$ .22 & \textbf{.9734} $\pm$ .0007 & \textbf{.9749} $\pm$ .0006 & \textbf{.1700} $\pm$ .015\\
&LVAE &  .4227 $\pm$ .047 & 13.49 $\pm$ 1.1 & 10.96 $\pm$ 5.4 & 7.054 $\pm$ .22 & .9537 $\pm$ .0080 & .9618 $\pm$ .0022 & .2080 $\pm$ .057\\
&GVAE & .1484 $\pm$ .069 & \textbf{10.19} $\pm$ .19 & 7.072 $\pm$ .57 & 5.738 $\pm$ .76 & .9621 $\pm$ .0010 & .9641 $\pm$ .0007 & .2023 $\pm$ .023\\
&ADA-GVAE & .3402 $\pm$ .073 & 10.67 $\pm$ .34 & 8.233 $\pm$ 1.2 & \textbf{4.940} $\pm$ 1.1 & .9585 $\pm$ .0008 & .9610 $\pm$ .0011 & .1801 $\pm$ .036\\
&VAE-CE & \textbf{.4923} $\pm$ .033 & 14.51 $\pm$ .60 & 7.808 $\pm$ .14 & 7.944 $\pm$ .46 & .9629 $\pm$ .0009 & .9657 $\pm$ .0004 & .1822 $\pm$ .014\\
\hline
MNIST & DVAE && 16.01 $\pm$ .35 & \textbf{4.341} $\pm$ .31 & 8.043 $\pm$ .25 & \textbf{.9911} $\pm$ .0007 & \textbf{.9940} $\pm$ .0005 & .1822 $\pm$ .0094\\
&ADA-GVAE && \textbf{13.64} $\pm$ .19 & 10.14 $\pm$ 1.1 & \textbf{3.604} $\pm$ .90 & .9645 $\pm$ .0017 & .9701 $\pm$ .0015 & .1997 $\pm$ .015\\
&VAE-CE && 21.67 $\pm$ 1.34 & 7.117 $\pm$ .29 & 6.646 $\pm$ .35 & .9795 $\pm$ .0019 & .9834 $\pm$ .0017 & \textbf{.1817} $\pm$ .012\\
\hline
\end{tabular}
\end{center}
\caption{Representation quality metrics for synthetic data and MNIST.}\label{t:rep}
\end{table*}
\section{Results}
All results are reported as mean $\pm$ standard deviation, with the best mean marked in bold. We do more extensive comparisons using the synthetic data (both more methods and metrics) as access to ground-truth generative factors allows for more supervision and metric evaluations. For qualitative comparisons, other methods' explanations are generated using the interpolation method with the lowest $eac$ ($sm$ or $dim$), whereas VAE-CE uses $graph$-based interpolation.

\textbf{Synthetic data}. For the \textit{explanation quality}, the full results are provided in Table~\ref{t:syneac}. VAE-CE provides the best results, the graph-based $eac$ (\ie the full method) is significantly lower than any other. As a small ablation study we also consider the $eac$ given naïve interpolation methods, and use the graph-interpolation procedure and components from VAE-CE and apply it to other models. While the use of either component shows performance improvements, the scores are dominated by the combination thereof. Pair-based explanations and the closest ground-truths (as used for the $eac$) are depicted in Fig.~\ref{fig:syncomp}. We can observe that VAE-CE transforms individual lines better than alternative approaches. Examples of individual explanations (from only a query datapoint) are depicted in Fig.~\ref{fig:synonly}.

An overview of \textit{representation-quality} metrics is provided in Table~\ref{t:rep} (top). We can observe that the $mig$ seems strongly correlated with the explanation quality, with VAE-CE performing best. Other metrics vary, with the baseline (DVAE) performing the best classification-wise. As such, the extra regularization comes at a cost. %

For the \textit{exemplar selection experiment}, the baseline probability of picking the evaluated variant of class 9 was $.7902 \pm .018$. Using classes 7 and 8 this variant was selected at a probability of $.9378 \pm .044$, hinting that a variant with more common factors is more likely to be chosen.

\definecolor{dgray}{gray}{0.3}

\begin{table}[!bt]
\fontsize{8.5}{10}\selectfont
\setlength\tabcolsep{3.5pt}
\begin{center}
\begin{tabular}{l l l l}
\cline{2-4}
\textbf{Model} & $eac$-$sm$ $\downarrow$ & $eac$-$dim$ $\downarrow$ & $eac$-$graph$ $\downarrow$\\ \hline
DVAE & 27.53 $\pm$ .29 & 27.66 $\pm$ .41 & \textcolor{dgray}{\textit{26.87 $\pm$ .31}}\\
LVAE & 26.95 $\pm$ .45 & 25.38 $\pm$ 1.6 & \textcolor{dgray}{\textit{23.86 $\pm$ 1.0}}\\
GVAE & 27.87 $\pm$ .60 & 28.90 $\pm$ .65 & \textcolor{dgray}{\textit{26.52 $\pm$ .72}}\\
ADA-GVAE & \textbf{26.84} $\pm$ .57 & 26.22 $\pm$ 1.1 & \textcolor{dgray}{\textit{23.24 $\pm$ 1.2}}\\
VAE-CE & 28.96 $\pm$ .82 & \textbf{21.78} $\pm$ .55 & \textbf{19.92} $\pm$ .64\\
\hline
\end{tabular}
\end{center}
\caption{Explanation quality results on the synthetic data. Note that $eac$-$graph$ relies on VAE-CE components, other methods cannot independently produce these explanations.}\label{t:syneac}
\end{table}

\newcommand{\figheightb}{.0157}
\newcommand{\vpaddingb}{.05\height}
\begin{figure}[!hb]
    \centering
    \addtolength{\tabcolsep}{-5pt}    
    \renewcommand*{\arraystretch}{1.05}
    \begin{tabular}{l c|c}
        \scriptsize \raisebox{-.065cm}{input} & \marginbox{0 \vpaddingb}{\adjincludegraphics[valign=c,height=\figheightb\textwidth]{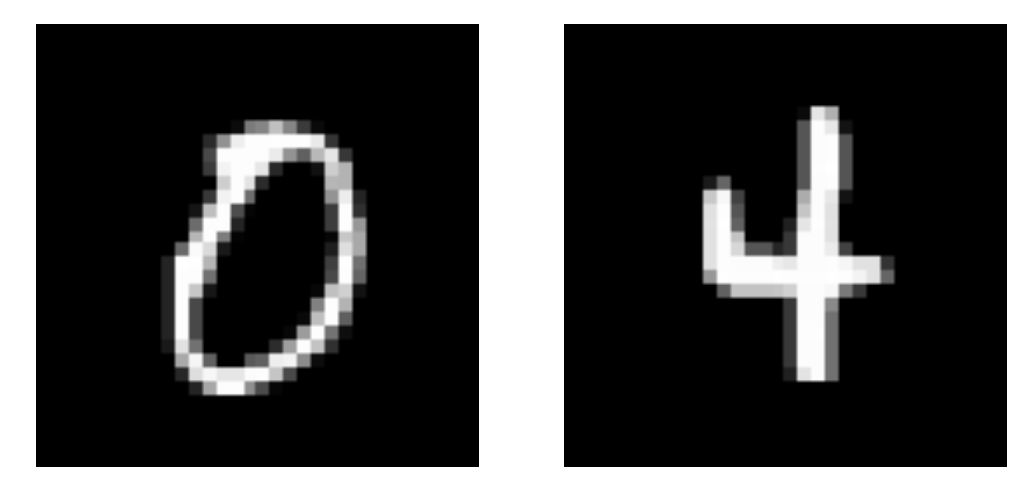}} & \marginbox{0 \vpaddingb}{\adjincludegraphics[valign=c,height=\figheightb\textwidth]{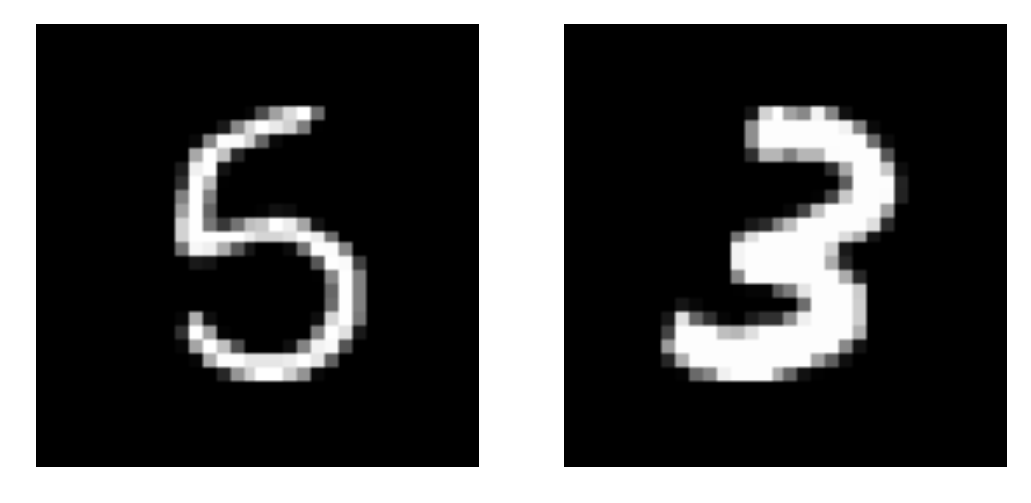}} \\\hline
        \scriptsize \raisebox{-.065cm}{DVAE} & \marginbox{0 \vpaddingb}{\adjincludegraphics[valign=c,height=\figheightb\textwidth]{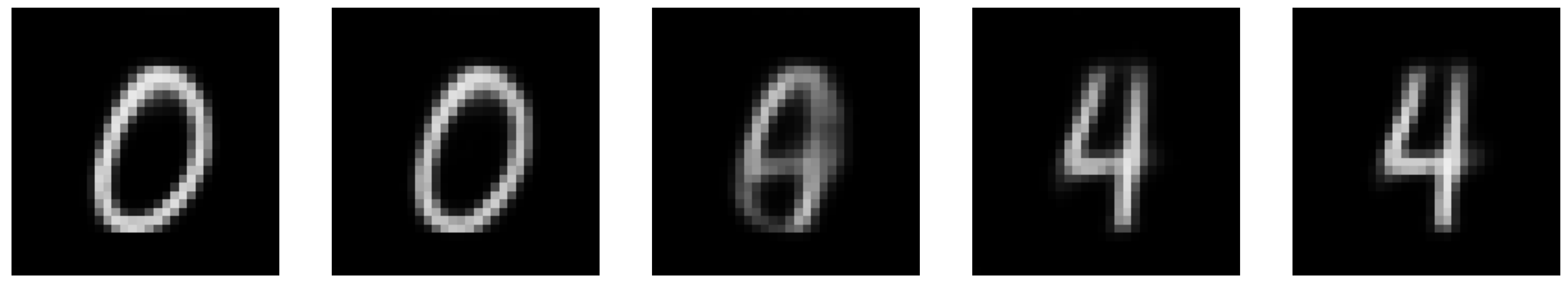}} & \marginbox{0 \vpaddingb}{\adjincludegraphics[valign=c,height=\figheightb\textwidth]{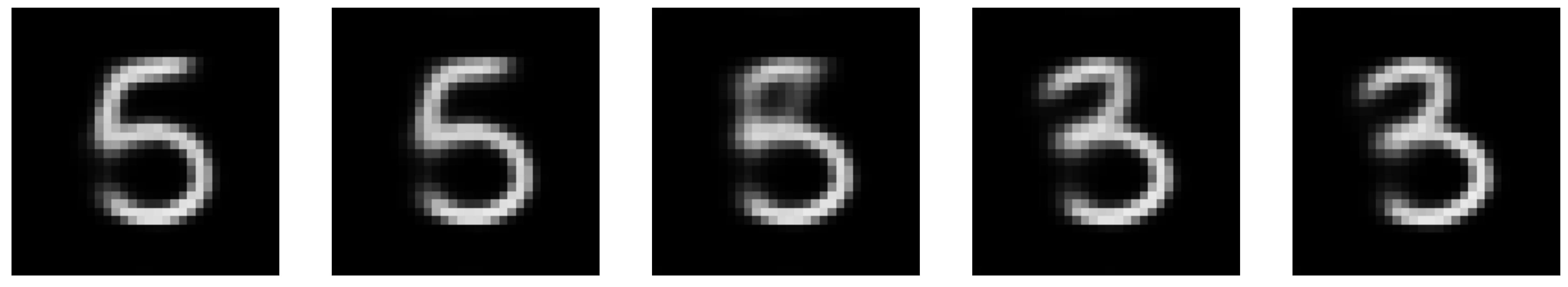}} \\\hline
        \makebox[0cm][l]{\scriptsize \raisebox{-.065cm}{ADA-GVAE}} & \marginbox{0 \vpaddingb}{\adjincludegraphics[valign=c,height=\figheightb\textwidth]{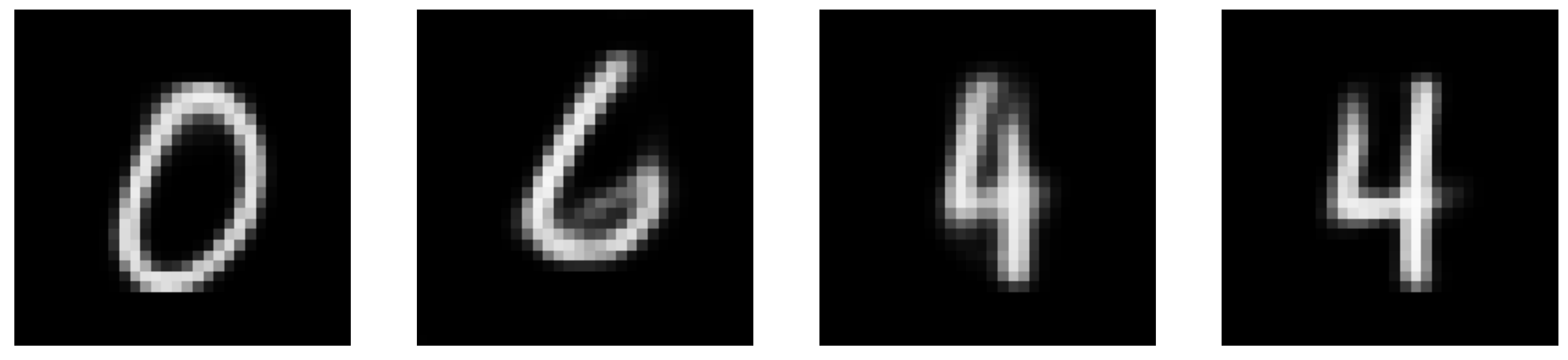}} & \marginbox{0 \vpaddingb}{\adjincludegraphics[valign=c,height=\figheightb\textwidth]{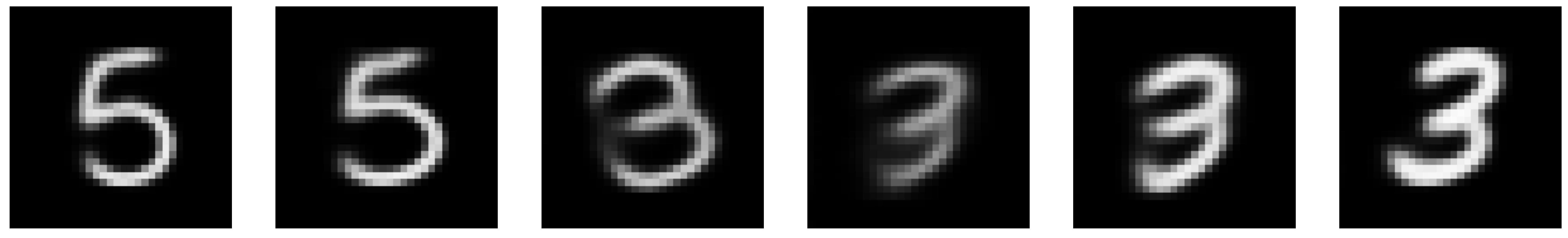}} \\\hline
        \scriptsize \raisebox{-.065cm}{VAE-CE}\hphantom{c} & \marginbox{0 \vpaddingb}{\adjincludegraphics[valign=c,height=\figheightb\textwidth]{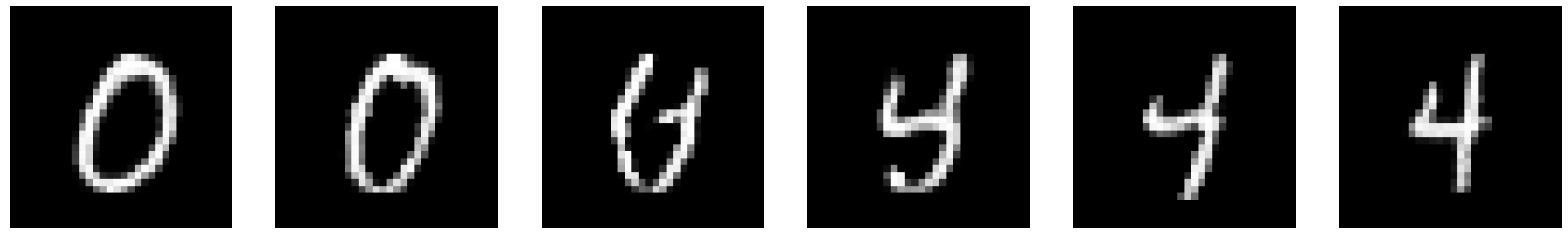}} & \marginbox{0 \vpaddingb}{\adjincludegraphics[valign=c,height=\figheightb\textwidth]{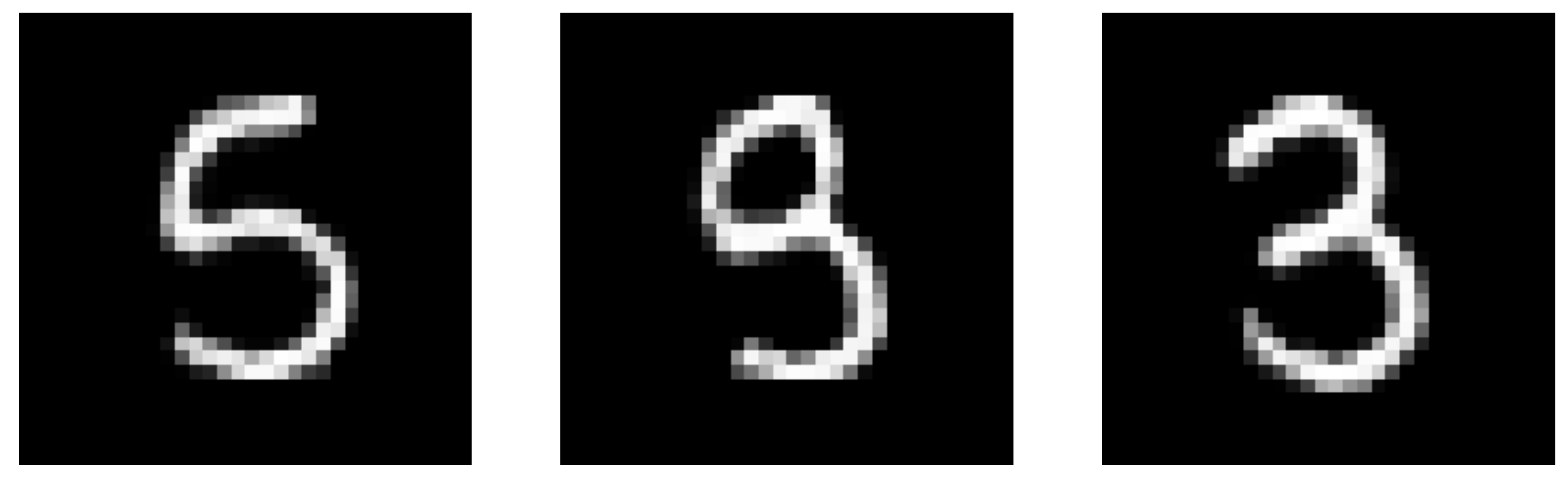}} \\
    \end{tabular}
    \addtolength{\tabcolsep}{5pt}
    \caption{MNIST explanations using provided input-pairs.}
    \label{fig:mnistcomp}
\end{figure}

\newcommand{\figcrop}{.1}
\newcommand{\figheight}{.037}
\newcommand{\vpadding}{.08\height}
\begin{figure}[!t]
    \centering
    \addtolength{\tabcolsep}{-5pt}    
    \renewcommand*{\arraystretch}{2}
    \begin{tabular}{l c|c}
        \footnotesize DVAE & \marginbox{0 \vpadding}{\adjincludegraphics[trim={0 0 0 {\figcrop\height}},clip,valign=c,height=\figheight\textwidth]{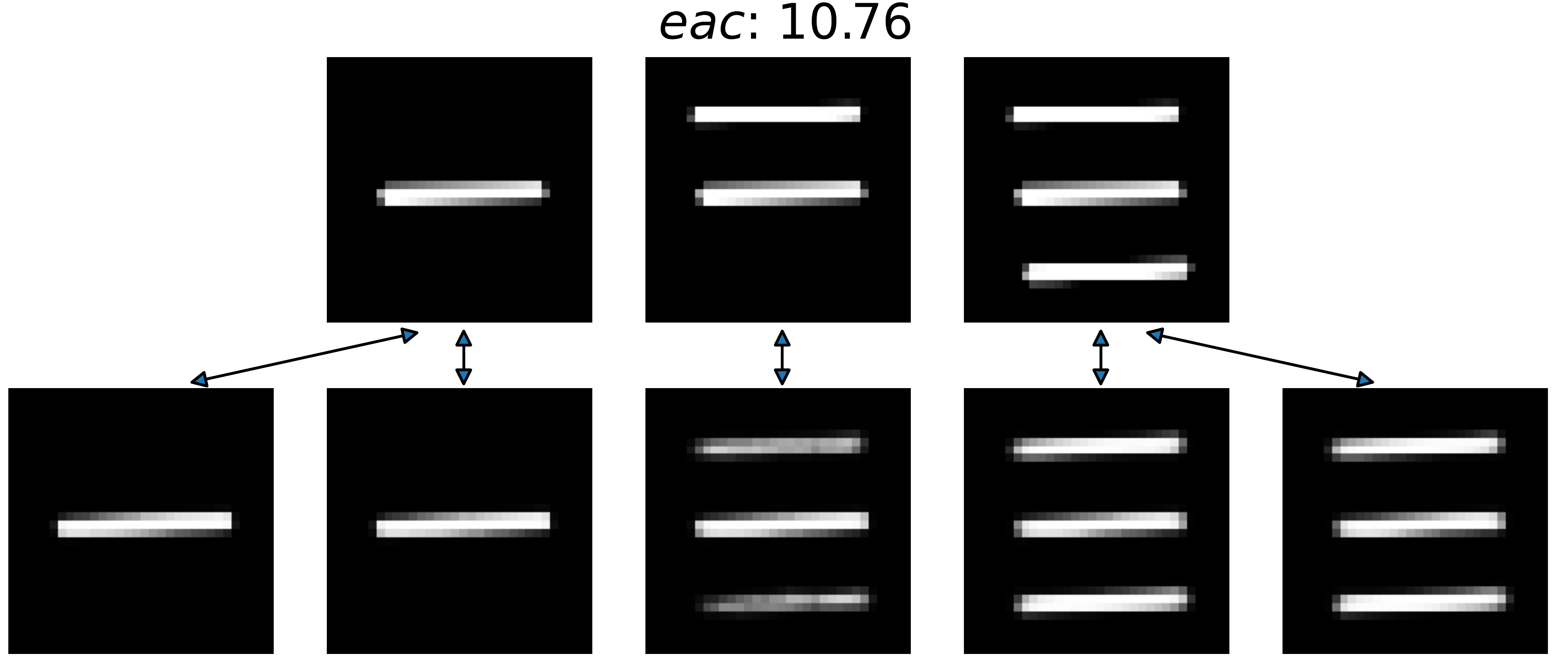}} & \marginbox{0 \vpadding}{\adjincludegraphics[trim={0 0 0 {\figcrop\height}},clip,valign=c,height=\figheight\textwidth]{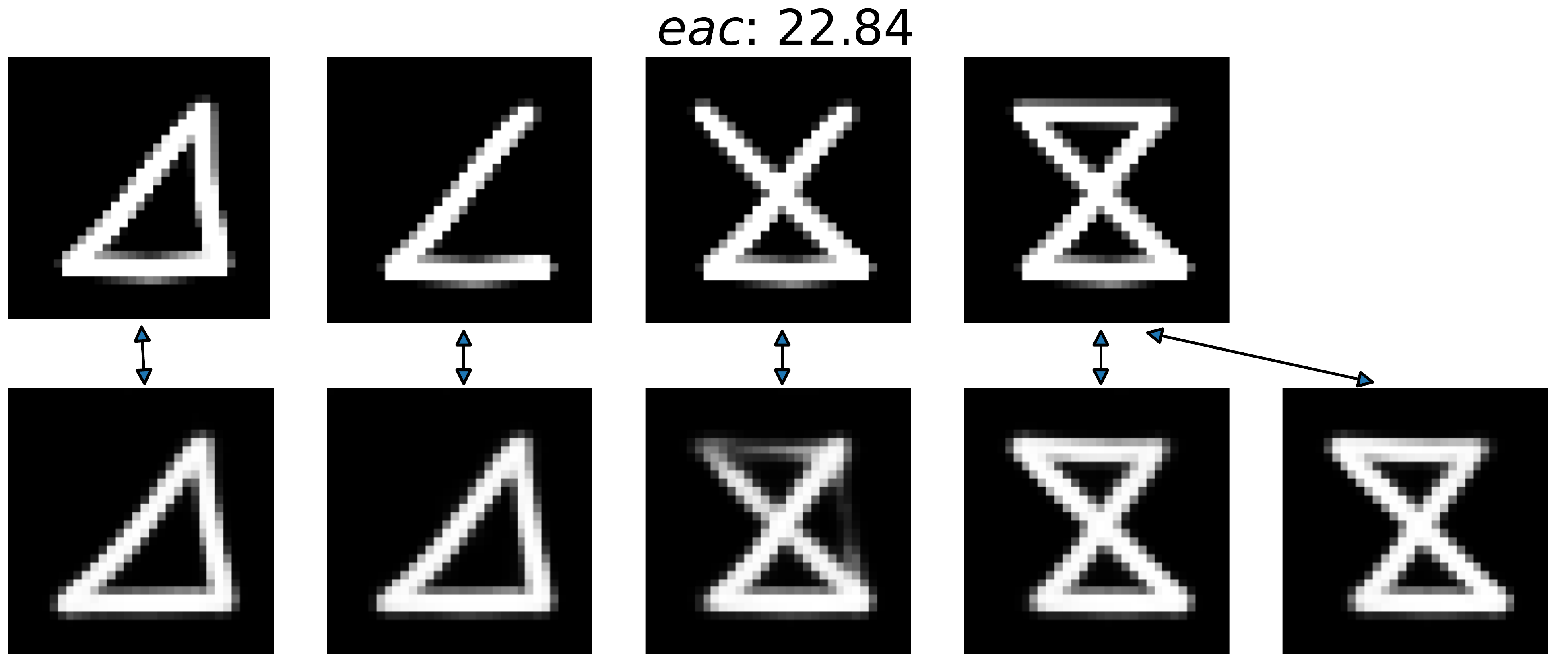}} \\\hline
        \footnotesize LVAE & \marginbox{0 \vpadding}{\adjincludegraphics[trim={0 0 0 {\figcrop\height}},clip,valign=c,height=\figheight\textwidth]{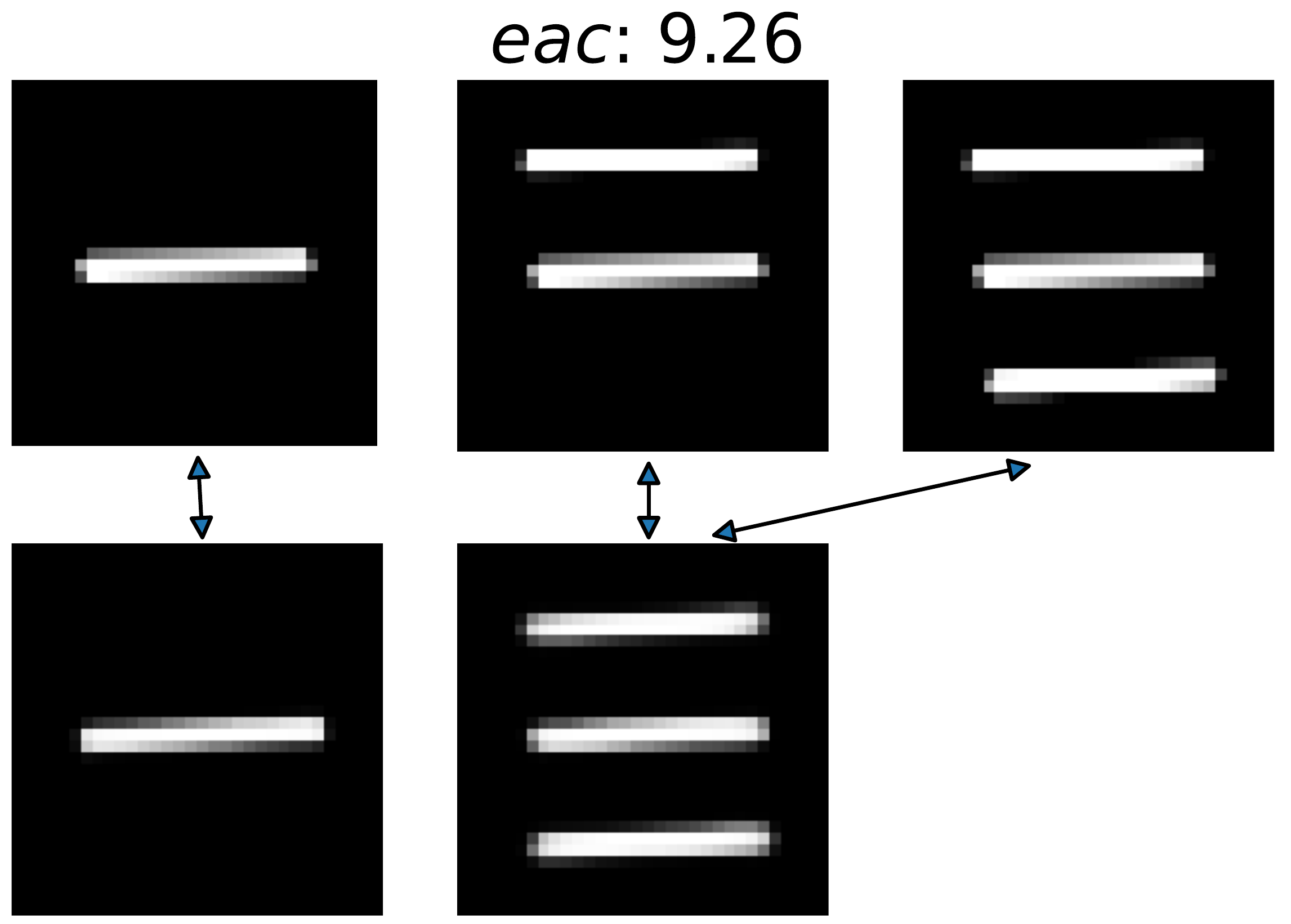}} & \marginbox{0 \vpadding}{\adjincludegraphics[trim={0 0 0 {\figcrop\height}},clip,valign=c,height=\figheight\textwidth]{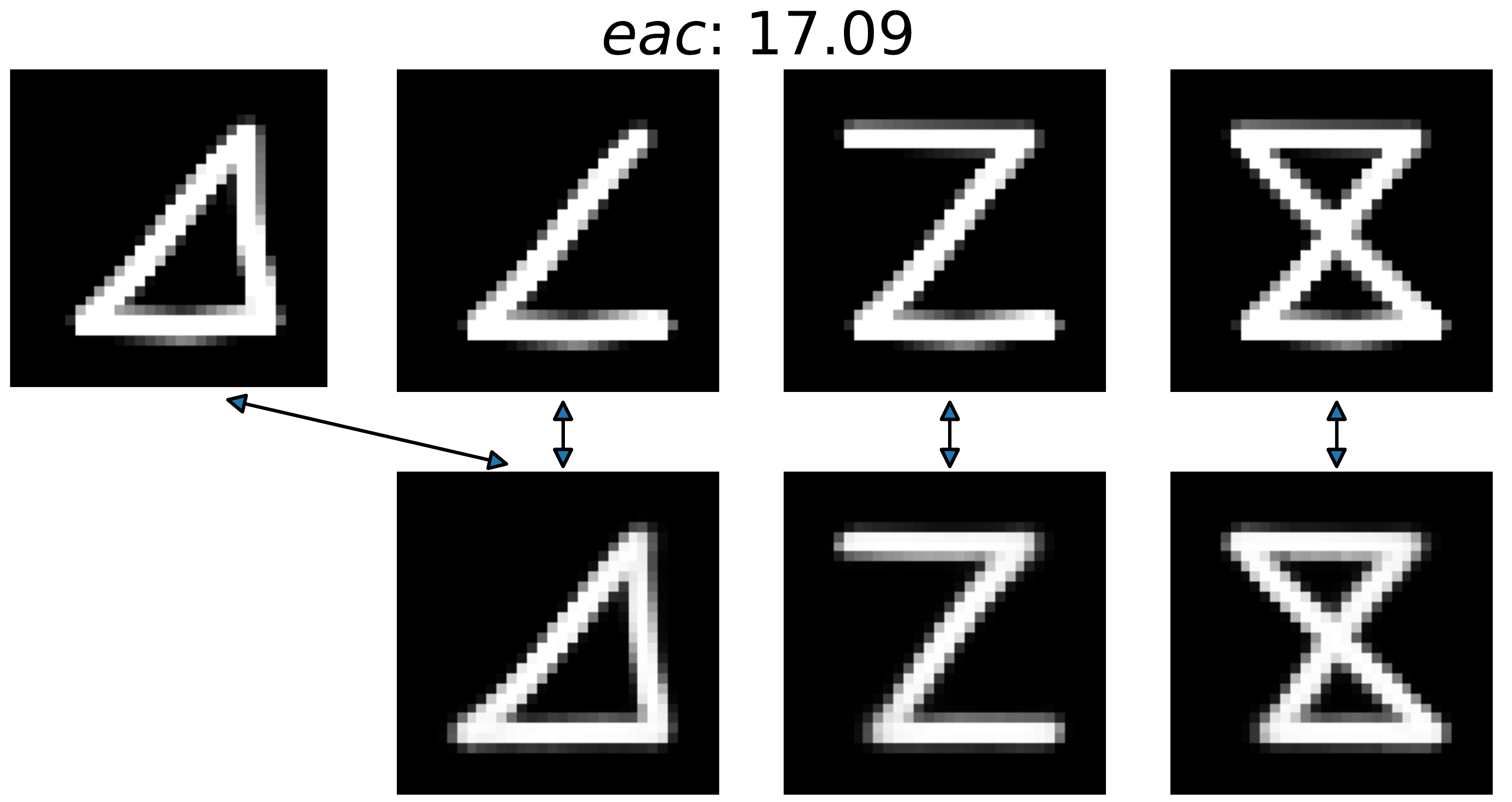}} \\\hline
        \footnotesize GVAE & \marginbox{0 \vpadding}{\adjincludegraphics[trim={0 0 0 {\figcrop\height}},clip,valign=c,height=\figheight\textwidth]{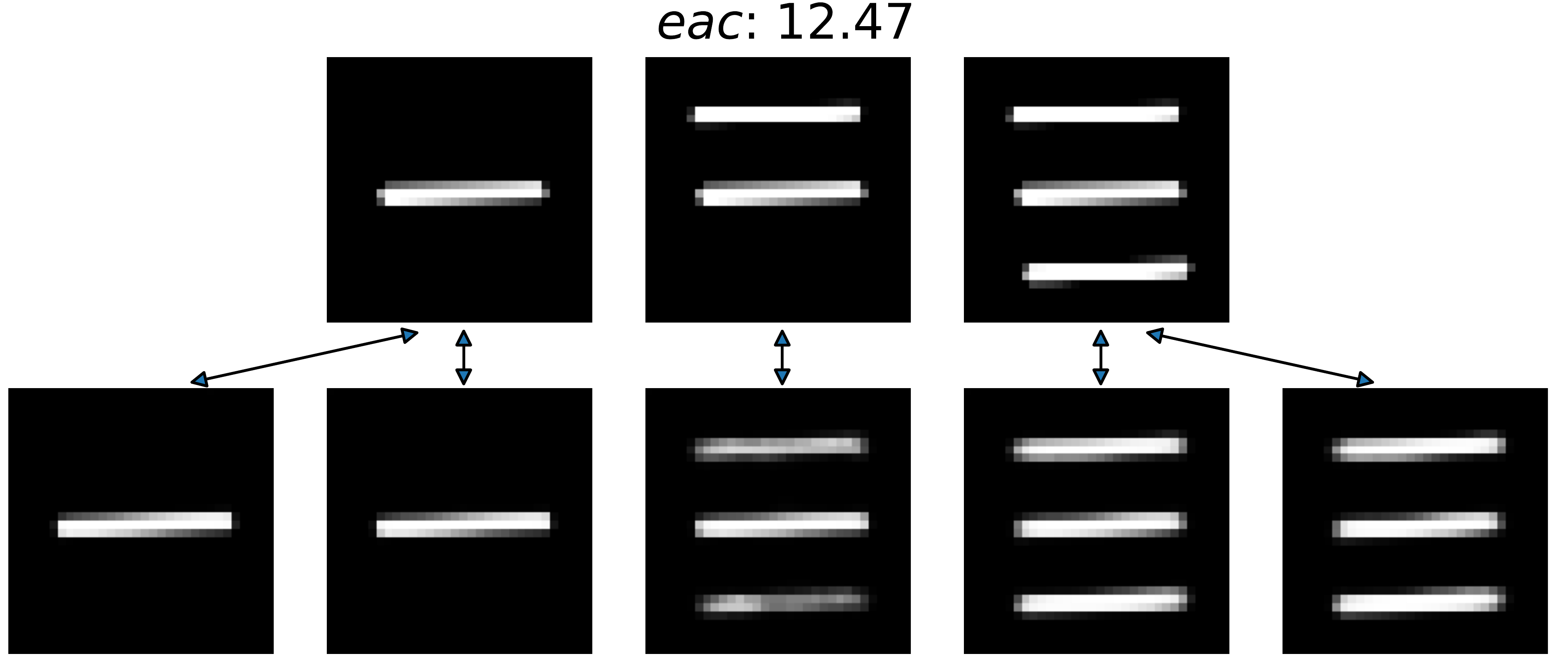}} & \marginbox{0 \vpadding}{\adjincludegraphics[trim={0 0 0 {\figcrop\height}},clip,valign=c,height=\figheight\textwidth]{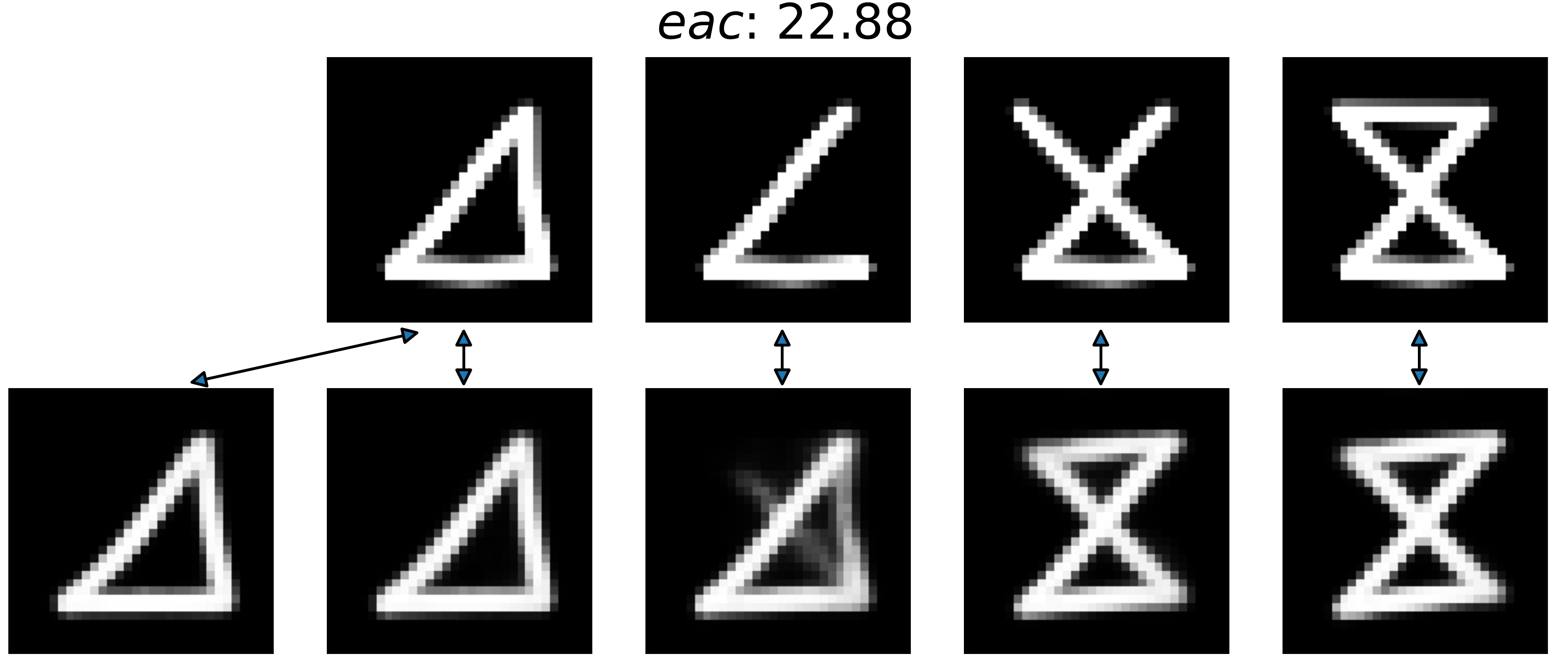}} \\\hline
        \makebox[0cm][l]{\footnotesize ADA-GVAE} & \marginbox{0 \vpadding}{\adjincludegraphics[trim={0 0 0 {\figcrop\height}},clip,valign=c,height=\figheight\textwidth]{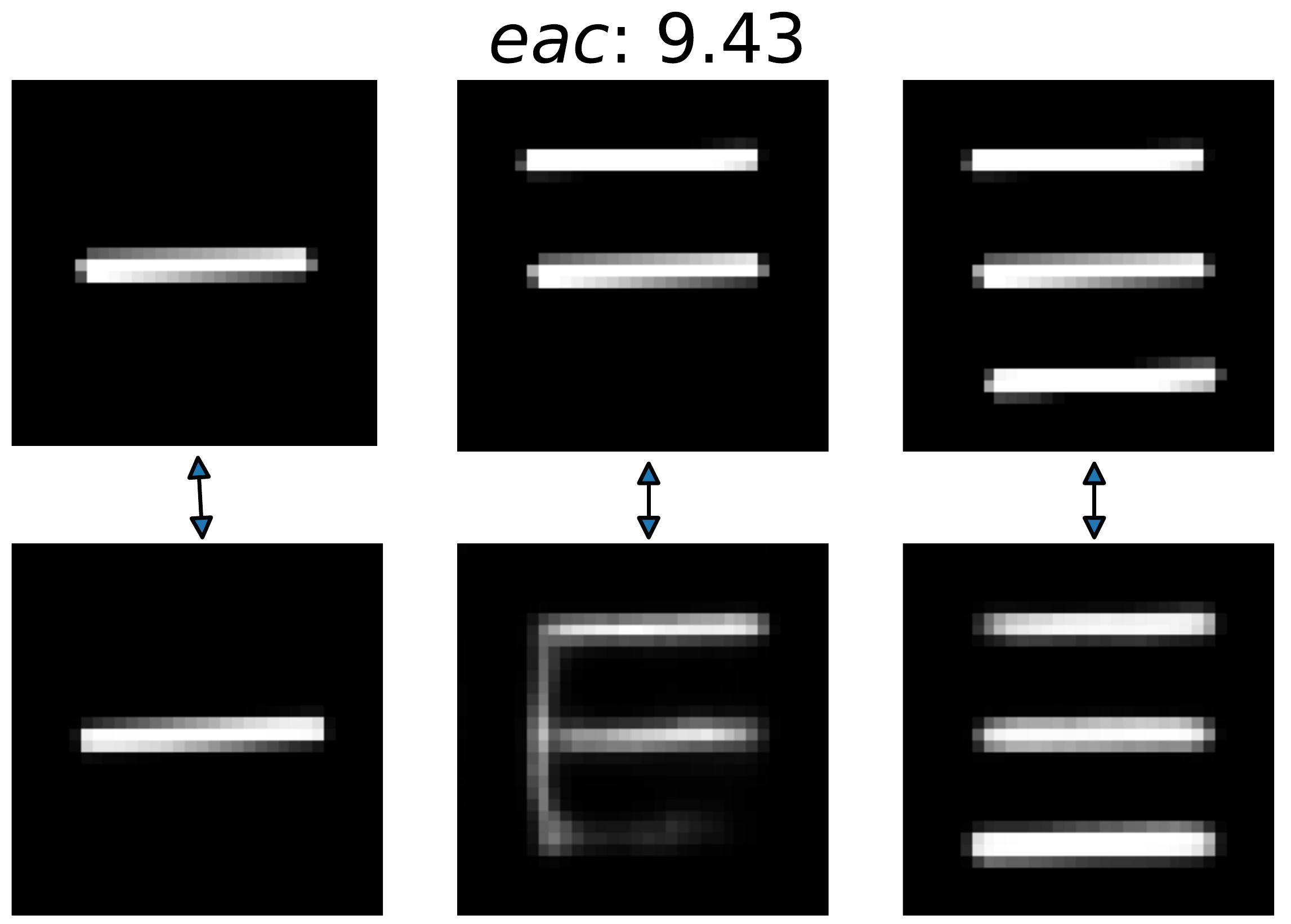}} & \marginbox{0 \vpadding}{\adjincludegraphics[trim={0 0 0 {\figcrop\height}},clip,valign=c,height=\figheight\textwidth]{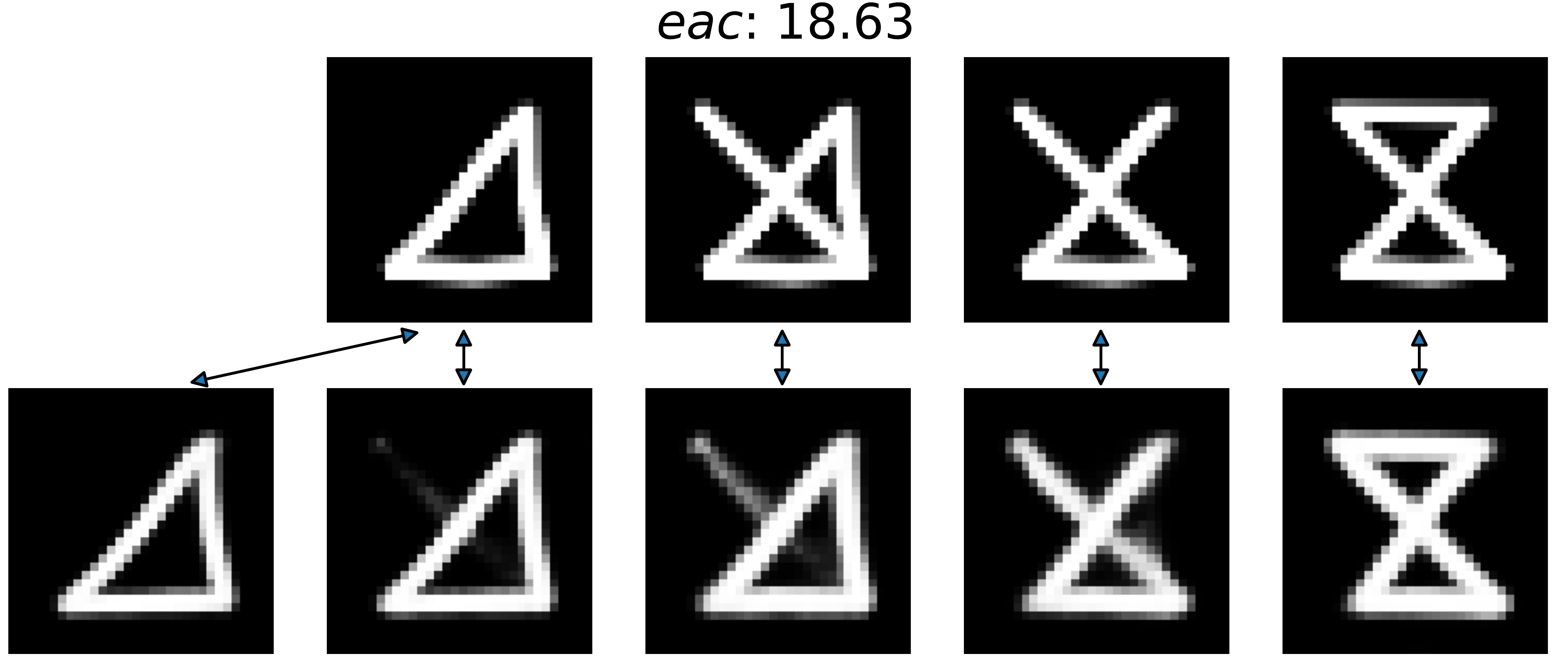}} \\\hline
        \footnotesize VAE-CE & \marginbox{0 \vpadding}{\adjincludegraphics[trim={0 0 0 {\figcrop\height}},clip,valign=c,height=\figheight\textwidth]{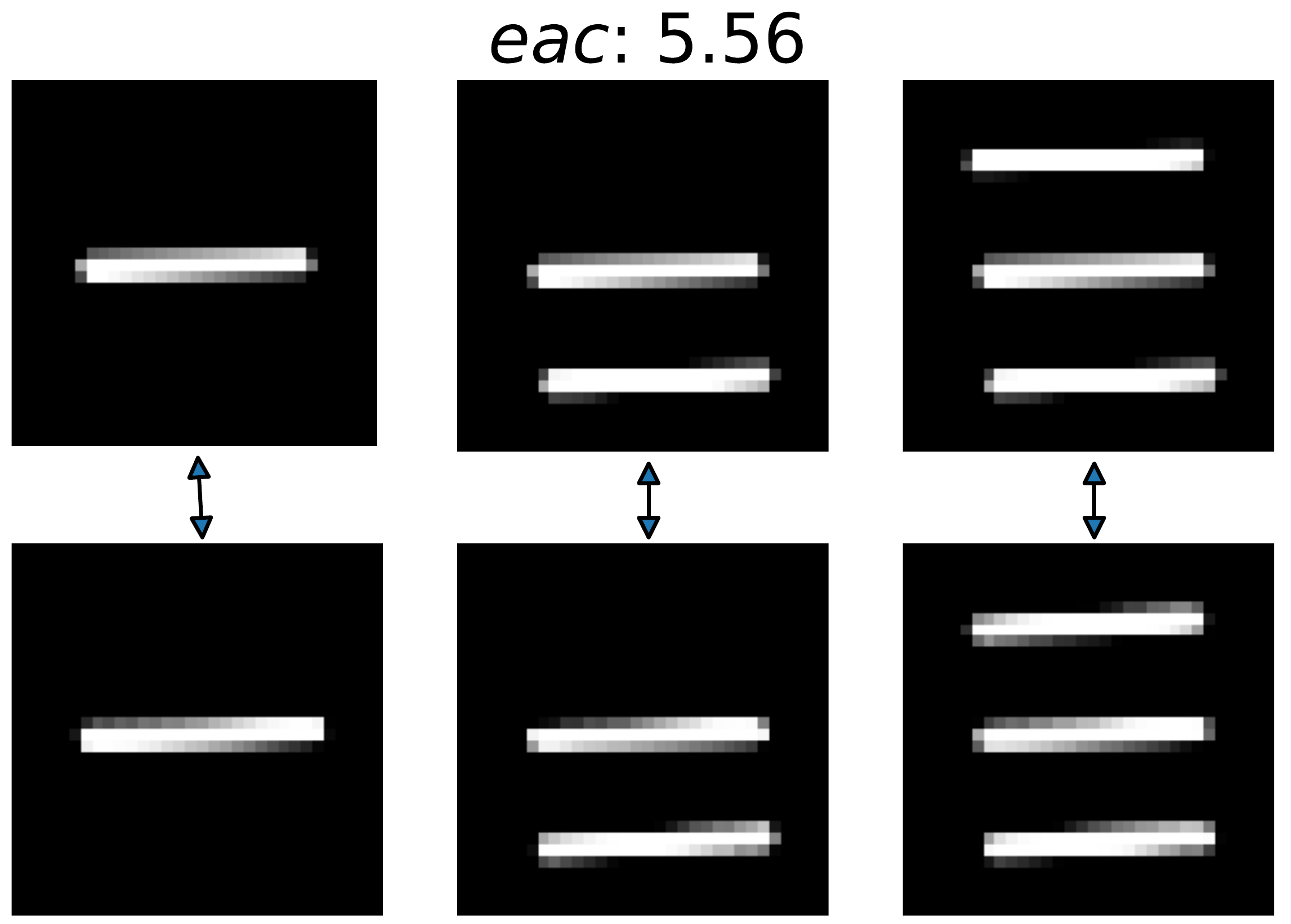}} & \marginbox{0 \vpadding}{\adjincludegraphics[trim={0 0 0 {\figcrop\height}},clip,valign=c,height=\figheight\textwidth]{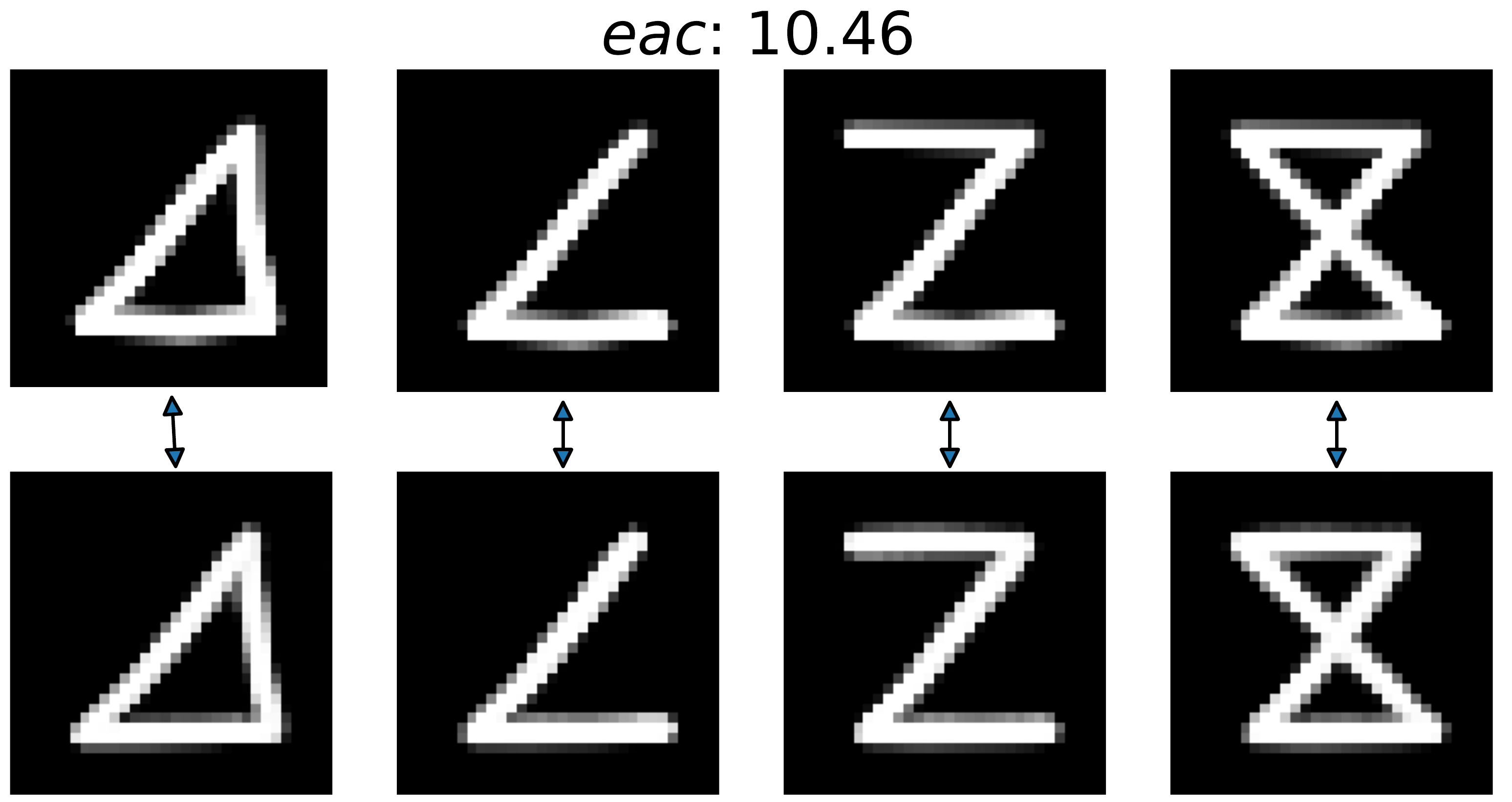}} \\
    \end{tabular}
    \addtolength{\tabcolsep}{5pt}
    \caption{Synthetic-data explanations using provided pairs from class 1 to 5 (left) and class 8 to 3 (right). The top row indicates the closest ground-truth explanation, whereas the bottom row depicts the created interpolation.}
    \label{fig:syncomp}
\end{figure}

\textbf{MNIST}. The \textit{explanation quality} is evaluated by comparing pair-based explanations and by evaluating datapoint-explanations. The resulting explanations are depicted in Figs.~\ref{fig:mnistcomp} and~\ref{fig:mnistonly}, respectively.
Although individual line changes are apparent in VAE-CE's explanations they are noisier than before, likely because of the more complex shapes and noisy supervision. This is very apparent in \eg the transition from 0-4 in Fig.~\ref{fig:mnistcomp}. However, the other approaches do not explain well using the assumed concept of lines, but rather change multiple components at once.

The \textit{representation quality} metrics are provided in Table~\ref{t:rep} (bottom). These results paint a similar picture as before, with no method dominating all metrics. We note that DVAE's accuracy is substantially higher than that of other methods, showing that the regularization methods again do not align perfectly with the representation of classes. %

\section{Conclusions}
In this paper, we proposed an interpretability-focused classification model that creates explanations in a concept domain $C$. This method extends a class-disentangled VAE with a new supervised regularization method for disentangling individual dimensions. Using this model we generate visual contrastive explanations, highlighting class-differing concepts using a sequence of transformations. An introductory empirical evaluation shows that the components of our method provide benefits over existing approaches, although applying it to more complex data remains future work. 
Ultimately, we believe that the proposed method allows us to effectively learn a model that represents and explains data in domain $C$, providing us with a more understandable and trustworthy classification model. Topics still of interest consider exploring more complex datasets, more efficient (heuristic search-based) approaches to explanation generation, and $CD$ implementations using less supervision.

\newpage
{\small
\bibliographystyle{ieee_fullname}
\bibliography{out}
}

\end{document}


\newcommand{\todo}[1]{{\color{red}\textbf{\textit{note: #1}}}}
\newcommand{\loss}{\mathcal{L}}
\def\Wrt{W.r.t\onedot}
\title{Supplementary Material \\ VAE-CE: Visual Contrastive Explanation using Disentangled VAEs}

\author{Yoeri Poels\hspace{1cm}Vlado Menkovski\\
Eindhoven University of Technology, the Netherlands
\\
{\tt\small \{y.r.j.poels, v.menkovski\}@tue.nl}
}

\maketitle

\section{Change Discriminator ($\bf{{CD}}$)}
The change discriminator ($CD$) is implemented as a binary classifier. It takes a pair of datapoints as input and predicts, in range $[0, 1]$, whether the change in datapoints is desirable, with 0 being undesirable and 1 being desirable. It is trained using pairs of datapoints depicting such changes, alongside a label indicating whether their change is desirable or not.

$CD$ is heavily regularized in an attempt to deal with samples dissimilar to the training set (as it will classify interpolation pairs) and to exclude class-agnostic noise in its prediction. We use the disentangled VAE (from \S3.1 in the main paper) as a base model for $CD$. In a separate pass we supply change pairs $(x_a, x_b)$ and use their class-related embedding, $z_{y_a}$ and $z_{y_b}$, to infer the change quality. We take the absolute difference of these embeddings and put them through a discriminative model to infer the final prediction for pair ($x_a, x_b$):
\begin{alignat}{1}
{CD}(x_a, x_b) = DISC(|q_{{\phi_y}}(z_{y_a}|x_a)-q_{{\phi_y}}(z_{y_b}|x_b)|).
\end{alignat}

\begin{figure}[!htb]
    \centering
    \includegraphics[width=.75\linewidth]{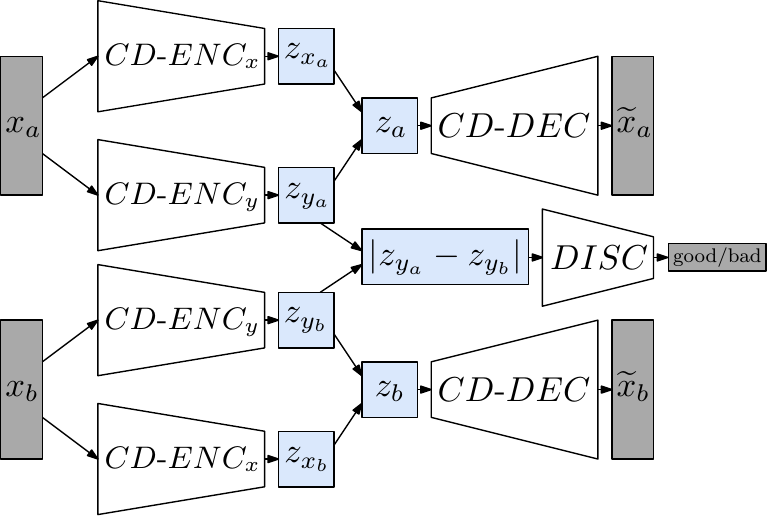}
    \caption{Components used during the change-discrimination pass of $CD$ (\ie the classifier components are omitted). The encoder and decoder distributions are written as $CD$-$ENC_y$, $CD$-$ENC_x$, and $CD$-$DEC$. We optimize the VAE objective and the change-discrimination objective. Note that the same encoder and decoder models are used for both datapoints.}
    \label{fig:cd}
\end{figure}

\begin{figure}[!htb]
    \centering
    \includegraphics[width=.75\linewidth]{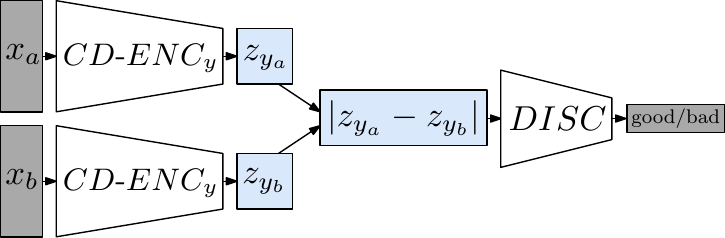}
    \caption{The components used when inferring a prediction from $CD$.}
    \label{fig:cd2}
\end{figure}

When training we minimize the binary cross-entropy between this prediction and change-quality labels $y_{cd}$. Simultaneously, we minimize the $ELBO$ for these samples, \ie the first 3 terms of the loss in \S3.1 of the main paper. The regular disentangled VAE objective is optimized in a separate pass, using regular datapoints and labels. An overview of the model is depicted in Fig.~\ref{fig:cd}. During inference, \ie while training VAE-CE, we only use the class-specific encoder and latent discriminator to make predictions. This submodel is depicted in Fig.~\ref{fig:cd2}.

\section{Interpolation-graph complexity}
To derive the size of the interpolation graph $\mathcal{G} = (V, E)$, we consider the graph as the discrete, dimension-wise interpolation of vector $a$ to vector $b$, where $|a| = |b| = n$. Nodes denote intermediate states in this transition, and each edge corresponds to a (set of) dimension(s) being changed from $a$ to $b$.

Each node in the graph considers one interpolation state, where for each dimension in this state, we pick the value from either $a$ or $b$. As such, there are $2^n$ possible combinations, giving us $|V| = 2^n$.

To derive the total number of edges, we reason about the number of edges that are attached to each node. Each node has an outgoing edge towards all nodes that have changed dimension values from $a_i$ to $b_i$, while not changing any such values from $b_i$ back to $a_i$. As such, given a node with $k$ dimensions already changed to $b_i$, we have $n-k$ dimensions that are yet to be changed. For each combination of $n-k$ changes we add an edge to the corresponding node, giving $2^{n-k}-1$ outgoing edges for a node with $k$ already-changed values (we subtract one as we do not consider the case where none of these $n-k$ values are changed). The number of nodes that exist for each value of $k$ is the number of combinations of size $k$ given $n$ values to change in total: ${n\choose k}$. The total number of edges can then be expressed by summing for all values of $k$. We denote and rewrite this sum as follows:
\begin{align}
    |E| &= \sum_{k=0}^{n} \Big[{n\choose k} (2^{n-k}-1)\Big]\\
    &= \sum_{k=0}^{n} \Big[{n\choose k} (2^{n-k})\Big] - \sum_{k=0}^{n} \Big[{n\choose k}\Big]\\
    &= 3^n - 2^n,\label{eq:last}
\end{align}
using the Binomial Theorem\cite{graham1989concrete} in step \ref{eq:last}. As such, we get $|E| = 3^n - 2^n$.
\section{Synthetic data generation}
Synthetic datapoints are generated by using the class-shapes, as defined in \S4.1 of the main paper, and adding noise. This process consists of two steps: (1) distorting the shape and (2) using this shape to create an image with varying stroke width.

\textbf{Shape distortion.} Shapes are distorted according to a randomly generated `vector field' that determines how much each piece of the shape will move. This field is created using a normalizing force and
3 to 6 points that force the field to flow in a certain direction. These point forces are placed at random positions, have a random orientation and a random weight $\sim\mathcal{U}_{[3, 6]}$. The direction of each point in the field is generated according to a weighted average of the forces. Each force's weight, for a given position, is computed as $w_p = w_f \cdot \frac{1}{d(p,f) + 1}$, where $w_f$ is the force's weight and $d(p,f)$ denotes the Euclidean distance between the position and the force.

\begin{figure}[!htb]
    \centering
    \subfloat[The input shape of class 0.\label{fig:exa}]{
    \hspace{.005\linewidth}
    \includegraphics[height=.25\linewidth]{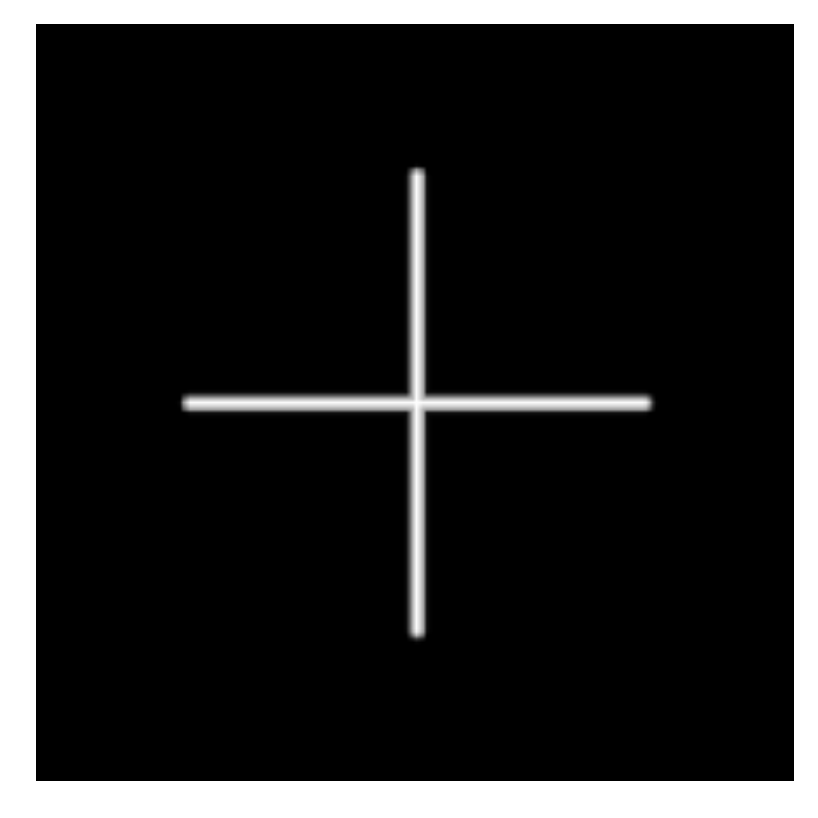}
    \hspace{.005\linewidth}
    }
        \hspace{.005\linewidth}
    \subfloat[The vector field, where each red dot depicts a `force'.\label{fig:exb}]{
    \hspace{.025\linewidth}
    \includegraphics[height=.25\linewidth]{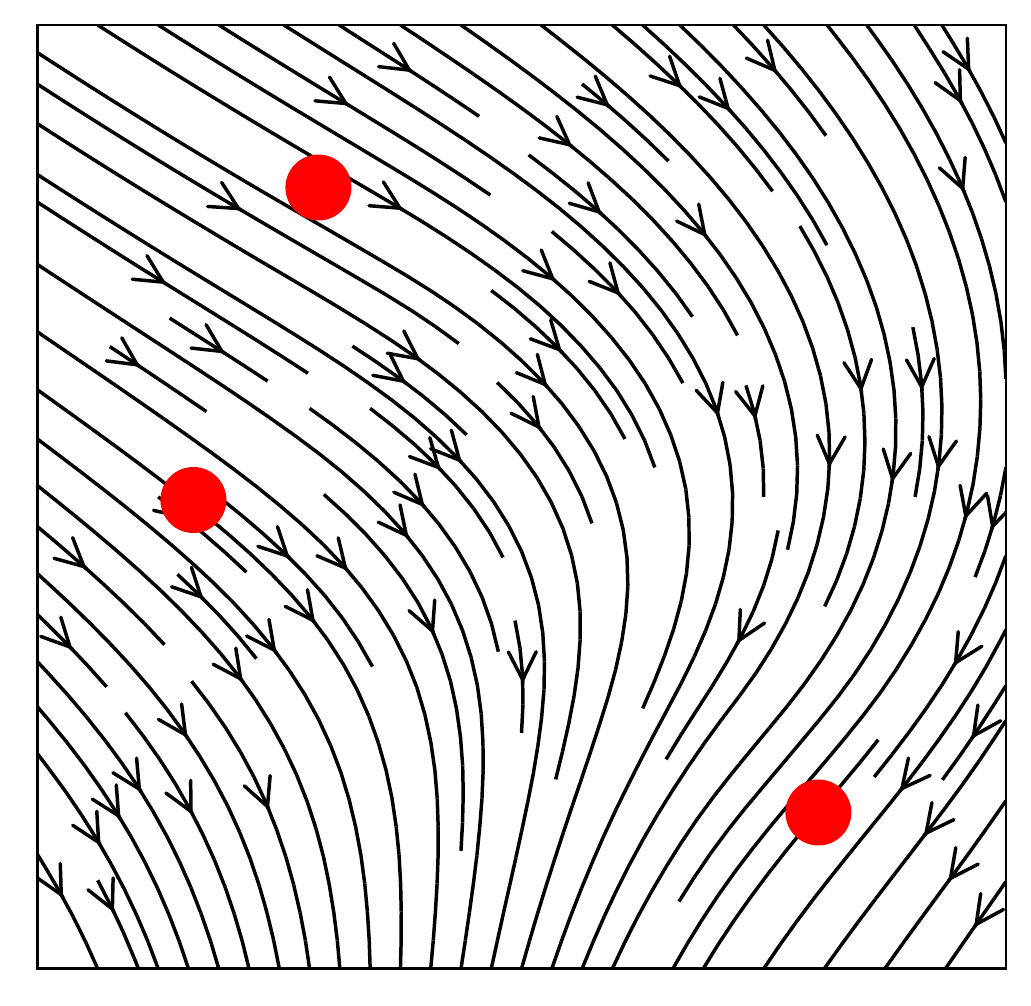}
    \hspace{.025\linewidth}
    }%
        \hspace{.015\linewidth}
    \subfloat[The final shape, distorted, rotated and normalized.\label{fig:exc}]{
    \hspace{.005\linewidth}
    \includegraphics[height=.25\linewidth]{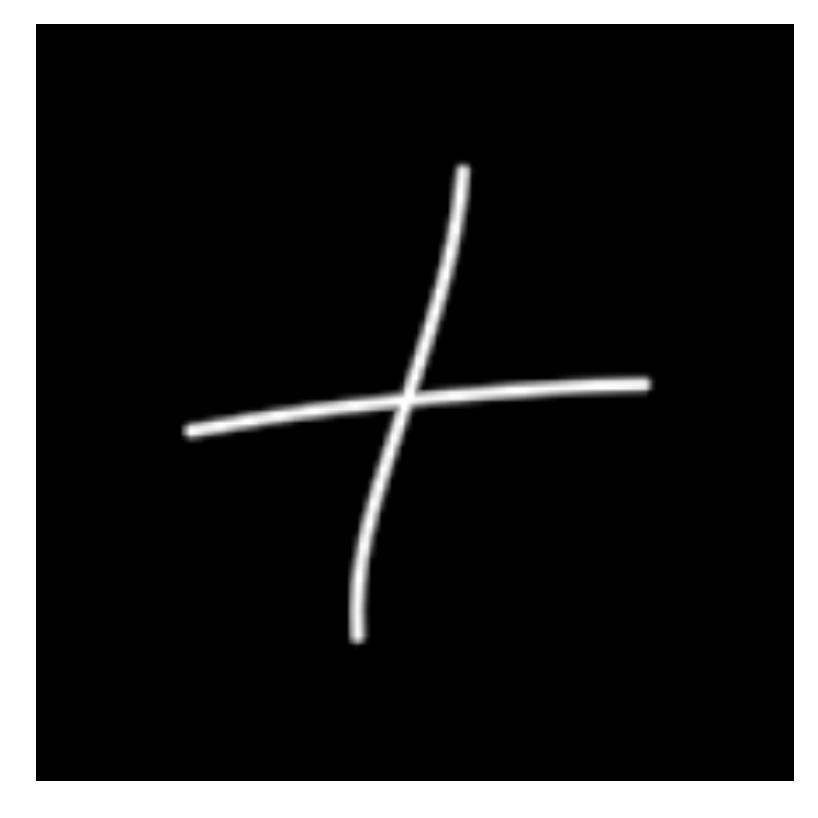}
    \hspace{.005\linewidth}
    }%
    \caption{The distortion step of the data generation process.}%
    \label{fig:exabc}%
\end{figure}

The shape is distorted by computing the offset of the shape's pieces according to this vector field, scaled according to an image-level weight $\sim\mathcal{U}_{[0.3, 0.6]}$. In practice, each line consists of a subdivision of 20 evenly-spaced line segments, of which the start- and end-coordinates are transformed. The transformed shape is additionally rotated $\sim \mathcal{N}{(0, \frac{\pi}{40}})$ degrees. The new shape is centered and normalized to keep the size and position of the data constant. The entire distortion process is depicted in Fig.~\ref{fig:exabc}.

\textbf{Stroke-width variation.} Next, we generate the final datapoint using the distorted shape. Rather than applying a uniform line width, we vary the width throughout the image. This variation is produced by generating a `stroke-thickness field' that determines the line width, as a proportion of the total image size, at any given point. This field is generated similarly to the vector field. We sample a base stroke thickness $\sim\mathcal{U}_{[0.001, 0.05]}$ and create 3 to 5 random points depicting a stroke-thickness force. Each of these points comes with a stroke thickness $\sim\mathcal{U}_{[0.001, 0.05]}$, weight $\sim\mathcal{U}_{[3, 6]}$, and decay exponent $\in \{2, 3, 4\}$. The stroke thickness at each point in the image is determined by a weighted average of the base thickness and the thickness of each force. Each force's weight, for a given position, is computed as $w_p = w_f \cdot e^{-exp \cdot d(p, f)}$, where $w_f$ depicts the weight parameter, $exp$ the decay exponent, and $d(p, f)$ the Euclidean distance between the position and the force.

\begin{figure}[!hbt]
    \centering
    \subfloat[The shape to draw.\label{fig:ex2a}]{
    \hspace{.005\linewidth}
    \includegraphics[height=.25\linewidth]{fig/c2b.pdf}
    \hspace{.005\linewidth}
    }
       \hspace{.005\linewidth}
    \subfloat[The stroke-thickness field, where each blue dot depicts a `force'.\label{fig:ex2b}]{
    \hspace{.0075\linewidth}
    \includegraphics[height=.25\linewidth]{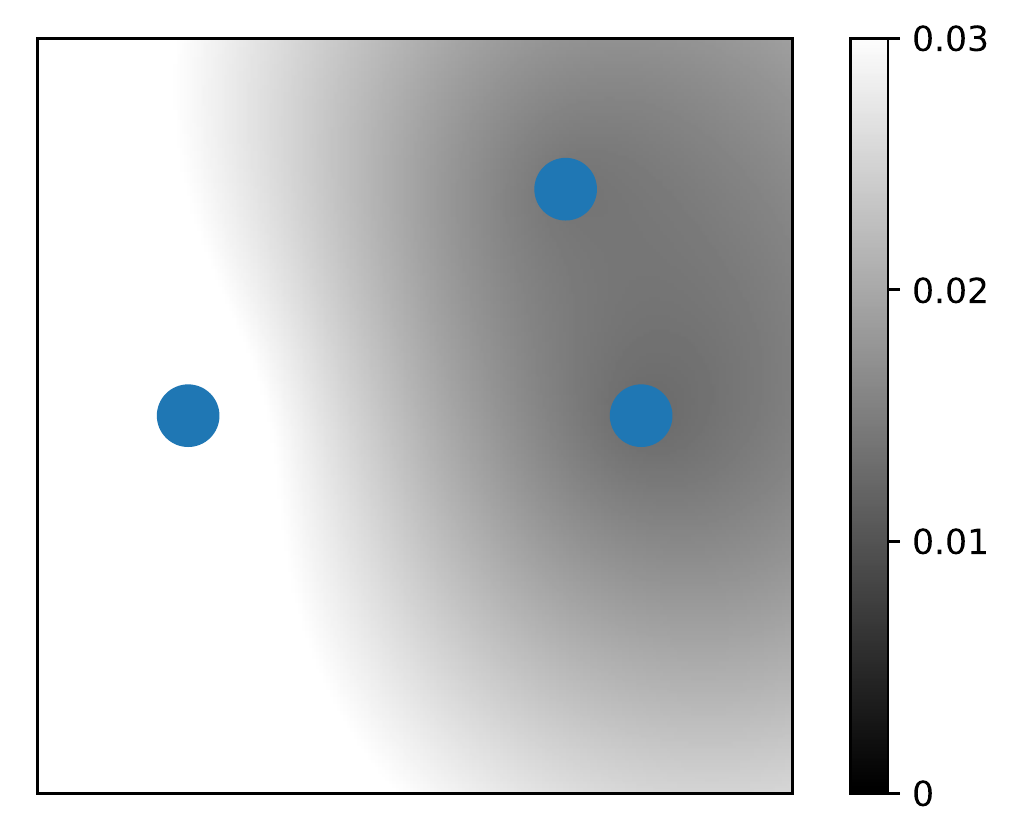}
    \hspace{.0075\linewidth}
    }%
        \hspace{.0075\linewidth}
    \subfloat[The final datapoint.\label{fig:ex2c}]{
    \hspace{.005\linewidth}
    \includegraphics[height=.25\linewidth]{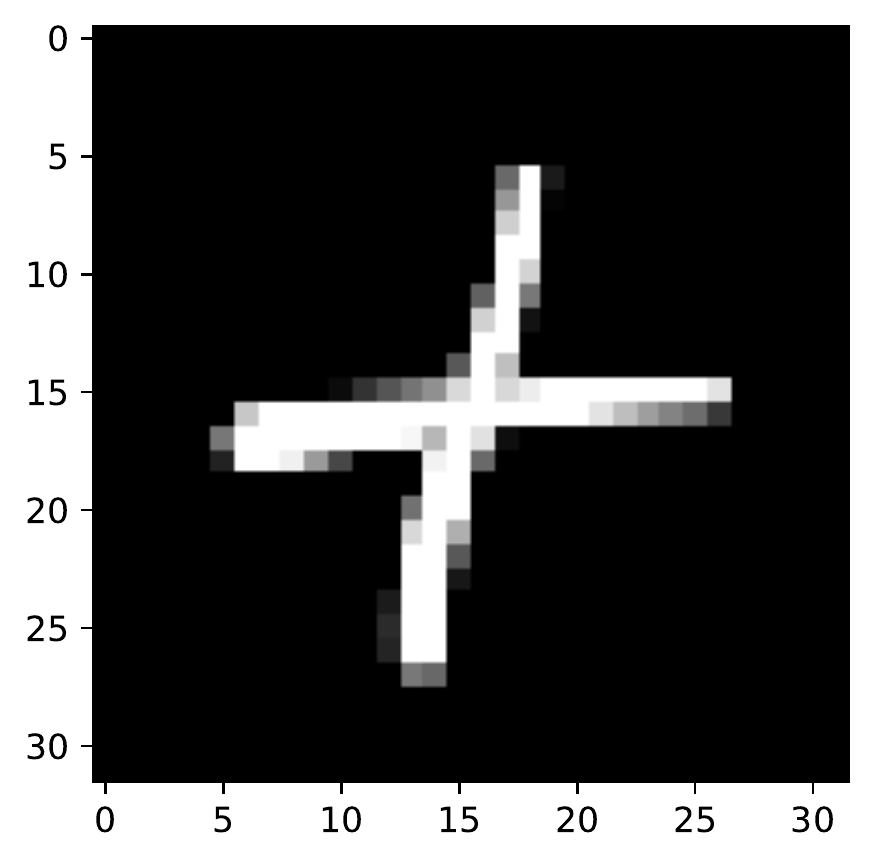}
    \hspace{.005\linewidth}
    }%
    \caption{Creating the final datapoint by using a stroke-thickness field.}%
    \label{fig:ex2abc}%
\end{figure}

The final image is created by computing the distance of each pixel to its nearest line and checking with the stroke-field whether the pixel should be drawn as a line or not. On the edge of lines (in a constant range of 0.04) we linearly interpolate between white and black in order to create smooth edges. Following this process we can generate images of arbitrary resolution; an example of generating a 32x32 image is depicted in Fig.~\ref{fig:ex2abc}.

\section{MNIST line-augmentation}
MNIST datapoints are transformed into individual lines in order to be able to create change pairs. By hiding (sub)sets of individual lines we can create pairs differing in a single line. We consider a line to be a smooth curve, which we define as a sequence of points where the angle of the consecutive points (\wrt a horizontal line) does not change faster than some predefined limit. The process works in three stages: (1) thinning images, (2) identifying starting points of lines and (3) generating lines.

\begin{figure}[!htb]
    \centering
    \subfloat[Input image.\label{fig:exmna}]{
    \hspace{.02\linewidth}
    \includegraphics[height=.25\linewidth]{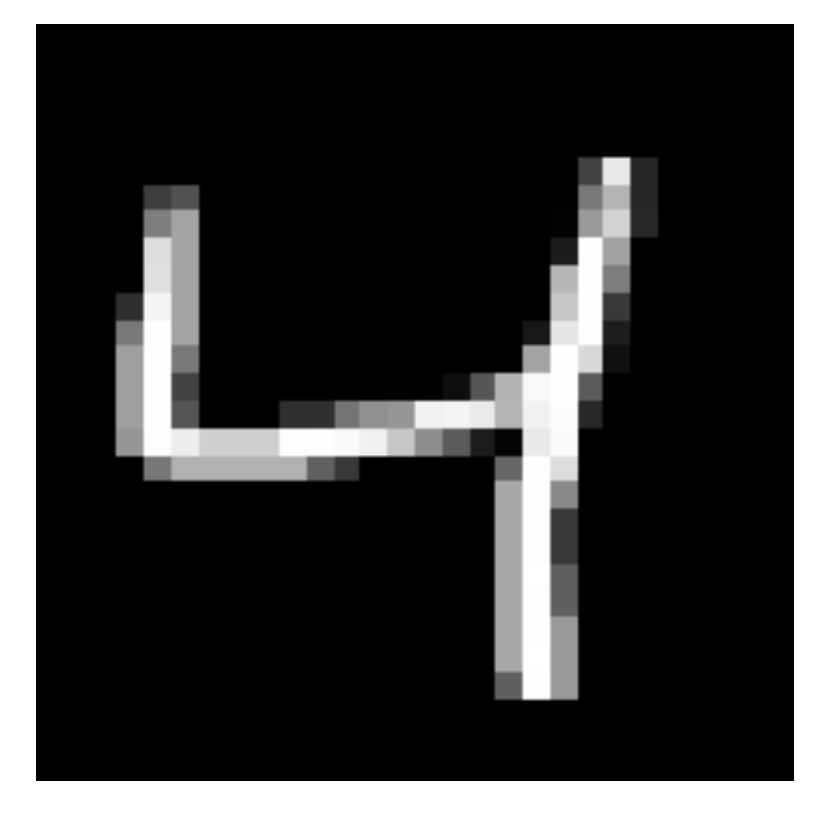}
    \hspace{.02\linewidth}
    }
    \subfloat[Thresholded image.\label{fig:exmnb}]{
    \hspace{.02\linewidth}
    \includegraphics[height=.25\linewidth]{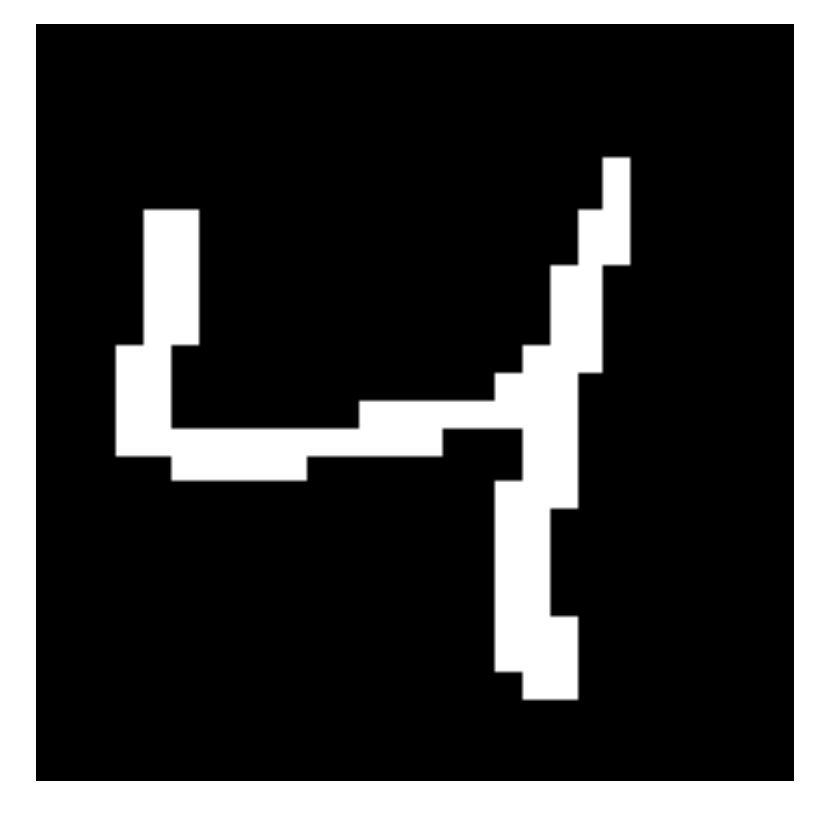}
    \hspace{.02\linewidth}
    }
    \subfloat[Thinned image.\label{fig:exmnc}]{
    \hspace{.02\linewidth}
    \includegraphics[height=.25\linewidth]{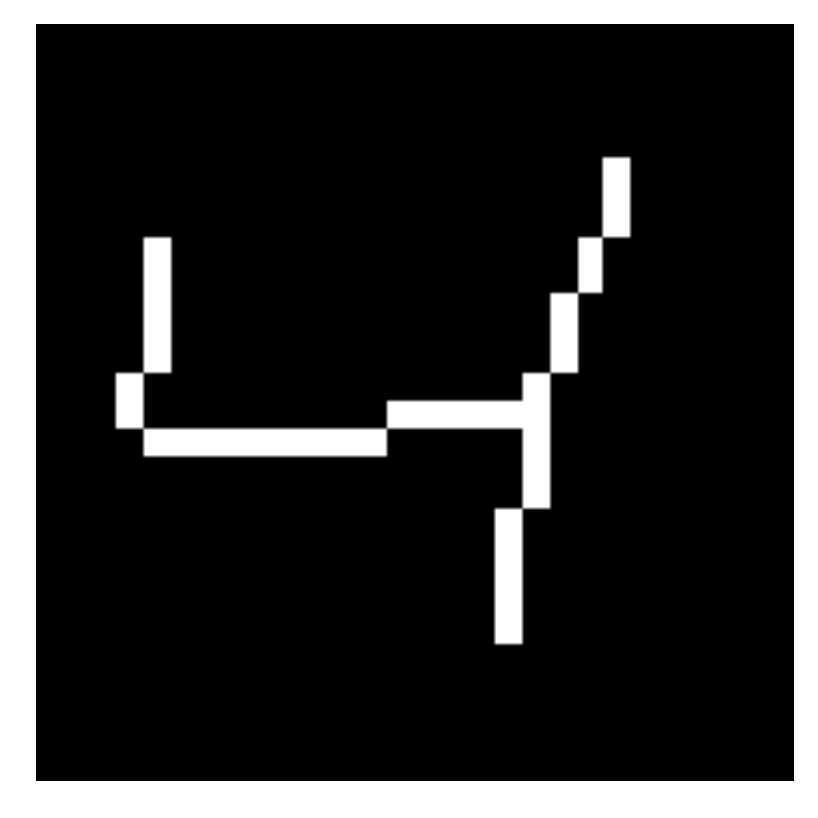}
    \hspace{.02\linewidth}
    }
    \caption{The thinning steps of augmenting MNIST-digits.}%
    \label{fig:exmnabc}%
\end{figure}

\textbf{Thinning images.} To reduce an image to thin lines we follow the same procedure as \cite{de2016incremental}. This procedure considers first thresholding the image, \ie marking all pixels above a specific value. The threshold is found by starting at 0 and increasing it iteratively until one of three conditions is reached: (1) the number of remaining (white) pixels drops under 50\%, or (2) the number of 4-connected or 8-connected components changes, or (3) the threshold reaches 250 (given a pixel-range of $[0, 255]$). This (binary) image is thinned using the Zhang-Suen thinning algorithm\cite{zhang1984fast}. An example of the thinning process is depicted in Fig.~\ref{fig:exmnabc}.

\begin{figure}[!htb]
    \centering
    \includegraphics[height=.25\linewidth]{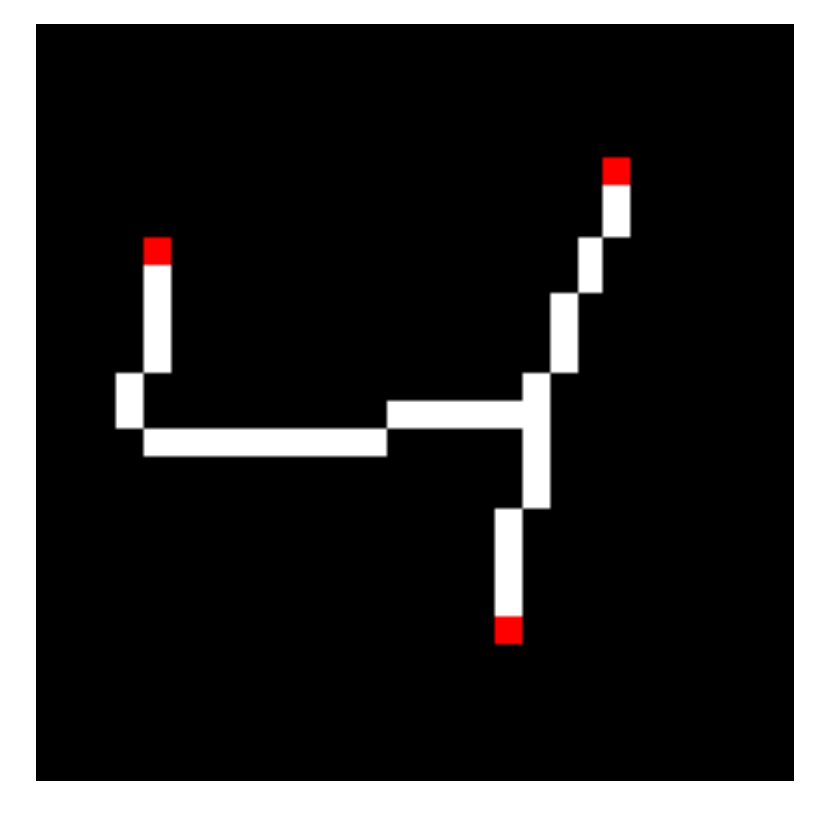}
    \caption{The starting pixels of the thinned image, marked as red pixels.}
    \label{fig:exmd}
\end{figure}

\textbf{Identifying starting points.} Lines are created by grouping neighboring white pixels together, where we consider 8-connected pixels as neighbors. To create lines from this representation, we first identify starting points and iteratively construct lines. We denote a starting pixel as a pixel with only one neighbor, where this neighboring pixel is connected to both the starting pixel and one other pixel. The first condition marks pixels likely to be the start point of a line, whereas the latter ensures that the starting pixel is not an `outlier' pixel connected to a pixel in the middle of a line. If no such starting pixels can be found, the condition is relaxed to only the first condition. If still no starting pixels can be identified, we mark all pixels with the maximum number of neighbors as starting pixels (as these pixels likely lie on intersections of lines). The starting pixels for the example 4-image are depicted in Fig.~\ref{fig:exmd}.

\begin{figure}[!htb]
    \centering
    \includegraphics[width=1\linewidth]{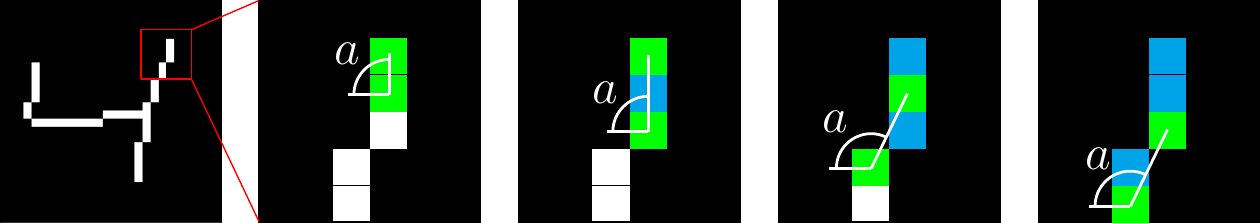}
    \caption{The line-creation process. We iteratively add neighbors such that angle $a$ changes the least. Green pixels indicate the pixels currently determining the line angle, whereas blue pixels indicate the remaining pixels that were added to the line.}
    \label{fig:exmline}
\end{figure}

\textbf{Line generation.} Lines are identified by starting at an arbitrary starting pixel and iteratively adding pixels such that the overall line does not curve faster than a set limit ($\frac{\pi}{4}$, in our case). First, an arbitrary neighbor of the starting pixel is added to the line. The angle between these two points (and a horizontal line) determines the line's current angle. Then, points are iteratively added by checking the last-added point's neighbors and adding the neighbor such that the angle of the line changes the least (ignoring already-added neighbors). The line angle is continuously updated according to the last 3 points added. When no neighboring points can be found such that the angle change does not exceed our chosen limit, the line is complete. The line-creation process is depicted in Fig.~\ref{fig:exmline}.

\begin{figure}[!htb]
    \centering
    \subfloat[The identified starting points.\label{fig:exmfa}]{
    \hspace{.01\linewidth}
    \includegraphics[height=.25\linewidth]{fig/m3.pdf}
    \hspace{.015\linewidth}
    }
        \hspace{.01\linewidth}
    \subfloat[The identified lines. \label{fig:exmfb}]{
    \hspace{.015\linewidth}
    \includegraphics[height=.25\linewidth]{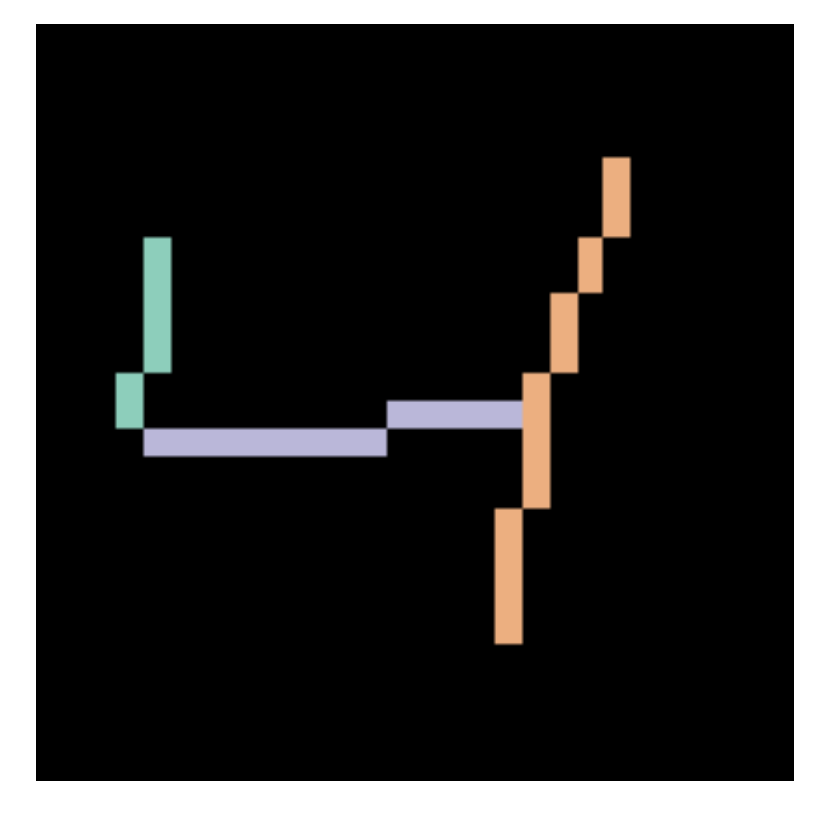}
    \hspace{.015\linewidth}
    }%
        \hspace{.01\linewidth}
    \subfloat[The final line split. \label{fig:exmfc}]{
    \hspace{.015\linewidth}
    \includegraphics[height=.25\linewidth]{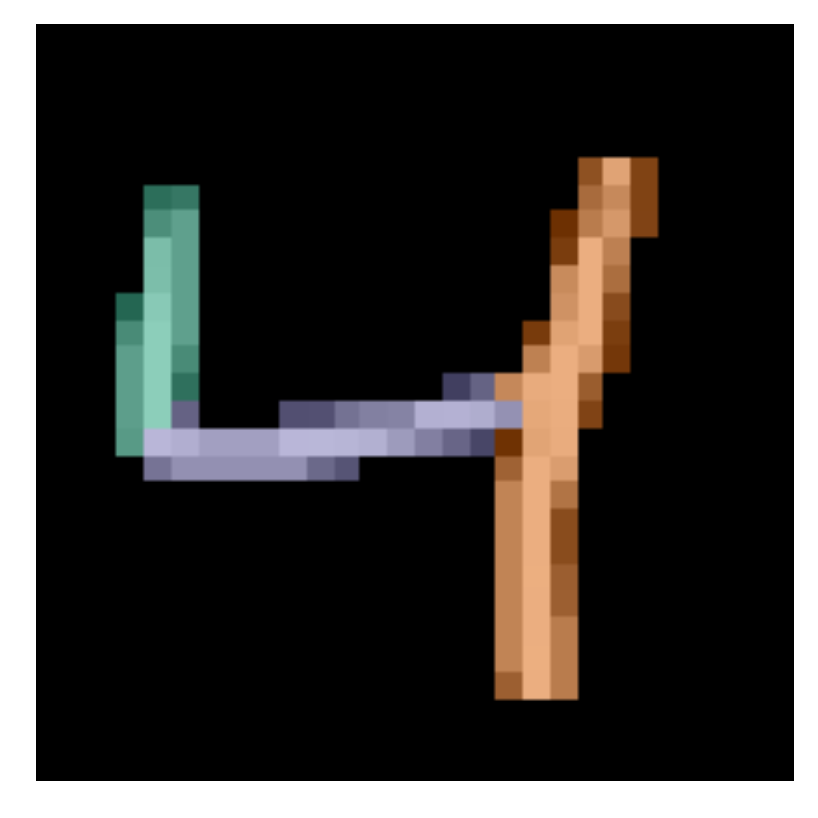}
    \hspace{.01\linewidth}
    }%
    \caption{The line-generation step of augmenting MNIST-digits, where each color indicates an identified line.}%
    \label{fig:exmfabc}%
\end{figure}

This process is repeated for all starting pixels. If we still have unmapped pixels, we create the remaining lines by starting from arbitrary points that neighbor already-found lines. Finally, outlier points (pixels with no neighbors) are added to their closest line. As the entire process can be noisy, it is possible that insignificant `lines' were identified. As a post-processing step, we add these small lines ($\leq$ 3 pixels) to their largest neighboring line (if a line is surrounded by only small lines, we add it to its neighbor-lines' neighbor, and so on). To conclude, we create the line split of the source digit. Pixels of the (unmodified) datapoint are mapped to lines by assigning them to the closest line (the line with the minimal Euclidean distance between the to-be-mapped pixel and the closest pixel on the line). The result of this line-generation process is depicted in Fig.~\ref{fig:exmfabc}.

\section{Model architecture}
We first provide shared architectures, after which we provide method-specific architectures. All regular models use 8 latent dimensions per subspace, $CD$ uses 16 latent dimensions per subspace. We denote (transposed) 2D convolutional layers as `\textit{(T)Conv2D(kernel size, stride, filters)}' and use `SAME' padding for all such layers. Fully connected layers are denoted as `\textit{FC(output size)}'. All models are implemented using Tensorflow\cite{tensorflow2015whitepaper}.

\subsection{Shared components}
\begin{table}[H]
\begin{center}
\begin{tabular}{l l l}\hline
 & \textbf{Layer} & \textbf{Activation}\\ \hline
$x$&Input $(32 \times 32 \times 1)$ & \\
1&Conv2D($4\times4$, 2, 32) & LReLU(0.1)\\
2&Batch Normalization & \\
3&
Conv2D($4\times4$, 1, 64)
& LReLU(0.1)  \\
4&Batch Normalization &  \\
5&
Conv2D($4\times4$, 1, 128) & LReLU(0.1)  \\
6&Batch Normalization & \\
$\mu_y/\mu_x$&FC(8) from layer 6 &  \\
$\sigma_y/\sigma_x$&FC(8) from layer 6 &  \\
\hline
\end{tabular}
\end{center}
\caption{The encoders, $q_{\phi{_y}}(z_y|x)$ and $q_{\phi{_x}}(z_x|x)$.}
\end{table}

\begin{table}[H]
\begin{center}
\begin{tabular}{l l l}\hline
 & \textbf{Layer} & \textbf{Activation}\\ \hline
$z$&Input (16)& \\
1&FC(32768) & LReLU(0.1) \\
2&Batch Normalization &  \\
3&TConv2D($4\times4$, 1, 64) & LReLU(0.1)  \\
4&Batch Normalization &  \\
5&TConv2D($4\times4$, 1, 32) & LReLU(0.1)  \\
6&Batch Normalization & \\
$\widetilde{x}$&TConv2D($4\times4$, 2, 1) & Sigmoid \\
\hline
\end{tabular}
\end{center}
\caption{The decoder, $p_\theta(x|z_y,z_x)$.}
\end{table}

\begin{table}[H]
\begin{center}
\begin{tabular}{l l l}\hline
 & \textbf{Layer} & \textbf{Activation}\\ \hline
$z_y$&Input (8)& \\
$\widetilde{y}$&FC(10) & Softmax \\
\hline
\end{tabular}
\end{center}
\caption{The label classifier, $q_{\psi{_y}}(y|z_y)$.}
\end{table}

\begin{table}[H]
\begin{center}
\begin{tabular}{l l l} \hline
 & \textbf{Layer} & \textbf{Activation}\\ \hline
$z_x$&Input (8)& \\
$1$&FC(50) & LReLU(0.1) \\
$2$&Batch Normalization & \\
$\widetilde{y}$&FC(10) & Softmax \\
\hline
\end{tabular}
\end{center}
\caption{The adverse label classifier, $q_{\psi{_x}}(y|z_x)$.}
\end{table}
\subsection{LVAE}

\begin{table}[H]
\begin{center}
\begin{tabular}{l l l}\hline
 & \textbf{Layer} & \textbf{Activation} \\ \hline
$z_{y_i}$&Input (1)& \\
$\widetilde{d}_i$&FC(2) & Softmax \\
\hline
\end{tabular}
\end{center}
\caption{The dimension-label classifiers.}
\end{table}

\begin{table}[H]
\begin{center}
\begin{tabular}{l l l}\hline
 & \textbf{Layer} & \textbf{Activation} \\ \hline
$z_{y_{ic}}$&Input (7)&\\
$1$&FC(50) & LReLU(0.1)\\
$2$&Batch Normalization & \\
$\widetilde{d}_i$&FC(2) & Softmax \\
\hline
\end{tabular}
\end{center}
\caption{The complementary dimension-label classifiers.}
\end{table}

\subsection{VAE-CE}
\begin{table}[H]
\begin{center}
\begin{tabular}{l l l}\hline
 & \textbf{Layer} & \textbf{Activation} \\ \hline
$x$&Input $(32 \times 32 \times 1)$ & \\
1&
Conv2D($4\times4$, 2, 32)
& LReLU(0.1)  \\
2&Batch Normalization & \\
3&Dropout(0.3) & \\
4&
Conv2D($4\times4$, 1, 64)
& LReLU(0.1)  \\
5&Batch Normalization &  \\
6&Dropout(0.3) &  \\
7&
Conv2D($4\times4$, 1, 128)
& LReLU(0.1)  \\
8&Batch Normalization &  \\
9&Dropout(0.3) &  \\
$real$&FC(2) & Softmax  \\
\hline
\end{tabular}
\end{center}
\caption{The realism discriminator, $D$.}
\end{table}
\subsection{CD}

\begin{table}[H]
\begin{center}
\begin{tabular}{l l l}\hline
 & \textbf{Layer} & \textbf{Activation} \\ \hline
$x$&Input $(32 \times 32 \times 1)$ & \\

1& Conv2D($4\times4$, 2, 32) & LReLU(0.1)  \\
2&Batch Normalization & \\
3&
Conv2D($4\times4$, 1, 128)
& LReLU(0.1)  \\
4&Batch Normalization &  \\
$\mu_y/\mu_x$&FC(16) from layer 4 &  \\
$\sigma_y/\sigma_x$&FC(16) from layer 4 &  \\
\hline
\end{tabular}
\end{center}
\caption{$CD$'s encoders.}
\end{table}
\begin{table}[H]
\begin{center}
\begin{tabular}{l l l}\hline
 & \textbf{Layer} & \textbf{Activation}\\ \hline
$z$&Input (32)&\\
1&FC(32768) & LReLU(0.1) \\
2&Batch Normalization & \\

3&
TConv2D($4\times4$, 1, 32)
& LReLU(0.1) \\
4&Batch Normalization &\\
$\widetilde{x}$&
TConv2D($4\times4$, 2, 1)
& Sigmoid\\
\hline
\end{tabular}
\end{center}
\caption{$CD$'s decoder.}
\end{table}
\begin{table}[H]
\begin{center}
\begin{tabular}{l l l}\hline
 & \textbf{Layer} & \textbf{Activation} \\ \hline
$z_y$ or $z_x$&Input (16)&\\
$1$&FC(50) & LReLU(0.1)\\
$2$&Batch Normalization & \\
$\widetilde{y}$&FC(10) & Softmax \\
\hline
\end{tabular}
\end{center}
\caption{$CD$'s label-disentanglement classifiers.}
\end{table}
\begin{table}[H]
\begin{center}
\begin{tabular}{l l l}\hline
 & \textbf{Layer} & \textbf{Activation}\\ \hline
$|z_{y_a} - z_{y_b}|$&Input (16)& \\
$1$&FC(50) & LReLU(0.1) \\
$2$&Batch Normalization & \\
$3$&Dropout(0.3) & \\
$change$&FC(2) & Softmax  \\
\hline
\end{tabular}
\end{center}
\caption{The latent-change discriminator, $DISC$.}
\end{table}

\section{Hyperparameters and training}
\subsection{Loss functions}
First, we briefly restate the loss functions. All methods extend DVAE: We sum the DVAE loss and the method-specific loss. Note that $CD$ also extends DVAE.

\textbf{DVAE}'s loss denotes the class-disentangled VAE optimization, as described in \S{3.1} of the main paper:
\begin{alignat}{1}
\loss_{\theta,{\phi{_y}},{\phi{_x}},{\psi{_y}}}&(x, y) = \beta_y KL(q_{{\phi{_y}}}(z_y|x)||p_\theta(z))\label{eq:kl0}\\
&+\beta_x KL(q_{{\phi{_x}}}(z_x|x)||p_\theta(z))\label{eq:kl1}\\
&-\mathbb{E}_{q_{{\phi{_y}}}(z_y|x),q_{{\phi{_x}}}(z_x|x)}[\log p_\theta(x|z_y,z_x)]\label{eq:rec}\\
&-\alpha \mathbb{E}_{q_{\phi{_y}}(z_y|x)}[\log(q_{{\psi{_y}}}(y|z_y))]\label{eq:cl0}\\
&+\alpha \mathbb{E}_{q_{\phi{_x}}(z_x|x)}[\log(q_{{\psi{_x}}}(y|z_x))],\label{eq:cl1}\\
&\hspace{-.576cm}\loss_{\psi{_x}}(x, y) = -\mathbb{E}_{q_{\phi{_x}}(z_x|x)}[\log(q_{{\psi{_x}}}(y|z_x))],\label{eq:cl1_adv}
\end{alignat}

\textbf{LVAE}'s loss optimizes individual dimensions using auxiliary classifiers. We denote dimension labels as $d_i$, individual dimensions as $z_{y_i}$. and the complementary dimensions (all dimensions but $i$) as $z_{y_{ci}}$. We use a classifier for each label $i$, denoted as categorical distribution $q_{\psi_{di}}(d_i|z_{y_i})$, and an adversarial classifier for the complementing dimensions $ci$, denoted as categorical distribution $q_{\psi_{dci}}(d_i|z_{y_{ci}})$. $n_c$ denotes the number of concepts. The extra loss terms can be denoted as follows:
\begin{alignat}{1}
\loss_{\theta,\phi{_y},\psi_{di}}(x, y, d_i)&= -\alpha_d {\log(q_{\psi_{di}}(d_i|z_{y_i}))} \\
&+ \alpha_d \log(q_{\psi_{dci}}(d_i|z_{y_{ci}})), \\
\loss_{\psi_{dci}} (x, y, d_i)&= -\log(q_{\psi_{dci}}(d_i|z_{y_{ci}})),\\
\loss_{LVAE}(x, y, d) &= \sum_{i}^{n_c} \big(\loss_{\theta,\phi{_y},\psi_{di}}(x, y, d_i) \\
&\hspace{.809cm} + \loss_{\psi_{dci}} (x, y, d_i)\big).
\end{alignat}

\textbf{VAE-CE}'s loss considers the pair-based dimension conditioning procedure as described in \S3.2 of the main paper. We create samples $\widetilde{x}_{p_a}$ and $\widetilde{x}_{p_b}$ (from datapoints that are used in the DVAE objective) and optimize the main VAE using $CD$ and $D$. Additionally, we train $D$ to distinguish between real/fake datapoints as a binary classification task, using datapoint-label pairs ($x$, $y_{d}$). We either use the synthesized datapoints and a 0-label, or training datapoints and a 1-label. The loss can be denoted as follows:
\begin{alignat}{1}
\hspace{0cm}\loss_{\theta,{\phi{_y}},{\phi{_x}}}(\widetilde{x}_{p_a}, \widetilde{x}_{p_b}) &=  {-}\alpha_r\log(D(\widetilde{x}_{p_a}))\\ &\hspace{0.465cm}{-}\alpha_r\log(D(\widetilde{x}_{p_b}))\\
&\hspace{-1.6cm}+ \alpha_{p} n_y\frac{|z_{p_a}-z_{p_b}|}{|z_{y_a}-z_{y_b}|} \cdot - \log(CD(\widetilde{x}_{p_a}, \widetilde{x}_{p_b})),\\
\loss_{D}(x, y_d) &= -\log(D({y}_{d}|x)).
\end{alignat}

\textbf{GVAE} and \textbf{ADA-GVAE} are optimized using the $ELBO$, \ie equations (\ref{eq:kl0}), (\ref{eq:kl1}), and (\ref{eq:rec}). We use specific pairings of datapoints and average out dimensions. Since these datapoints might not have class labels, we compute their loss \wrt the $ELBO$ in a separate pass.

\textbf{CD}'s loss optimizes the change-discrimination objective. We denote the change-quality label as $y_{cd}$, and infer a change-quality prediction of a pair $(x_a, x_b)$. The resulting loss can be denoted as follows:
\begin{alignat}{1}
&\hspace{-.16cm}{CD}(x_a, x_b) = DISC(|q_{{\phi_y}}(z_{y_a}|x_a)-q_{{\phi_y}}(z_{y_b}|x_b)|), \\
&\hspace{-.16cm}\loss_{\phi{_y},DISC}(x_a, x_b, y_{cd}) = -\alpha_c\log(CD(y_{cd}|x_a, x_b)).
\end{alignat}

Finally, we note that for models using multiple training passes with uneven numbers of datapoints (GVAE, ADA-GVAE, and CD), we scale the respective losses by this ratio in order to balance the different passes.

\subsection{Hyperparameter optimization}
The settings shared between all training procedures are depicted in Table~\ref{ap:hpsh}. We tune the hyperparameters as follows. First, we identify a non-degenerate solution by hand, defined as a solution where none of the objectives are ignored (\ie no collapsed latent spaces, uninformative classifiers, or all-zero outputs). Next, we define a range of parameters around this solution, and explore all configurations in this range. The hyperparameters are provided in Table~\ref{ap:hp}, whereas their explored values are denoted in Table~\ref{ap:hpv}. 

We use the $eac$ for model selection, using 90 ($a, b$) pairs (note that these pairs are distinct from the pairs used for the final results). We generate interpolations using all methods ($sm$, $dim$, and $graph$ for VAE-CE) and take minimum $eac$ out of these. As computing the $eac$ requires access to the ground-truth generating process, we search for hyperparameters using the synthetic data. The identified parameters are also used for the MNIST models.

Each model is trained for $\approx$ \num{2000000} steps: 20 epochs on the synthetic dataset and 33 epochs on MNIST. We found that training models with fewer steps generally resulted in worse $eac$-values, whereas training for significantly longer did not improve (and sometimes regressed) the $eac$-score. 

\begin{figure*}
    \centering
    \subfloat[DVAE\label{fig:advae}]{\includegraphics[height=.32\textwidth]{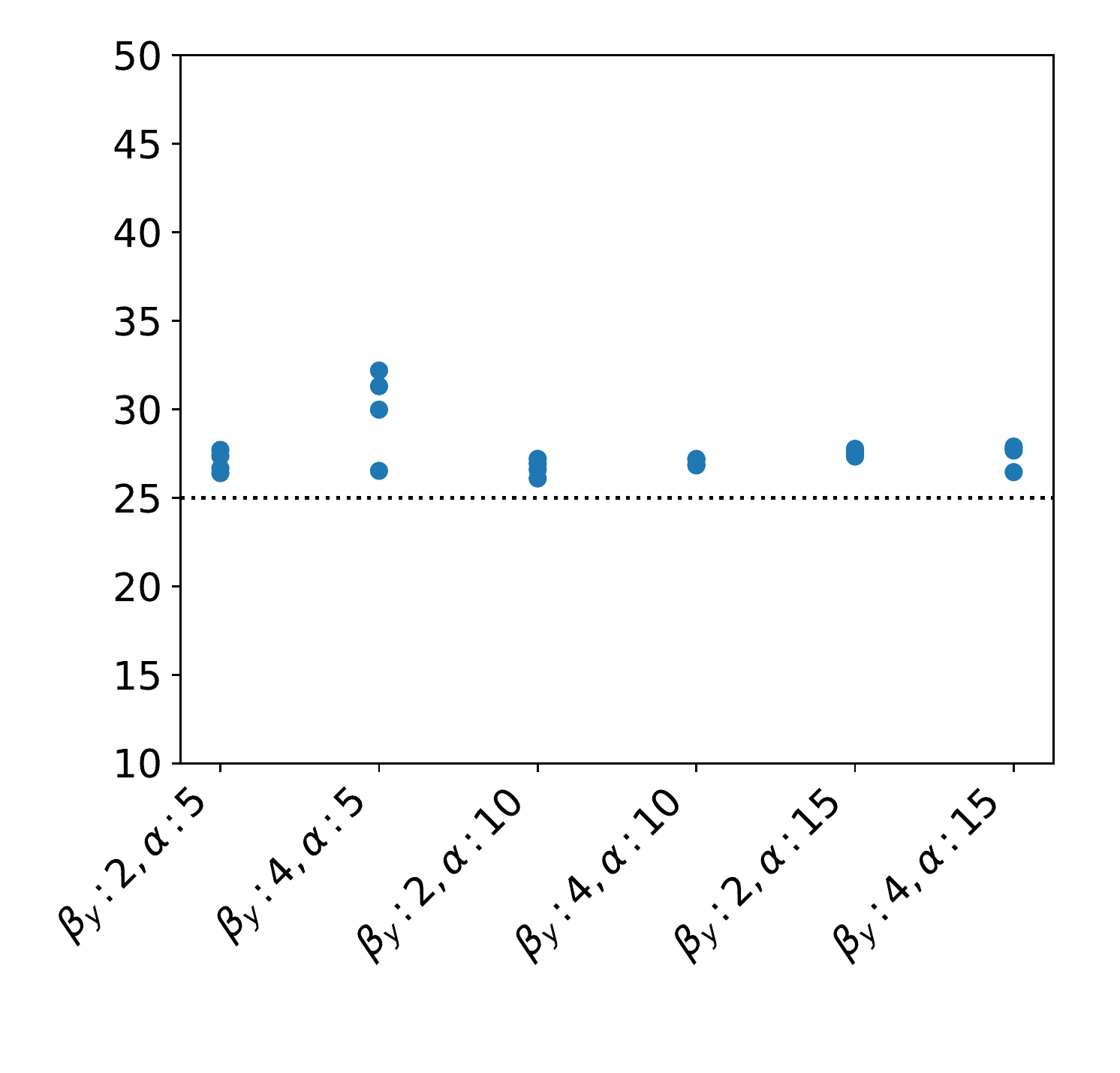}}
    \subfloat[LVAE\label{fig:alvae}]{\includegraphics[height=.32\textwidth]{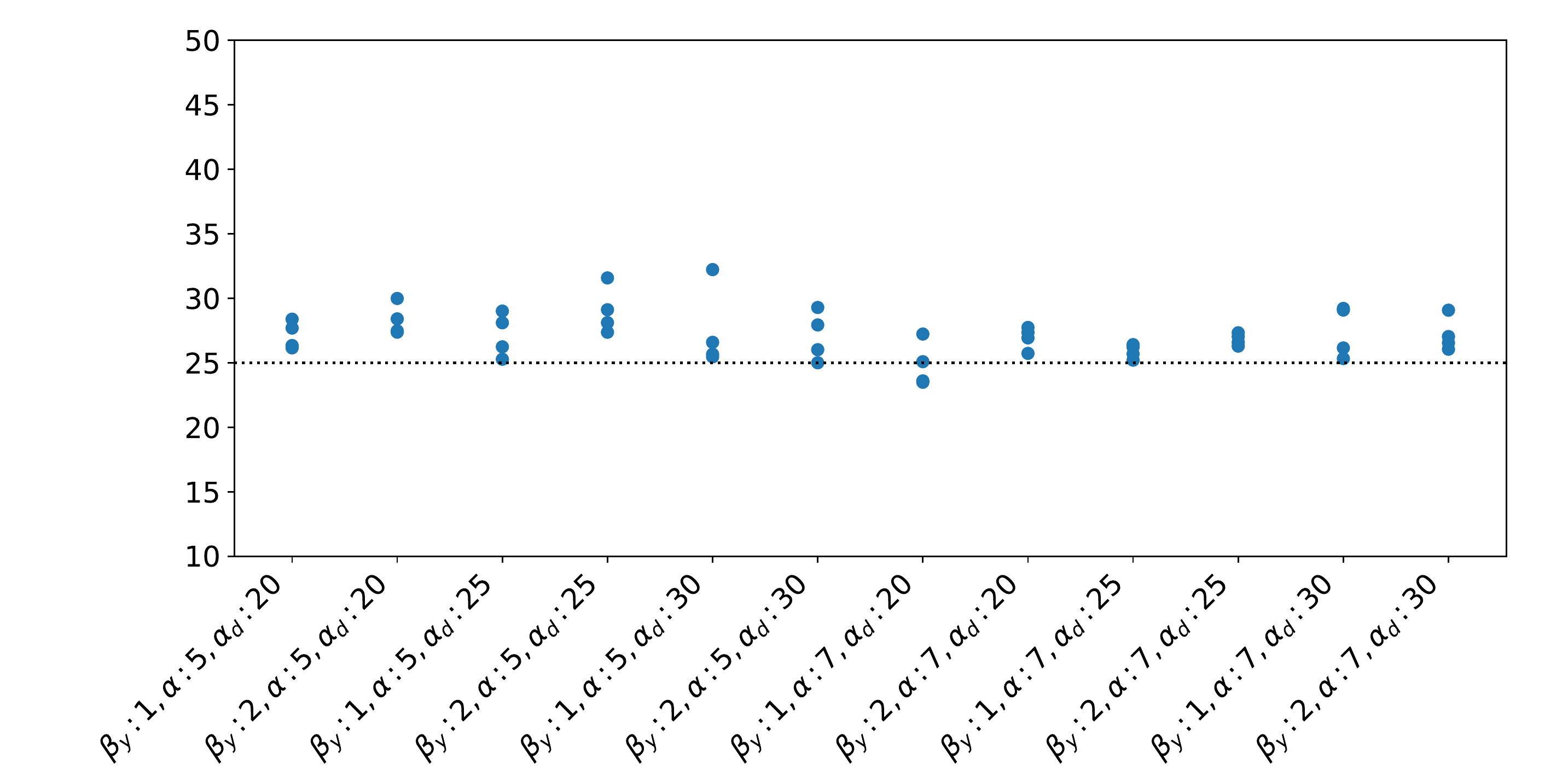}}
    \\
    \centering
    \subfloat[GVAE\label{fig:agvae}]{\includegraphics[height=.32\textwidth]{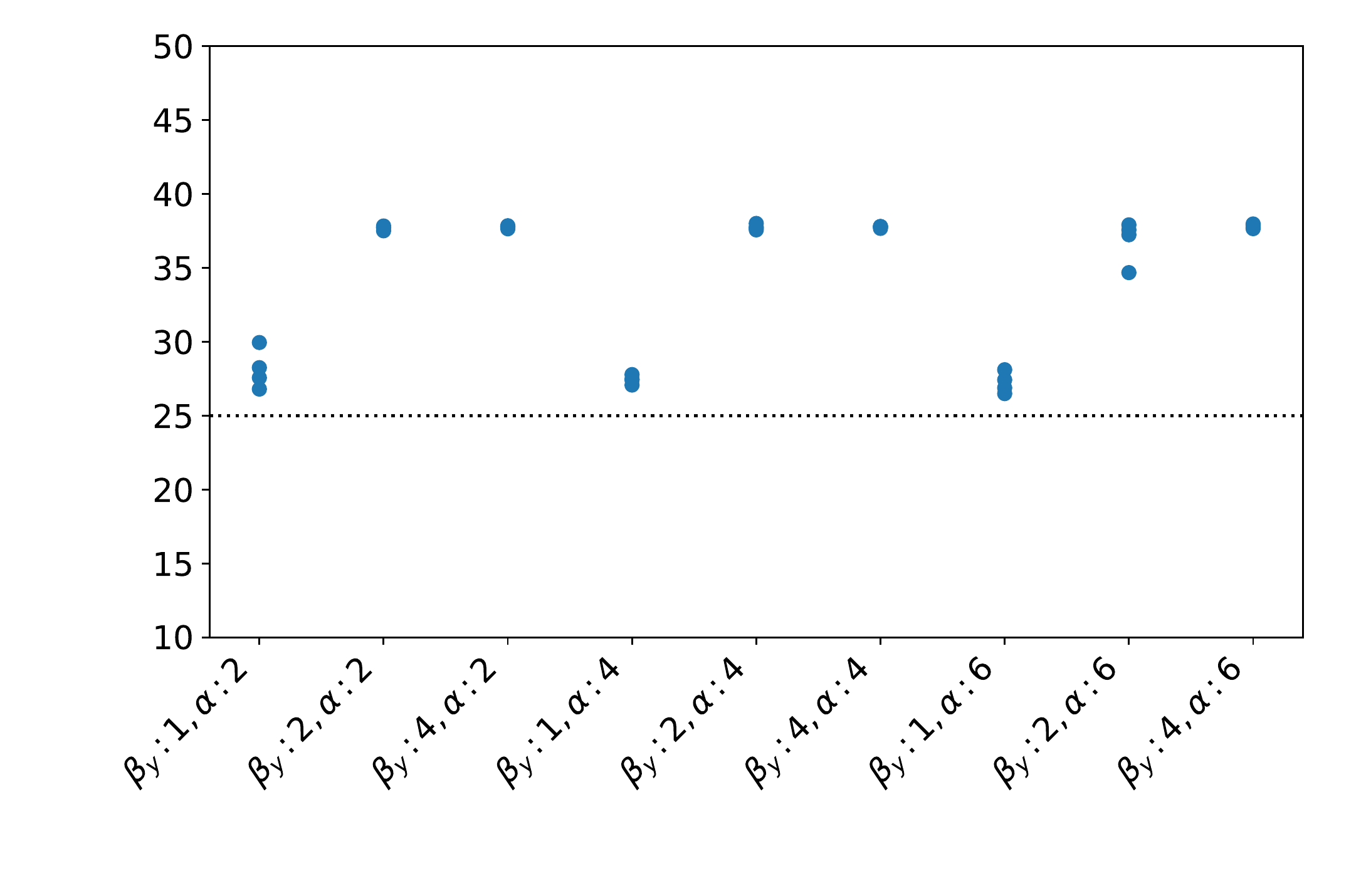}}
    \subfloat[ADA-GVAE\label{fig:adavae}]{\includegraphics[height=.32\textwidth]{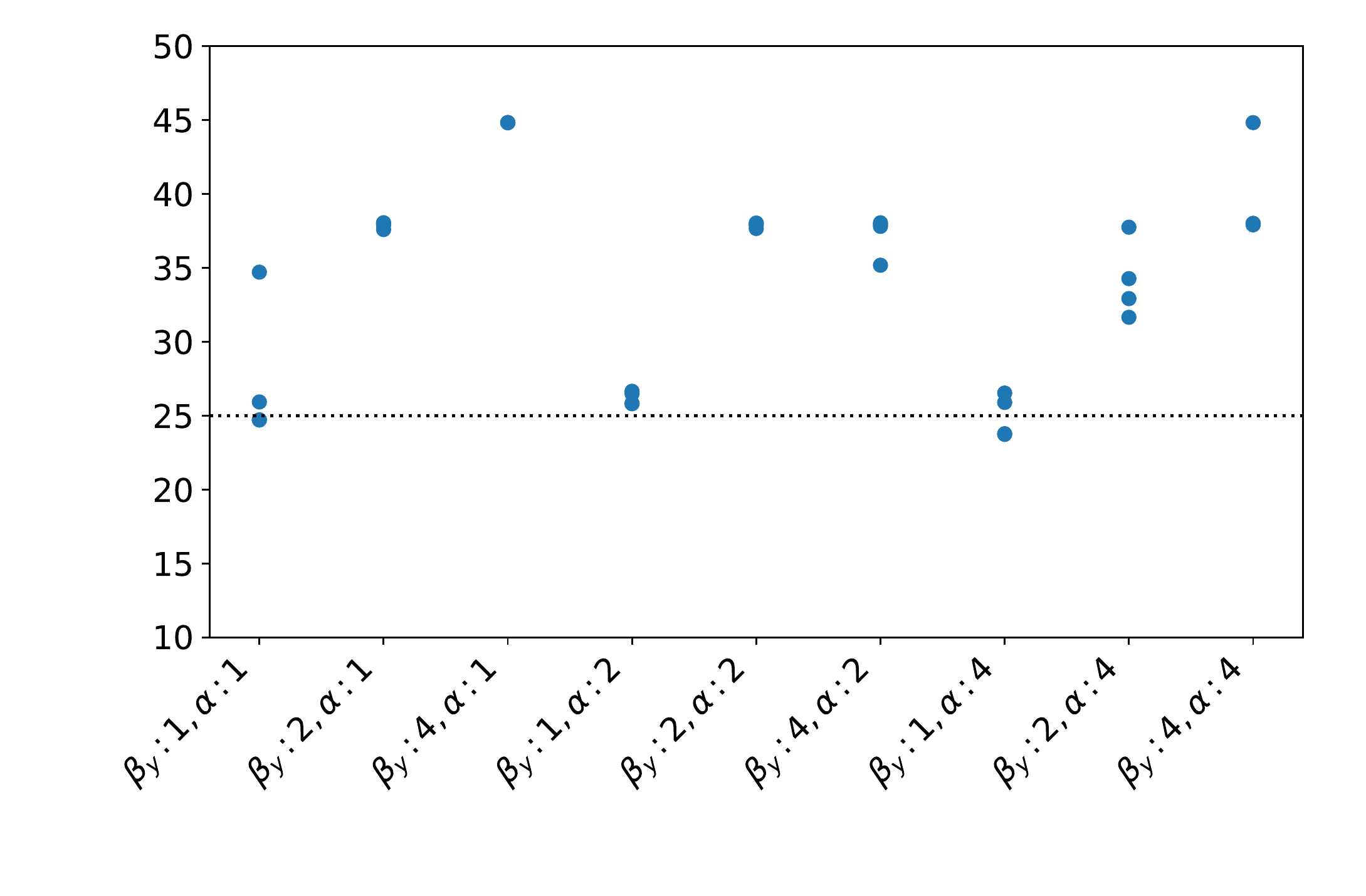}}
    \\
    \centering
    \subfloat[VAE-CE\label{fig:avaece}]{\includegraphics[height=.32\textwidth]{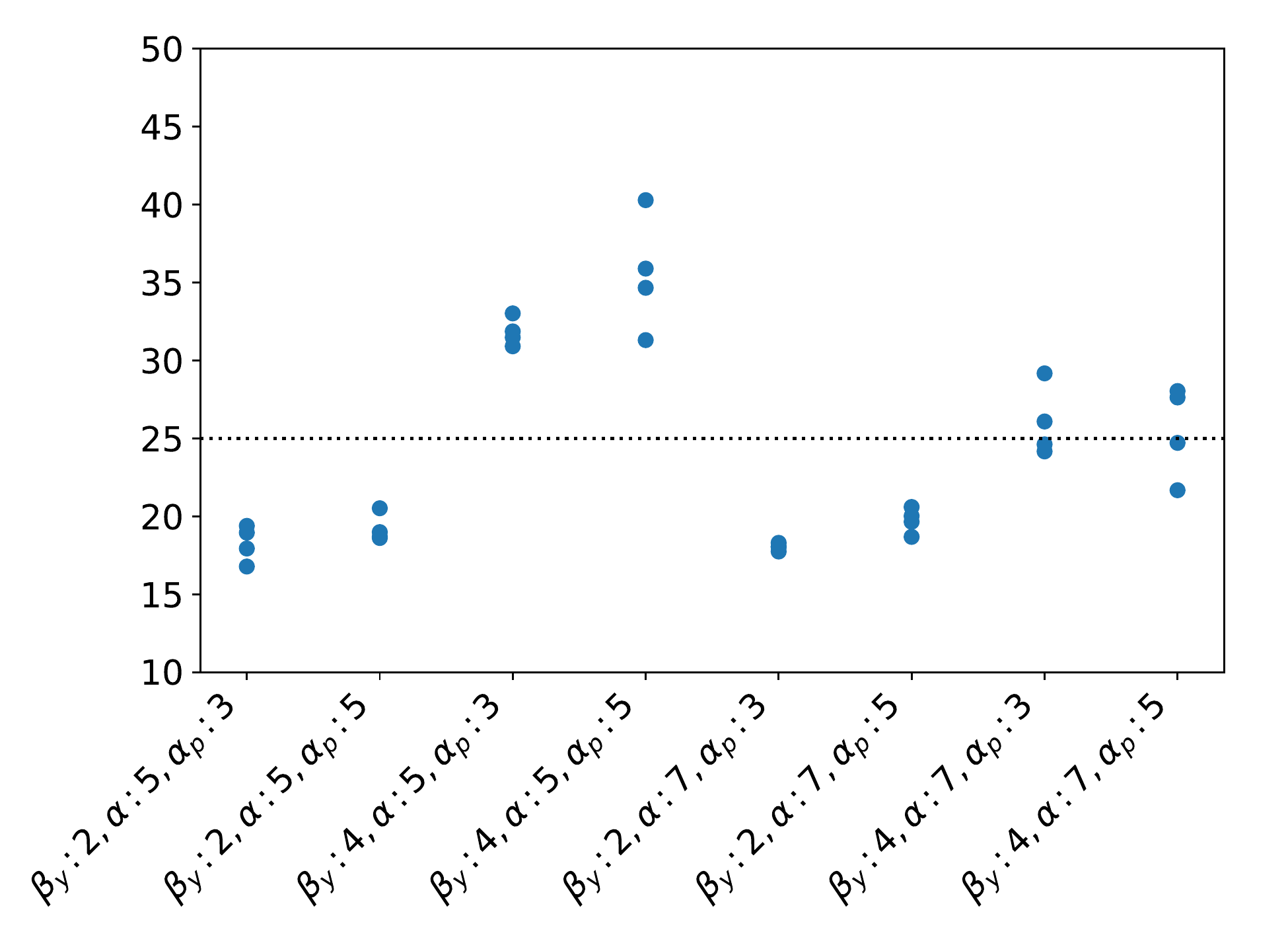}}
    \caption{The minimum $eac$ of each model-run (on validation data). The x-axis depicts the different configurations, whereas the y-axis depicts the $eac$ (lower is better).}%
    \vspace{3cm}
    \label{fig:allplots}%
\end{figure*}

For each model type we consider the Cartesian product of the hyperparameter values denoted in Table~\ref{ap:hpv}. We train each configuration four times, giving us a total of 176 models. For each configuration we compute the average $eac$ and select the hyperparameters corresponding to the lowest cost. These identified hyperparameters are depicted in Table~\ref{ap:hpf}. An overview of all model runs is provided in Fig.~\ref{fig:allplots}.

$CD$ is trained beforehand. For simplicity, we used a single configuration that gave a satisfactory test-set change-pair accuracy ($96.4\%$ on the synthetic data and $87.1\%$ on MNIST-pairs). This configuration is depicted in Table~\ref{ap:hpcd}. $CD$ was trained for \num{5000000} steps (for both datasets).

\begin{table}[H]
\begin{center}
\begin{tabular}{l l}\hline
 \textbf{Setting}& \textbf{Value}\\ \hline
Batch size & 128 \\
Optimizer & Adam\cite{kingma2014adam} \\ 
Adam: learning rate & 0.001\\
Adam: $\beta_1$ & 0.9\\
Adam: $\beta_2$ & 0.999\\
Adam: $\epsilon$ & 0.0001\\
\hline
\end{tabular}
\end{center}
\caption{Shared settings.}\label{ap:hpsh}
\end{table}
\begin{table}[H]
\begin{center}
\begin{tabular}{l l l}\hline
 \textbf{}& \textbf{Meaning} & \textbf{Model} \\ \hline
$\beta_y$ & $z_y$ KL divergence weight & $all$ \\ 
$\beta_x$ & $z_x$ KL divergence weight & $all$ \\ 
$\alpha$ & class-label classification weight & $all$ \\ 
$\alpha_d$ & per-dimension classification weight & LVAE \\
$\alpha_r$ & $D$-prediction weight & VAE-CE \\
$\alpha_p$ & $CD$-prediction weight & VAE-CE \\
\hline
\end{tabular}
\end{center}
\caption{All hyperparameters.}\label{ap:hp}
\end{table}
\begin{table}[H]
\begin{center}
\begin{tabular}{l l l}\hline
 \textbf{Model} & \textbf{} & \textbf{Values} \\ \hline
DVAE & $\beta_y$ &  \{2, 4\} \\
 & $\alpha$ &  \{5, 10, 15\} \\ \hline
LVAE & $\beta_y$ &  \{1, 2\} \\
 & $\alpha$ &  \{5, 7\} \\
 & $\alpha_d$ &  \{20, 25, 30\} \\ \hline
 GVAE & $\beta_y$ &  \{1, 2, 4\} \\
 & $\alpha$ &  \{2, 4, 6\} \\ \hline
 ADA-GVAE & $\beta_y$ &  \{1, 2, 4\} \\
 & $\alpha$ &  \{1, 2, 4\} \\ \hline
 VAE-CE & $\beta_y$ &  \{2, 4\} \\
 & $\alpha$ &  \{5, 7\} \\ 
  & $\alpha_p$ &  \{3, 5\} \\
\hline
\end{tabular}
\end{center}
\caption{All explored hyperparameter values. Parameters not mentioned are set to 1. The Cartesian product of the per-parameter values denotes all configurations we train.}\label{ap:hpv}
\end{table}
\begin{table}[H]
\begin{center}
\begin{tabular}{l l l}\hline
 \textbf{Model}& \textbf{} & \textbf{Value} \\ \hline
DVAE & $\beta_y$ &  2 \\
 & $\alpha$ &  10 \\ \hline
LVAE & $\beta_y$ &  1 \\
 & $\alpha$ &  7 \\
 & $\alpha_d$ &  20 \\ \hline
 GVAE & $\beta_y$ &  1 \\
 & $\alpha$ &  6 \\ \hline
 ADA-GVAE & $\beta_y$ &  1 \\
 & $\alpha$ &  4 \\ \hline
 VAE-CE & $\beta_y$ &  2 \\
 & $\alpha$ &  7 \\ 
  & $\alpha_p$ & 3 \\
\hline
\end{tabular}
\end{center}
\caption{The selected hyperparameter values.}\label{ap:hpf}
\end{table}
\begin{table}[H]
\begin{center}
\begin{tabular}{l l l}\hline
 \textbf{}& \textbf{Meaning} &\textbf{Value} \\ \hline
$\beta_y$ & $z_y$ KL divergence weight &  1\\ 
$\beta_x$ & $z_x$ KL divergence weight &  0.5\\ 
$\alpha$ & class-label classification weight & 16 \\ 
$\alpha_c$ & $DISC$ classification weight & 50\\
\hline
\end{tabular}
\end{center}
\caption{$CD$ hyperparameters.}\label{ap:hpcd}
\end{table}

\newpage
{\small
\bibliographystyle{ieee_fullname}
\bibliography{out}
}